%% file: main.tex
\documentclass[sigconf,nonacm]{acmart}
\setcopyright{none}
\renewcommand\footnotetextcopyrightpermission[1]{}

\usepackage{amsmath,amsfonts}
\usepackage{booktabs}
\usepackage{graphicx}
\usepackage{enumitem}
\usepackage{tikz}
\usetikzlibrary{arrows.meta,patterns}
\newcommand{\ManuscriptMainPaper}{}

\begin{document}

\title{Yielding Fairness: Coupled Learning and Fairness for Value-Weighted Ranking}

\author{Ami Tavory}
\affiliation{\institution{Meta Platforms}\country{}}
\author{Yaniv Sheena}
\affiliation{\institution{Meta Platforms}\country{}}
\author{Shaked Bar}
\affiliation{\institution{Meta Platforms}\country{}}
\author{Yair Zagdansky}
\affiliation{\institution{Meta Platforms}\country{}}



\input{sections/score_fixed_counts}
\input{sections/fairness_strengthening_macros}
\input{sections/corrected_experiment_macros}
\input{sections/direct_selector_macros}
\input{sections/empirical_cut_macros}
\input{sections/objection_safe_macros}
\input{sections/attainable_disparity_macros}
\input{sections/abstract}

\maketitle
\hypersetup{pdfauthor={Ami Tavory, Yaniv Sheena, Shaked Bar, Yair Zagdansky}}

\input{sections/introduction}
\input{sections/yield_model_form}
\input{sections/remediable_adverse_impact}
\input{sections/group_fairness_conditions}
\input{sections/experiments}
\input{sections/discussion_limitations_and_conclusion}

\input{sections/ethical_considerations}

\bibliographystyle{ACM-Reference-Format}
\bibliography{references}

\appendix
\input{sections/experiment_protocol_appendix}
\input{sections/preprocessing_appendix}
\input{sections/regularity_appendix}
\input{sections/objection_algorithm_appendix}
\input{sections/score_account_results_appendix}
\input{sections/group_disparity_decomposition_appendix}
\input{sections/normalized_price_sensitivity_appendix}

\input{sections/direct_selector_details}
\input{sections/empirical_cut_details}
\input{sections/objection_safe_details}

\end{document}

%% file: sections/score_fixed_counts.tex
\newcommand{\ManuscriptRawGiniPositive}{9}
\newcommand{\ManuscriptRawGiniTotal}{10}
\newcommand{\ManuscriptRawCapturePositive}{8}
\newcommand{\ManuscriptRawCaptureTotal}{10}

%% file: sections/fairness_strengthening_macros.tex
\newcommand{\ManuscriptFWCellTotal}{240}
\newcommand{\ManuscriptFWSignedGapReduced}{151}
\newcommand{\ManuscriptFWSignedGapReducedWithIncrease}{150}
\newcommand{\ManuscriptNWCappedGapReduced}{174}
\newcommand{\ManuscriptNWNoIncrease}{151}
\newcommand{\ManuscriptNWAllConstraintsMet}{53}
\newcommand{\ManuscriptNWImprovesUniform}{106}
\newcommand{\ManuscriptNWImprovesUniformWithConstraints}{34}
\newcommand{\ManuscriptNWMedianRetainedPercent}{98.04}
\newcommand{\ManuscriptNWMedianGapImprovement}{0.0064}

%% file: sections/corrected_experiment_macros.tex
\newcommand{\ManuscriptCorrectedCells}{240}
\newcommand{\ManuscriptCorrectedCommonGapValFeasible}{240}
\newcommand{\ManuscriptCorrectedCommonGapCAccepted}{75}
\newcommand{\ManuscriptCorrectedCommonGapTProposalNoIncrease}{231}
\newcommand{\ManuscriptCorrectedCommonGapTAppliedNoIncrease}{238}
\newcommand{\ManuscriptCorrectedCommonGapTProposalJoint}{80}
\newcommand{\ManuscriptCorrectedCommonGapTAppliedJoint}{125}
\newcommand{\ManuscriptCorrectedCommonGapTAppliedQualifiedGain}{22}
\newcommand{\ManuscriptCorrectedCommonCapsValFeasible}{240}
\newcommand{\ManuscriptCorrectedCommonCapsCAccepted}{75}
\newcommand{\ManuscriptCorrectedCommonCapsTProposalNoIncrease}{231}
\newcommand{\ManuscriptCorrectedCommonCapsTAppliedNoIncrease}{238}
\newcommand{\ManuscriptCorrectedCommonCapsTProposalJoint}{80}
\newcommand{\ManuscriptCorrectedCommonCapsTAppliedJoint}{125}
\newcommand{\ManuscriptCorrectedCommonCapsTAppliedQualifiedGain}{22}
\newcommand{\ManuscriptCorrectedUnionGapValFeasible}{240}
\newcommand{\ManuscriptCorrectedUnionGapCAccepted}{15}
\newcommand{\ManuscriptCorrectedUnionGapTProposalNoIncrease}{60}
\newcommand{\ManuscriptCorrectedUnionGapTAppliedNoIncrease}{238}
\newcommand{\ManuscriptCorrectedUnionGapTProposalJoint}{19}
\newcommand{\ManuscriptCorrectedUnionGapTAppliedJoint}{127}
\newcommand{\ManuscriptCorrectedUnionGapTAppliedQualifiedGain}{4}
\newcommand{\ManuscriptCorrectedUnionCapsValFeasible}{240}
\newcommand{\ManuscriptCorrectedUnionCapsCAccepted}{42}
\newcommand{\ManuscriptCorrectedUnionCapsTProposalNoIncrease}{144}
\newcommand{\ManuscriptCorrectedUnionCapsTAppliedNoIncrease}{229}
\newcommand{\ManuscriptCorrectedUnionCapsTProposalJoint}{53}
\newcommand{\ManuscriptCorrectedUnionCapsTAppliedJoint}{119}
\newcommand{\ManuscriptCorrectedUnionCapsTAppliedQualifiedGain}{10}

%% file: sections/direct_selector_macros.tex
\newcommand{\ManuscriptDirectCells}{240}
\newcommand{\ManuscriptDirectFullCommonAccepted}{48}
\newcommand{\ManuscriptDirectFullCommonFittedFloor}{126}
\newcommand{\ManuscriptDirectFullCommonProposalG}{205}
\newcommand{\ManuscriptDirectFullCommonProposalN}{203}
\newcommand{\ManuscriptDirectFullCommonProposalF}{37}
\newcommand{\ManuscriptDirectFullCommonProposalNoIncrease}{233}
\newcommand{\ManuscriptDirectFullCommonProposalJoint}{45}
\newcommand{\ManuscriptDirectFullCommonProposalU}{35}
\newcommand{\ManuscriptDirectFullCommonProposalQ}{17}
\newcommand{\ManuscriptDirectFullCommonProposalQFrozen}{20}
\newcommand{\ManuscriptDirectFullCommonProposalProgramFrozen}{75}
\newcommand{\ManuscriptDirectFullCommonProposalUQ}{17}
\newcommand{\ManuscriptDirectFullCommonProposalUQFrozen}{19}
\newcommand{\ManuscriptDirectFullCommonProposalRetentionPercent}{91.74}
\newcommand{\ManuscriptDirectFullCommonAppliedG}{38}
\newcommand{\ManuscriptDirectFullCommonAppliedN}{38}
\newcommand{\ManuscriptDirectFullCommonAppliedF}{23}
\newcommand{\ManuscriptDirectFullCommonAppliedNoIncrease}{240}
\newcommand{\ManuscriptDirectFullCommonAppliedJoint}{124}
\newcommand{\ManuscriptDirectFullCommonAppliedU}{21}
\newcommand{\ManuscriptDirectFullCommonAppliedQ}{13}
\newcommand{\ManuscriptDirectFullCommonAppliedQFrozen}{19}
\newcommand{\ManuscriptDirectFullCommonAppliedProgramFrozen}{23}
\newcommand{\ManuscriptDirectFullCommonAppliedUQ}{13}
\newcommand{\ManuscriptDirectFullCommonAppliedUQFrozen}{13}
\newcommand{\ManuscriptDirectFullCommonAppliedRetentionPercent}{100.00}
\newcommand{\ManuscriptDirectFullUnionAccepted}{45}
\newcommand{\ManuscriptDirectFullUnionFittedFloor}{222}
\newcommand{\ManuscriptDirectFullUnionProposalG}{175}
\newcommand{\ManuscriptDirectFullUnionProposalN}{107}
\newcommand{\ManuscriptDirectFullUnionProposalF}{36}
\newcommand{\ManuscriptDirectFullUnionProposalNoIncrease}{141}
\newcommand{\ManuscriptDirectFullUnionProposalJoint}{49}
\newcommand{\ManuscriptDirectFullUnionProposalU}{34}
\newcommand{\ManuscriptDirectFullUnionProposalQ}{20}
\newcommand{\ManuscriptDirectFullUnionProposalQFrozen}{20}
\newcommand{\ManuscriptDirectFullUnionProposalProgramFrozen}{40}
\newcommand{\ManuscriptDirectFullUnionProposalUQ}{20}
\newcommand{\ManuscriptDirectFullUnionProposalUQFrozen}{12}
\newcommand{\ManuscriptDirectFullUnionProposalRetentionPercent}{99.08}
\newcommand{\ManuscriptDirectFullUnionAppliedG}{32}
\newcommand{\ManuscriptDirectFullUnionAppliedN}{29}
\newcommand{\ManuscriptDirectFullUnionAppliedF}{24}
\newcommand{\ManuscriptDirectFullUnionAppliedNoIncrease}{234}
\newcommand{\ManuscriptDirectFullUnionAppliedJoint}{121}
\newcommand{\ManuscriptDirectFullUnionAppliedU}{22}
\newcommand{\ManuscriptDirectFullUnionAppliedQ}{15}
\newcommand{\ManuscriptDirectFullUnionAppliedQFrozen}{16}
\newcommand{\ManuscriptDirectFullUnionAppliedProgramFrozen}{18}
\newcommand{\ManuscriptDirectFullUnionAppliedUQ}{15}
\newcommand{\ManuscriptDirectFullUnionAppliedUQFrozen}{9}
\newcommand{\ManuscriptDirectFullUnionAppliedRetentionPercent}{100.00}
\newcommand{\ManuscriptDirectFullDualAccepted}{38}
\newcommand{\ManuscriptDirectFullDualFittedFloor}{208}
\newcommand{\ManuscriptDirectFullDualProposalG}{171}
\newcommand{\ManuscriptDirectFullDualProposalN}{72}
\newcommand{\ManuscriptDirectFullDualProposalF}{27}
\newcommand{\ManuscriptDirectFullDualProposalNoIncrease}{102}
\newcommand{\ManuscriptDirectFullDualProposalJoint}{36}
\newcommand{\ManuscriptDirectFullDualProposalU}{26}
\newcommand{\ManuscriptDirectFullDualProposalQ}{7}
\newcommand{\ManuscriptDirectFullDualProposalQFrozen}{16}
\newcommand{\ManuscriptDirectFullDualProposalProgramFrozen}{23}
\newcommand{\ManuscriptDirectFullDualProposalUQ}{7}
\newcommand{\ManuscriptDirectFullDualProposalUQFrozen}{5}
\newcommand{\ManuscriptDirectFullDualProposalRetentionPercent}{99.83}
\newcommand{\ManuscriptDirectFullDualAppliedG}{27}
\newcommand{\ManuscriptDirectFullDualAppliedN}{24}
\newcommand{\ManuscriptDirectFullDualAppliedF}{14}
\newcommand{\ManuscriptDirectFullDualAppliedNoIncrease}{233}
\newcommand{\ManuscriptDirectFullDualAppliedJoint}{125}
\newcommand{\ManuscriptDirectFullDualAppliedU}{14}
\newcommand{\ManuscriptDirectFullDualAppliedQ}{4}
\newcommand{\ManuscriptDirectFullDualAppliedQFrozen}{7}
\newcommand{\ManuscriptDirectFullDualAppliedProgramFrozen}{10}
\newcommand{\ManuscriptDirectFullDualAppliedUQ}{4}
\newcommand{\ManuscriptDirectFullDualAppliedUQFrozen}{2}
\newcommand{\ManuscriptDirectFullDualAppliedRetentionPercent}{100.00}
\newcommand{\ManuscriptDirectTCommonAccepted}{48}
\newcommand{\ManuscriptDirectTCommonFittedFloor}{126}
\newcommand{\ManuscriptDirectTCommonProposalG}{206}
\newcommand{\ManuscriptDirectTCommonProposalN}{205}
\newcommand{\ManuscriptDirectTCommonProposalF}{41}
\newcommand{\ManuscriptDirectTCommonProposalNoIncrease}{232}
\newcommand{\ManuscriptDirectTCommonProposalJoint}{50}
\newcommand{\ManuscriptDirectTCommonProposalU}{38}
\newcommand{\ManuscriptDirectTCommonProposalQ}{19}
\newcommand{\ManuscriptDirectTCommonProposalQFrozen}{26}
\newcommand{\ManuscriptDirectTCommonProposalProgramFrozen}{66}
\newcommand{\ManuscriptDirectTCommonProposalUQ}{18}
\newcommand{\ManuscriptDirectTCommonProposalUQFrozen}{23}
\newcommand{\ManuscriptDirectTCommonProposalRetentionPercent}{92.76}
\newcommand{\ManuscriptDirectTCommonAppliedG}{41}
\newcommand{\ManuscriptDirectTCommonAppliedN}{40}
\newcommand{\ManuscriptDirectTCommonAppliedF}{17}
\newcommand{\ManuscriptDirectTCommonAppliedNoIncrease}{238}
\newcommand{\ManuscriptDirectTCommonAppliedJoint}{121}
\newcommand{\ManuscriptDirectTCommonAppliedU}{16}
\newcommand{\ManuscriptDirectTCommonAppliedQ}{8}
\newcommand{\ManuscriptDirectTCommonAppliedQFrozen}{21}
\newcommand{\ManuscriptDirectTCommonAppliedProgramFrozen}{26}
\newcommand{\ManuscriptDirectTCommonAppliedUQ}{8}
\newcommand{\ManuscriptDirectTCommonAppliedUQFrozen}{9}
\newcommand{\ManuscriptDirectTCommonAppliedRetentionPercent}{100.00}
\newcommand{\ManuscriptDirectTUnionAccepted}{45}
\newcommand{\ManuscriptDirectTUnionFittedFloor}{222}
\newcommand{\ManuscriptDirectTUnionProposalG}{164}
\newcommand{\ManuscriptDirectTUnionProposalN}{107}
\newcommand{\ManuscriptDirectTUnionProposalF}{43}
\newcommand{\ManuscriptDirectTUnionProposalNoIncrease}{150}
\newcommand{\ManuscriptDirectTUnionProposalJoint}{63}
\newcommand{\ManuscriptDirectTUnionProposalU}{40}
\newcommand{\ManuscriptDirectTUnionProposalQ}{18}
\newcommand{\ManuscriptDirectTUnionProposalQFrozen}{23}
\newcommand{\ManuscriptDirectTUnionProposalProgramFrozen}{37}
\newcommand{\ManuscriptDirectTUnionProposalUQ}{18}
\newcommand{\ManuscriptDirectTUnionProposalUQFrozen}{15}
\newcommand{\ManuscriptDirectTUnionProposalRetentionPercent}{99.24}
\newcommand{\ManuscriptDirectTUnionAppliedG}{33}
\newcommand{\ManuscriptDirectTUnionAppliedN}{24}
\newcommand{\ManuscriptDirectTUnionAppliedF}{16}
\newcommand{\ManuscriptDirectTUnionAppliedNoIncrease}{228}
\newcommand{\ManuscriptDirectTUnionAppliedJoint}{119}
\newcommand{\ManuscriptDirectTUnionAppliedU}{15}
\newcommand{\ManuscriptDirectTUnionAppliedQ}{9}
\newcommand{\ManuscriptDirectTUnionAppliedQFrozen}{19}
\newcommand{\ManuscriptDirectTUnionAppliedProgramFrozen}{20}
\newcommand{\ManuscriptDirectTUnionAppliedUQ}{9}
\newcommand{\ManuscriptDirectTUnionAppliedUQFrozen}{8}
\newcommand{\ManuscriptDirectTUnionAppliedRetentionPercent}{100.00}
\newcommand{\ManuscriptDirectTDualAccepted}{38}
\newcommand{\ManuscriptDirectTDualFittedFloor}{208}
\newcommand{\ManuscriptDirectTDualProposalG}{166}
\newcommand{\ManuscriptDirectTDualProposalN}{58}
\newcommand{\ManuscriptDirectTDualProposalF}{22}
\newcommand{\ManuscriptDirectTDualProposalNoIncrease}{80}
\newcommand{\ManuscriptDirectTDualProposalJoint}{29}
\newcommand{\ManuscriptDirectTDualProposalU}{21}
\newcommand{\ManuscriptDirectTDualProposalQ}{3}
\newcommand{\ManuscriptDirectTDualProposalQFrozen}{12}
\newcommand{\ManuscriptDirectTDualProposalProgramFrozen}{23}
\newcommand{\ManuscriptDirectTDualProposalUQ}{3}
\newcommand{\ManuscriptDirectTDualProposalUQFrozen}{3}
\newcommand{\ManuscriptDirectTDualProposalRetentionPercent}{100.04}
\newcommand{\ManuscriptDirectTDualAppliedG}{28}
\newcommand{\ManuscriptDirectTDualAppliedN}{17}
\newcommand{\ManuscriptDirectTDualAppliedF}{7}
\newcommand{\ManuscriptDirectTDualAppliedNoIncrease}{223}
\newcommand{\ManuscriptDirectTDualAppliedJoint}{124}
\newcommand{\ManuscriptDirectTDualAppliedU}{7}
\newcommand{\ManuscriptDirectTDualAppliedQ}{0}
\newcommand{\ManuscriptDirectTDualAppliedQFrozen}{12}
\newcommand{\ManuscriptDirectTDualAppliedProgramFrozen}{15}
\newcommand{\ManuscriptDirectTDualAppliedUQ}{0}
\newcommand{\ManuscriptDirectTDualAppliedUQFrozen}{1}
\newcommand{\ManuscriptDirectTDualAppliedRetentionPercent}{100.00}

%% file: sections/empirical_cut_macros.tex
\newcommand{\ManuscriptEmpiricalCells}{240}
\newcommand{\ManuscriptEmpiricalFullAccepted}{27}
\newcommand{\ManuscriptEmpiricalFullSelectionFloor}{234}
\newcommand{\ManuscriptEmpiricalFullProposalG}{186}
\newcommand{\ManuscriptEmpiricalFullProposalN}{87}
\newcommand{\ManuscriptEmpiricalFullProposalF}{24}
\newcommand{\ManuscriptEmpiricalFullProposalNoIncrease}{100}
\newcommand{\ManuscriptEmpiricalFullProposalJoint}{27}
\newcommand{\ManuscriptEmpiricalFullProposalU}{23}
\newcommand{\ManuscriptEmpiricalFullProposalQ}{15}
\newcommand{\ManuscriptEmpiricalFullProposalQFrozen}{15}
\newcommand{\ManuscriptEmpiricalFullProposalProgramFrozen}{31}
\newcommand{\ManuscriptEmpiricalFullProposalUQ}{14}
\newcommand{\ManuscriptEmpiricalFullProposalUQFrozen}{10}
\newcommand{\ManuscriptEmpiricalFullProposalRetentionPercent}{97.86}
\newcommand{\ManuscriptEmpiricalFullAppliedG}{24}
\newcommand{\ManuscriptEmpiricalFullAppliedN}{21}
\newcommand{\ManuscriptEmpiricalFullAppliedF}{16}
\newcommand{\ManuscriptEmpiricalFullAppliedNoIncrease}{237}
\newcommand{\ManuscriptEmpiricalFullAppliedJoint}{125}
\newcommand{\ManuscriptEmpiricalFullAppliedU}{16}
\newcommand{\ManuscriptEmpiricalFullAppliedQ}{9}
\newcommand{\ManuscriptEmpiricalFullAppliedQFrozen}{13}
\newcommand{\ManuscriptEmpiricalFullAppliedProgramFrozen}{15}
\newcommand{\ManuscriptEmpiricalFullAppliedUQ}{9}
\newcommand{\ManuscriptEmpiricalFullAppliedUQFrozen}{7}
\newcommand{\ManuscriptEmpiricalFullAppliedRetentionPercent}{100.00}
\newcommand{\ManuscriptEmpiricalTAccepted}{27}
\newcommand{\ManuscriptEmpiricalTSelectionFloor}{234}
\newcommand{\ManuscriptEmpiricalTProposalG}{174}
\newcommand{\ManuscriptEmpiricalTProposalN}{93}
\newcommand{\ManuscriptEmpiricalTProposalF}{23}
\newcommand{\ManuscriptEmpiricalTProposalNoIncrease}{113}
\newcommand{\ManuscriptEmpiricalTProposalJoint}{29}
\newcommand{\ManuscriptEmpiricalTProposalU}{23}
\newcommand{\ManuscriptEmpiricalTProposalQ}{11}
\newcommand{\ManuscriptEmpiricalTProposalQFrozen}{19}
\newcommand{\ManuscriptEmpiricalTProposalProgramFrozen}{33}
\newcommand{\ManuscriptEmpiricalTProposalUQ}{11}
\newcommand{\ManuscriptEmpiricalTProposalUQFrozen}{11}
\newcommand{\ManuscriptEmpiricalTProposalRetentionPercent}{98.58}
\newcommand{\ManuscriptEmpiricalTAppliedG}{22}
\newcommand{\ManuscriptEmpiricalTAppliedN}{16}
\newcommand{\ManuscriptEmpiricalTAppliedF}{9}
\newcommand{\ManuscriptEmpiricalTAppliedNoIncrease}{233}
\newcommand{\ManuscriptEmpiricalTAppliedJoint}{123}
\newcommand{\ManuscriptEmpiricalTAppliedU}{9}
\newcommand{\ManuscriptEmpiricalTAppliedQ}{4}
\newcommand{\ManuscriptEmpiricalTAppliedQFrozen}{13}
\newcommand{\ManuscriptEmpiricalTAppliedProgramFrozen}{16}
\newcommand{\ManuscriptEmpiricalTAppliedUQ}{4}
\newcommand{\ManuscriptEmpiricalTAppliedUQFrozen}{3}
\newcommand{\ManuscriptEmpiricalTAppliedRetentionPercent}{100.00}

%% file: sections/objection_safe_macros.tex
\newcommand{\ManuscriptSafeCAppliedAllUnitsRemovalJointSupport}{236}
\newcommand{\ManuscriptSafeCAppliedAllUnitsRemovalJointUndefined}{4}
\newcommand{\ManuscriptSafeCAppliedAllUnitsRemovalSupport}{236}
\newcommand{\ManuscriptSafeCAppliedAllUnitsRemovalUndefined}{4}
\newcommand{\ManuscriptSafeCAppliedAllUnitsStrictRemoved}{236}
\newcommand{\ManuscriptSafeCAppliedAllUnitsStrictRemovedJoint}{193}
\newcommand{\ManuscriptSafeCAppliedCapture}{197}
\newcommand{\ManuscriptSafeCAppliedCountRemoval}{46.46}
\newcommand{\ManuscriptSafeCAppliedCountRemovalJointSupport}{240}
\newcommand{\ManuscriptSafeCAppliedCountRemovalJointUndefined}{0}
\newcommand{\ManuscriptSafeCAppliedCountRemovalMedian}{36.41}
\newcommand{\ManuscriptSafeCAppliedCountRemovalSupport}{240}
\newcommand{\ManuscriptSafeCAppliedCountRemovalUndefined}{0}
\newcommand{\ManuscriptSafeCAppliedCountStrictRemoved}{240}
\newcommand{\ManuscriptSafeCAppliedCountStrictRemovedJoint}{197}
\newcommand{\ManuscriptSafeCAppliedJoint}{197}
\newcommand{\ManuscriptSafeCAppliedLossRemoval}{46.46}
\newcommand{\ManuscriptSafeCAppliedLossRemovalJointSupport}{240}
\newcommand{\ManuscriptSafeCAppliedLossRemovalJointUndefined}{0}
\newcommand{\ManuscriptSafeCAppliedLossRemovalMedian}{36.41}
\newcommand{\ManuscriptSafeCAppliedLossRemovalSupport}{240}
\newcommand{\ManuscriptSafeCAppliedLossRemovalUndefined}{0}
\newcommand{\ManuscriptSafeCAppliedLossStrictRemoved}{240}
\newcommand{\ManuscriptSafeCAppliedLossStrictRemovedJoint}{197}
\newcommand{\ManuscriptSafeCAppliedMeanRetained}{95.94}
\newcommand{\ManuscriptSafeCAppliedRetained}{95.00}
\newcommand{\ManuscriptSafeCAppliedValue}{239}
\newcommand{\ManuscriptSafeCAppliedValueRemoval}{47.06}
\newcommand{\ManuscriptSafeCAppliedValueRemovalJointSupport}{236}
\newcommand{\ManuscriptSafeCAppliedValueRemovalJointUndefined}{4}
\newcommand{\ManuscriptSafeCAppliedValueRemovalMedian}{37.83}
\newcommand{\ManuscriptSafeCAppliedValueRemovalSupport}{236}
\newcommand{\ManuscriptSafeCAppliedValueRemovalUndefined}{4}
\newcommand{\ManuscriptSafeCAppliedValueStrictRemoved}{236}
\newcommand{\ManuscriptSafeCAppliedValueStrictRemovedJoint}{193}
\newcommand{\ManuscriptSafeCCalibrationSolved}{197}
\newcommand{\ManuscriptSafeCFitSolved}{240}
\newcommand{\ManuscriptSafeCProposedAllUnitsRemovalJointSupport}{236}
\newcommand{\ManuscriptSafeCProposedAllUnitsRemovalJointUndefined}{4}
\newcommand{\ManuscriptSafeCProposedAllUnitsRemovalSupport}{236}
\newcommand{\ManuscriptSafeCProposedAllUnitsRemovalUndefined}{4}
\newcommand{\ManuscriptSafeCProposedAllUnitsStrictRemoved}{236}
\newcommand{\ManuscriptSafeCProposedAllUnitsStrictRemovedJoint}{47}
\newcommand{\ManuscriptSafeCProposedCapture}{90}
\newcommand{\ManuscriptSafeCProposedCountRemoval}{42.90}
\newcommand{\ManuscriptSafeCProposedCountRemovalJointSupport}{240}
\newcommand{\ManuscriptSafeCProposedCountRemovalJointUndefined}{0}
\newcommand{\ManuscriptSafeCProposedCountRemovalMedian}{34.11}
\newcommand{\ManuscriptSafeCProposedCountRemovalSupport}{240}
\newcommand{\ManuscriptSafeCProposedCountRemovalUndefined}{0}
\newcommand{\ManuscriptSafeCProposedCountStrictRemoved}{240}
\newcommand{\ManuscriptSafeCProposedCountStrictRemovedJoint}{47}
\newcommand{\ManuscriptSafeCProposedJoint}{47}
\newcommand{\ManuscriptSafeCProposedLossRemoval}{42.90}
\newcommand{\ManuscriptSafeCProposedLossRemovalJointSupport}{240}
\newcommand{\ManuscriptSafeCProposedLossRemovalJointUndefined}{0}
\newcommand{\ManuscriptSafeCProposedLossRemovalMedian}{34.11}
\newcommand{\ManuscriptSafeCProposedLossRemovalSupport}{240}
\newcommand{\ManuscriptSafeCProposedLossRemovalUndefined}{0}
\newcommand{\ManuscriptSafeCProposedLossStrictRemoved}{240}
\newcommand{\ManuscriptSafeCProposedLossStrictRemovedJoint}{47}
\newcommand{\ManuscriptSafeCProposedMeanRetained}{89.72}
\newcommand{\ManuscriptSafeCProposedRetained}{94.46}
\newcommand{\ManuscriptSafeCProposedValue}{148}
\newcommand{\ManuscriptSafeCProposedValueRemoval}{43.32}
\newcommand{\ManuscriptSafeCProposedValueRemovalJointSupport}{236}
\newcommand{\ManuscriptSafeCProposedValueRemovalJointUndefined}{4}
\newcommand{\ManuscriptSafeCProposedValueRemovalMedian}{34.27}
\newcommand{\ManuscriptSafeCProposedValueRemovalSupport}{236}
\newcommand{\ManuscriptSafeCProposedValueRemovalUndefined}{4}
\newcommand{\ManuscriptSafeCProposedValueStrictRemoved}{236}
\newcommand{\ManuscriptSafeCProposedValueStrictRemovedJoint}{47}
\newcommand{\ManuscriptSafeFullAppliedAllUnitsRemovalJointSupport}{240}
\newcommand{\ManuscriptSafeFullAppliedAllUnitsRemovalJointUndefined}{0}
\newcommand{\ManuscriptSafeFullAppliedAllUnitsRemovalSupport}{240}
\newcommand{\ManuscriptSafeFullAppliedAllUnitsRemovalUndefined}{0}
\newcommand{\ManuscriptSafeFullAppliedAllUnitsStrictRemoved}{240}
\newcommand{\ManuscriptSafeFullAppliedAllUnitsStrictRemovedJoint}{59}
\newcommand{\ManuscriptSafeFullAppliedCapture}{89}
\newcommand{\ManuscriptSafeFullAppliedCountRemoval}{46.46}
\newcommand{\ManuscriptSafeFullAppliedCountRemovalJointSupport}{240}
\newcommand{\ManuscriptSafeFullAppliedCountRemovalJointUndefined}{0}
\newcommand{\ManuscriptSafeFullAppliedCountRemovalMedian}{36.41}
\newcommand{\ManuscriptSafeFullAppliedCountRemovalSupport}{240}
\newcommand{\ManuscriptSafeFullAppliedCountRemovalUndefined}{0}
\newcommand{\ManuscriptSafeFullAppliedCountStrictRemoved}{240}
\newcommand{\ManuscriptSafeFullAppliedCountStrictRemovedJoint}{59}
\newcommand{\ManuscriptSafeFullAppliedJoint}{59}
\newcommand{\ManuscriptSafeFullAppliedLossRemoval}{46.46}
\newcommand{\ManuscriptSafeFullAppliedLossRemovalJointSupport}{240}
\newcommand{\ManuscriptSafeFullAppliedLossRemovalJointUndefined}{0}
\newcommand{\ManuscriptSafeFullAppliedLossRemovalMedian}{36.41}
\newcommand{\ManuscriptSafeFullAppliedLossRemovalSupport}{240}
\newcommand{\ManuscriptSafeFullAppliedLossRemovalUndefined}{0}
\newcommand{\ManuscriptSafeFullAppliedLossStrictRemoved}{240}
\newcommand{\ManuscriptSafeFullAppliedLossStrictRemovedJoint}{59}
\newcommand{\ManuscriptSafeFullAppliedMeanRetained}{93.90}
\newcommand{\ManuscriptSafeFullAppliedRetained}{93.98}
\newcommand{\ManuscriptSafeFullAppliedValue}{166}
\newcommand{\ManuscriptSafeFullAppliedValueRemoval}{46.46}
\newcommand{\ManuscriptSafeFullAppliedValueRemovalJointSupport}{240}
\newcommand{\ManuscriptSafeFullAppliedValueRemovalJointUndefined}{0}
\newcommand{\ManuscriptSafeFullAppliedValueRemovalMedian}{36.41}
\newcommand{\ManuscriptSafeFullAppliedValueRemovalSupport}{240}
\newcommand{\ManuscriptSafeFullAppliedValueRemovalUndefined}{0}
\newcommand{\ManuscriptSafeFullAppliedValueStrictRemoved}{240}
\newcommand{\ManuscriptSafeFullAppliedValueStrictRemovedJoint}{59}
\newcommand{\ManuscriptSafeFullCalibrationSolved}{197}
\newcommand{\ManuscriptSafeFullFitSolved}{240}
\newcommand{\ManuscriptSafeFullProposedAllUnitsRemovalJointSupport}{240}
\newcommand{\ManuscriptSafeFullProposedAllUnitsRemovalJointUndefined}{0}
\newcommand{\ManuscriptSafeFullProposedAllUnitsRemovalSupport}{240}
\newcommand{\ManuscriptSafeFullProposedAllUnitsRemovalUndefined}{0}
\newcommand{\ManuscriptSafeFullProposedAllUnitsStrictRemoved}{240}
\newcommand{\ManuscriptSafeFullProposedAllUnitsStrictRemovedJoint}{64}
\newcommand{\ManuscriptSafeFullProposedCapture}{100}
\newcommand{\ManuscriptSafeFullProposedCountRemoval}{42.90}
\newcommand{\ManuscriptSafeFullProposedCountRemovalJointSupport}{240}
\newcommand{\ManuscriptSafeFullProposedCountRemovalJointUndefined}{0}
\newcommand{\ManuscriptSafeFullProposedCountRemovalMedian}{34.11}
\newcommand{\ManuscriptSafeFullProposedCountRemovalSupport}{240}
\newcommand{\ManuscriptSafeFullProposedCountRemovalUndefined}{0}
\newcommand{\ManuscriptSafeFullProposedCountStrictRemoved}{240}
\newcommand{\ManuscriptSafeFullProposedCountStrictRemovedJoint}{64}
\newcommand{\ManuscriptSafeFullProposedJoint}{64}
\newcommand{\ManuscriptSafeFullProposedLossRemoval}{42.90}
\newcommand{\ManuscriptSafeFullProposedLossRemovalJointSupport}{240}
\newcommand{\ManuscriptSafeFullProposedLossRemovalJointUndefined}{0}
\newcommand{\ManuscriptSafeFullProposedLossRemovalMedian}{34.11}
\newcommand{\ManuscriptSafeFullProposedLossRemovalSupport}{240}
\newcommand{\ManuscriptSafeFullProposedLossRemovalUndefined}{0}
\newcommand{\ManuscriptSafeFullProposedLossStrictRemoved}{240}
\newcommand{\ManuscriptSafeFullProposedLossStrictRemovedJoint}{64}
\newcommand{\ManuscriptSafeFullProposedMeanRetained}{89.51}
\newcommand{\ManuscriptSafeFullProposedRetained}{94.70}
\newcommand{\ManuscriptSafeFullProposedValue}{155}
\newcommand{\ManuscriptSafeFullProposedValueRemoval}{42.90}
\newcommand{\ManuscriptSafeFullProposedValueRemovalJointSupport}{240}
\newcommand{\ManuscriptSafeFullProposedValueRemovalJointUndefined}{0}
\newcommand{\ManuscriptSafeFullProposedValueRemovalMedian}{34.11}
\newcommand{\ManuscriptSafeFullProposedValueRemovalSupport}{240}
\newcommand{\ManuscriptSafeFullProposedValueRemovalUndefined}{0}
\newcommand{\ManuscriptSafeFullProposedValueStrictRemoved}{240}
\newcommand{\ManuscriptSafeFullProposedValueStrictRemovedJoint}{64}
\newcommand{\ManuscriptSafeTAppliedAllUnitsRemovalJointSupport}{239}
\newcommand{\ManuscriptSafeTAppliedAllUnitsRemovalJointUndefined}{1}
\newcommand{\ManuscriptSafeTAppliedAllUnitsRemovalSupport}{239}
\newcommand{\ManuscriptSafeTAppliedAllUnitsRemovalUndefined}{1}
\newcommand{\ManuscriptSafeTAppliedAllUnitsStrictRemoved}{239}
\newcommand{\ManuscriptSafeTAppliedAllUnitsStrictRemovedJoint}{55}
\newcommand{\ManuscriptSafeTAppliedCapture}{93}
\newcommand{\ManuscriptSafeTAppliedCountRemoval}{46.46}
\newcommand{\ManuscriptSafeTAppliedCountRemovalJointSupport}{240}
\newcommand{\ManuscriptSafeTAppliedCountRemovalJointUndefined}{0}
\newcommand{\ManuscriptSafeTAppliedCountRemovalMedian}{36.41}
\newcommand{\ManuscriptSafeTAppliedCountRemovalSupport}{240}
\newcommand{\ManuscriptSafeTAppliedCountRemovalUndefined}{0}
\newcommand{\ManuscriptSafeTAppliedCountStrictRemoved}{240}
\newcommand{\ManuscriptSafeTAppliedCountStrictRemovedJoint}{56}
\newcommand{\ManuscriptSafeTAppliedJoint}{56}
\newcommand{\ManuscriptSafeTAppliedLossRemoval}{46.24}
\newcommand{\ManuscriptSafeTAppliedLossRemovalJointSupport}{239}
\newcommand{\ManuscriptSafeTAppliedLossRemovalJointUndefined}{1}
\newcommand{\ManuscriptSafeTAppliedLossRemovalMedian}{35.87}
\newcommand{\ManuscriptSafeTAppliedLossRemovalSupport}{239}
\newcommand{\ManuscriptSafeTAppliedLossRemovalUndefined}{1}
\newcommand{\ManuscriptSafeTAppliedLossStrictRemoved}{239}
\newcommand{\ManuscriptSafeTAppliedLossStrictRemovedJoint}{55}
\newcommand{\ManuscriptSafeTAppliedMeanRetained}{93.17}
\newcommand{\ManuscriptSafeTAppliedRetained}{92.97}
\newcommand{\ManuscriptSafeTAppliedValue}{154}
\newcommand{\ManuscriptSafeTAppliedValueRemoval}{46.46}
\newcommand{\ManuscriptSafeTAppliedValueRemovalJointSupport}{240}
\newcommand{\ManuscriptSafeTAppliedValueRemovalJointUndefined}{0}
\newcommand{\ManuscriptSafeTAppliedValueRemovalMedian}{36.41}
\newcommand{\ManuscriptSafeTAppliedValueRemovalSupport}{240}
\newcommand{\ManuscriptSafeTAppliedValueRemovalUndefined}{0}
\newcommand{\ManuscriptSafeTAppliedValueStrictRemoved}{240}
\newcommand{\ManuscriptSafeTAppliedValueStrictRemovedJoint}{56}
\newcommand{\ManuscriptSafeTCalibrationSolved}{197}
\newcommand{\ManuscriptSafeTFitSolved}{240}
\newcommand{\ManuscriptSafeTProposedAllUnitsRemovalJointSupport}{239}
\newcommand{\ManuscriptSafeTProposedAllUnitsRemovalJointUndefined}{1}
\newcommand{\ManuscriptSafeTProposedAllUnitsRemovalSupport}{239}
\newcommand{\ManuscriptSafeTProposedAllUnitsRemovalUndefined}{1}
\newcommand{\ManuscriptSafeTProposedAllUnitsStrictRemoved}{239}
\newcommand{\ManuscriptSafeTProposedAllUnitsStrictRemovedJoint}{62}
\newcommand{\ManuscriptSafeTProposedCapture}{102}
\newcommand{\ManuscriptSafeTProposedCountRemoval}{42.90}
\newcommand{\ManuscriptSafeTProposedCountRemovalJointSupport}{240}
\newcommand{\ManuscriptSafeTProposedCountRemovalJointUndefined}{0}
\newcommand{\ManuscriptSafeTProposedCountRemovalMedian}{34.11}
\newcommand{\ManuscriptSafeTProposedCountRemovalSupport}{240}
\newcommand{\ManuscriptSafeTProposedCountRemovalUndefined}{0}
\newcommand{\ManuscriptSafeTProposedCountStrictRemoved}{240}
\newcommand{\ManuscriptSafeTProposedCountStrictRemovedJoint}{63}
\newcommand{\ManuscriptSafeTProposedJoint}{63}
\newcommand{\ManuscriptSafeTProposedLossRemoval}{42.88}
\newcommand{\ManuscriptSafeTProposedLossRemovalJointSupport}{239}
\newcommand{\ManuscriptSafeTProposedLossRemovalJointUndefined}{1}
\newcommand{\ManuscriptSafeTProposedLossRemovalMedian}{34.11}
\newcommand{\ManuscriptSafeTProposedLossRemovalSupport}{239}
\newcommand{\ManuscriptSafeTProposedLossRemovalUndefined}{1}
\newcommand{\ManuscriptSafeTProposedLossStrictRemoved}{239}
\newcommand{\ManuscriptSafeTProposedLossStrictRemovedJoint}{62}
\newcommand{\ManuscriptSafeTProposedMeanRetained}{90.65}
\newcommand{\ManuscriptSafeTProposedRetained}{94.88}
\newcommand{\ManuscriptSafeTProposedValue}{147}
\newcommand{\ManuscriptSafeTProposedValueRemoval}{42.90}
\newcommand{\ManuscriptSafeTProposedValueRemovalJointSupport}{240}
\newcommand{\ManuscriptSafeTProposedValueRemovalJointUndefined}{0}
\newcommand{\ManuscriptSafeTProposedValueRemovalMedian}{34.11}
\newcommand{\ManuscriptSafeTProposedValueRemovalSupport}{240}
\newcommand{\ManuscriptSafeTProposedValueRemovalUndefined}{0}
\newcommand{\ManuscriptSafeTProposedValueStrictRemoved}{240}
\newcommand{\ManuscriptSafeTProposedValueStrictRemovedJoint}{63}

%% file: sections/attainable_disparity_macros.tex
\newcommand{\ManuscriptAttainCountCells}{150}
\newcommand{\ManuscriptAttainCountFeasible}{150}
\newcommand{\ManuscriptAttainCountNoWorseOff}{146}
\newcommand{\ManuscriptAttainValueCells}{150}
\newcommand{\ManuscriptAttainValueFeasible}{150}
\newcommand{\ManuscriptAttainValueNoWorseOff}{150}
\newcommand{\ManuscriptAttainLossCells}{150}
\newcommand{\ManuscriptAttainLossFeasible}{150}
\newcommand{\ManuscriptAttainLossNoWorseOff}{90}
\newcommand{\ManuscriptAttainTotalCells}{450}
\newcommand{\ManuscriptAttainBindingShare}{68.5}

%% file: sections/abstract.tex
\begin{abstract}
Blocking a payment can prevent fraud loss and give up processing
revenue. If a platform caps the revenue it can give up, it should rank
payments by expected fraud loss per dollar of expected revenue.
We study how to learn this ranking, which of its blocks other accepted
rules would contest, and what information a screen needs to change
those decisions. We first motivate weighted binary cross-entropy
(W-BCE) as a practical way to learn this ranking with one score from
observed value and loss.

We call a block an objection when an accepted alternative rule would
pass the candidate at the same budget. We study how to reduce the
concentration of objections across groups while retaining loss capture.
Our contraction and replacement rule scales the blocking rate of each
contested candidate by a common factor and adds replacement blocks
among candidates whom all the criteria would block. Every group's
objection level scales by that factor in every measurement unit,
including for groups identified after fitting.

We show that the population-optimal W-BCE score recovers the yield
order. On nine of ten public datasets, W-BCE improves mean yield Gini
over ordinary binary cross-entropy. On ten public source families,
each repair fitted to a group objective raises some group's objection
level on held-out data. Common contraction keeps every group's level
at or below baseline, removes about half of objections in count, value,
and loss units, and retains $94.6\%$ of baseline loss capture on average.
\end{abstract}

%% file: sections/introduction.tex
\section{Introduction}
\label{sec:intro}

\begin{figure}[t]
\centering
\includegraphics[width=\columnwidth]{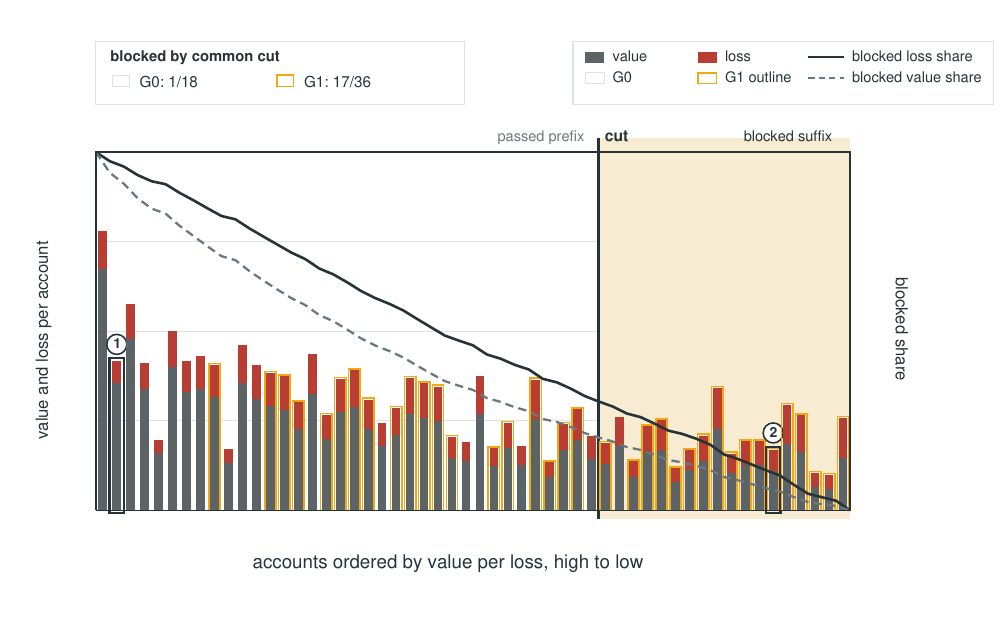}
\caption{Value-weighted yield ranking on realized accounts.  Each bar is an
account with realized value $a_i$ and loss $b_i$, and bars are sorted by
realized value per unit loss $a_i/b_i$, from high to low.  A vertical rank cut
passes accounts to its left and blocks accounts to its right;
equivalently, it blocks accounts at or below the cut value
on the realized value-per-loss scale.  Moving the cut from right to left blocks more accounts.
The curves show, for each possible cut, the blocked value and loss prevented
by the blocked suffix.  The group-1 orange outline shows how lower-value accounts can be
blocked more often under a common value-per-loss rule; the inset reports the
group block counts for the shown cut, with G0 and G1 denoting groups 0 and 1.
Markers 1 and 2 identify the similar-loss, different-value pair used in
Figure~\ref{fig:objection-pricing}.}
\Description{An illustrative account ranking ordered by value per unit loss, with curves for blocked value and captured loss and an inset comparing group block counts. Lower value can lead to more blocking despite similar loss. Markers 1 and 2 identify the example pair in the objection-penalty figure.}
\label{fig:greedy-overview}
\end{figure}

Risk screening and recommendation systems rank candidates whose outcomes
carry both value and loss. We study a binary pass/block decision, calling
each candidate an account. For account $i$, passing creates nonnegative
value $a_i$ and nonnegative cost or loss $b_i$; blocking prevents both.
In insurance, $a_i$ can be exposure and $b_i$ paid claims; in lending and
mortgages, they can be return or principal and credit loss; in web
recommendation, they can be engagement and negative feedback.
If these realized outcomes were known before the decision, a
value-budgeted rule would sort accounts by decreasing realized value per unit
loss, $a_i/b_i$, pass accounts to the left of a vertical cut, and block the
low-score accounts to its right.  Moving that cut left blocks more accounts
and plots blocked value against loss prevented, as in
Figure~\ref{fig:greedy-overview}.
We take $a_i/b_i=+\infty$ when $b_i=0<a_i$; candidates with
$a_i=b_i=0$ contribute no value or loss to this objective. Figure~\ref{fig:greedy-overview} labels groups 0 and 1 as G0 and G1.

In realistic systems, the ranking is over expected outcomes.  At decision
time, the system observes covariates $\mathbf X$ and chooses pass or block;
$A$ and $B$ are the random value and loss outcomes to be learned.  $A$ and $B$
may both be positive, and their conditional means may depend on $\mathbf X$;
when an account is blocked, the value and loss that passing it would have
produced are not observed. Let $\pi(\mathbf X)\in[0,1]$ be the probability of
blocking: $\pi=1$ blocks, $\pi=0$ passes, and intermediate values randomize
the decision. We call $E[\pi B]$ loss capture, the expected loss prevented.

For some business operating point, say
a blocked-value budget $\beta$, an upper bound on $E[\pi A]$, the population objective is
\begin{equation}
 \max_{\pi} E[\pi(\mathbf X)B]
 \quad\text{subject to}\quad
 E[\pi(\mathbf X)A]\le \beta .
 \label{eq:intro-objective}
\end{equation}
The standard budgeted-ranking ordering argument
\cite{neyman1933problem,dantzig1957discrete,clemencon2008ranking,kalashnov2021} says that a
solution ranks by the conditional value per unit loss,
\begin{equation}
 \Phi(\mathbf{x})
 =\frac{E[A\mid\mathbf X=\mathbf{x}]}
        {E[B\mid\mathbf X=\mathbf{x}]},
 \label{eq:intro-ratio}
\end{equation}
and blocks accounts with the lowest values of this value-per-loss ratio.
We set $\Phi=+\infty$ when mean loss is zero and mean value is positive;
when both means vanish, this objective is indifferent to the decision. This
ranking is followed by setting a threshold based on business considerations.

The ratio ranking supplies the unpenalized utility order, with ties allowing
multiple optimal policies. Our common controller reduces baseline blocking
on a fixed criterion-pass indicator and retains or adds blocks outside that
indicator along the learned order. Weighted binary cross-entropy (W-BCE)
estimates the order with one weighted pointwise classifier; a two-model
plug-in estimator can target the same order. The accepted indicators define
which blocks have objections, while the ratio orders the nonobjecting
candidates available for additions.

Our fairness question concerns conflicts between specified decision norms,
within the broader normative questions studied in algorithmic fairness
\cite{binns2018fairness}.
A value-per-loss rule can disagree with an accepted harm or net-harm rule at
the same blocked-value budget.  Because
value appears in the score, two accounts or groups with similar loss can be
ordered differently and blocked at different rates under the same
value-per-loss rule.  If harm or net-harm criteria are accepted as admissible
review criteria, then a block those criteria would have passed is a
contestable value-driven block, not merely a different outcome rate.
Figure~\ref{fig:greedy-overview} illustrates this effect with realized accounts
and the induced yield curve.  Ordinary pass/block, blocked-value, or
protected-loss parity checks do not ask whether the blocked accounts had such
objections. We weight objection removal against loss capture at the
design cap and constrain each contested candidate's block probability
relative to a fixed baseline.

\paragraph{Contributions.}
Our contributions are:
\begin{enumerate}[nosep,leftmargin=*]
 \item \textbf{Value-budgeted estimation and decision information.} Building
 on budgeted allocation and proper losses
 \cite{neyman1933problem,dantzig1957discrete,menon2018cost,savage1971,reid2010composite},
 W-BCE estimates the ratio-of-expectations order with one bounded score.
 We state the limitation of expected-label reductions and characterize
 outcome transformations that preserve this order. For population outcome
 coefficients, fixed objection indicators, and group normalizers,
 value- and loss-unit score adjustments
 use the ratio, while count-unit adjustments generally require absolute
 outcome scale (Proposition~\ref{prop:unit-score-information}).
 \item \textbf{Common-budget objections and group disparity.}  Fair
 classification and fair ranking commonly measure decision rates,
 label-conditioned rates, or exposure
 \cite{hardt2016equality,agarwal2018reductions,singh2018fairness,
 singh2026scoring}.  We instead start from account-level objections: when an
 accepted alternative criterion would pass a candidate at the same
 blocked-value budget, a value-per-loss block of that candidate is
 objectionable under that criterion.  We then define protected-group disparity
 by normalizing each group's realized objections by its own count, value, or
 loss. This explains why
 ordinary block, blocked-value, or protected-loss gaps can miss the complaint
 studied here.
 \item \textbf{Weighted objections and pointwise protection.} We optimize
 the trade-off between weighted objections and loss capture under a
 candidate-level bound on contested block rates. A retention band
 characterizes the corresponding bounds on proportional reduction across
 subgroups (Proposition~\ref{prop:retention-band}). Common contraction
 fixes one factor for every contested candidate and uses nonobjector
 backfill to recover capture.
 \item \textbf{Public-data evaluation.} The W-BCE score improves mean yield
 Gini on \ManuscriptRawGiniPositive{} of \ManuscriptRawGiniTotal{} datasets relative
 to ordinary BCE. With boosted predictors and two fixed criteria, common
 contraction removes about $48\%$ of objections in each of the three units
 and retains $94.6\%$ of baseline capture on average. Both operating
 targets hold in $57$ of $240$ held-out cohort-seed cells.
\end{enumerate}

\paragraph{Paper outline.}
In Section~\ref{sec:yield_model_form}, we analyze the value-per-loss target,
label-only limitations, and the W-BCE reduction that estimates it.
In Section~\ref{sec:remediable}, we define norm objections, their measurement
units, and the objection-penalized ranking rule.
In Section~\ref{sec:group-conditions}, we aggregate account-level objections into
protected-group disparity and analyze the weighted allocation,
the retention band, and the common contraction controller.
In Section~\ref{sec:experiments}, we compare W-BCE with ordinary BCE and
evaluate common contraction and backfill alongside signed, empirical-cut,
and parity controls.
In Section~\ref{sec:related}, we discuss related work, limitations, and
conclusions.

%% file: sections/yield_model_form.tex
\section{Yield-Model Form}
\label{sec:yield_model_form}

In Section~\ref{sec:intro}, we introduced the value-versus-loss ranking
problem and its value-per-loss target.  Here we justify W-BCE as an
operational way to estimate that order: W-BCE uses one weighted pointwise
classifier, trains on realized value and loss outcomes without preprocessing,
and avoids inference-time division by an estimated denominator.  In
Sections~\ref{sec:remediable} and~\ref{sec:group-conditions}, we reuse this
value-budgeted ranking form for account-level objections and protected-group
complaint disparities.

We use a simple single-output classifier trained on $n$ accounts
$(\mathbf{x}_i,a_i,b_i)$.  Observation $i$ uses soft label $q_i$ and weight
$w_i$,
\begin{equation}
 q_i=\frac{b_i}{a_i+b_i},\qquad w_i=a_i+b_i,
 \label{eq:model-form}
\end{equation}
where $a_i,b_i\geq0$ are realized value and loss.  Observations with
$a_i+b_i=0$ receive zero weight and are omitted from W-BCE training because
the metric is indifferent to them.
Thus $b_i=0<a_i$ gives label $0$, while $a_i=0<b_i$ gives label $1$.
Let $W=A+B$ and $Q=B/(A+B)$ when $W>0$, with $Q=0$ when $W=0$.  The classifier emits
$\hat y_\theta(\mathbf{x})=\sigma(s_\theta(\mathbf{x}))\in(0,1)$, where
$\sigma$ is the logistic sigmoid, $s_\theta$ is the learned logit, and
$\theta$ denotes model parameters. The empirical W-BCE loss is
\begin{equation}
 \widehat{\mathcal L}_{\rm W\text{-}BCE}
 =-\frac1n\sum_{i=1}^n
 \left[b_i\log\hat y_\theta(\mathbf{x}_i)
 +a_i\log\{1-\hat y_\theta(\mathbf{x}_i)\}\right].
 \label{eq:wbce-general}
\end{equation}
At inference the system ranks by $\hat y_\theta(\mathbf{x})$ and applies the
selected threshold.

\paragraph{Yield-order estimation necessity.}
For any blocked-value budget $\beta\in[0,E[A]]$, the classical
fractional-knapsack and Neyman--Pearson ordering argument
\cite{neyman1933problem,dantzig1957discrete,clemencon2008ranking,kalashnov2021} shows that
an optimizer of Eq.~\eqref{eq:intro-objective} blocks accounts with the lowest
values of the value-per-loss ratio in Eq.~\eqref{eq:intro-ratio}, with
randomization on ties if necessary.

We can consider several ways to estimate this order.  W-BCE does not use a
ranking loss, but it is known from weighted proper-loss theory
\cite{savage1971,reid2010composite} that its label-plus-weight construction in
Eq.~\eqref{eq:model-form} elicits a score monotone decreasing in $\Phi$, as
stated in Theorem~\ref{thm:wbce}.
We also consider alternative \textit{label-only} transformations that assign
each account one scalar relevance label and then optimize a pointwise,
pairwise, or listwise surrogate \cite{cao2007learning,xia2008listwise}.
Our impossibility claim concerns methods whose optimal order is a monotone
function of the conditional mean of that scalar label, as defined below.

\begin{theorem}[Weighted proper-loss consistency]
\label{thm:wbce}
Optimize predictions over $[0,1]$ with proper boundary extensions.
For any strictly proper binary loss $\ell$, at every covariate value with
$0<E[A+B\mid\mathbf X=\mathbf{x}]<\infty$, the population minimizer of
\[
 E[W\{Q\ell(1,\hat y)+(1-Q)\ell(0,\hat y)\}\mid\mathbf X=\mathbf{x}]
\]
is
\[
 \hat y^*(\mathbf{x})
 =\frac{E[B\mid\mathbf X=\mathbf{x}]}
        {E[A+B\mid\mathbf X=\mathbf{x}]}
 =\frac{1}{1+\Phi(\mathbf{x})}.
\]
Here the last expression equals one when $\Phi(\mathbf{x})=0$.
\end{theorem}

Consequently, descending classifier scores induce the same blocking order as
ascending $\Phi$; where the conditional total stake is zero, the loss and
decision are both indifferent. Finite-logit BCE approaches endpoint optima
as infima; Appendix~\ref{app:regularity} details these boundary cases.

\begin{proof}
Conditioning on $\mathbf X=\mathbf{x}$, the identities $WQ=B$ and
$W(1-Q)=A$ reduce the risk to
$E[A+B\mid\mathbf{x}]\{\bar q\ell(1,\hat y)+(1-\bar q)\ell(0,\hat y)\}$,
where $\bar q=E[B\mid\mathbf{x}]/E[A+B\mid\mathbf{x}]$.
Strict properness gives the unique optimum $\bar q=1/(1+\Phi)$,
which decreases strictly with $\Phi$.
\end{proof}

Define a method as \emph{label-only} when its population-optimal order is a
strictly monotone transform of $E[g(B,A)\mid\mathbf X=\mathbf{x}]$ for one
scalar label transformation $g$.  Such methods do not rank correctly in
general.

\begin{proposition}[Label-only limitation]
\label{prop:fixed-label-limitation}
No label-only reduction recovers the yield order universally.
Specifically, there do not exist a measurable scalar label $g(B,A)$ and a
fixed strictly monotone function $H$ such that
\[
 E[g(B,A)\mid\mathbf X=\mathbf{x}]
 =H\!\left(
   \frac{E[A\mid\mathbf X=\mathbf{x}]}
        {E[B\mid\mathbf X=\mathbf{x}]}
 \right)
\]
for every conditional distribution of $(B,A)$ with finite means and positive
mean loss.
\end{proposition}

\begin{proof}
Consider a degenerate distribution concentrated at $(B_0,A_0)$ with $B_0>0$.
The assumed identity requires $g(B_0,A_0)=H(A_0/B_0)$, so any universally
valid scalar label must depend on an observation only through $A/B$.
We write $\delta_{(b,a)}$ for the distribution assigning probability one
to $(B,A)=(b,a)$. Now let
$P_1=\tfrac12\delta_{(1,1)}+\tfrac12\delta_{(2,1)}$ and
$P_2=\tfrac12\delta_{(1,1)}+\tfrac12\delta_{(6,3)}$.
Under both distributions, $A/B$ equals $1$ or $1/2$ with equal probability, so
$E_{P_1}[g]=E_{P_2}[g]$.  Yet
$E_{P_1}[A]/E_{P_1}[B]=2/3$ while $E_{P_2}[A]/E_{P_2}[B]=4/7$.
Strict monotonicity of $H$ requires different expected labels, giving a
contradiction.
\end{proof}

This explains common failures: predicting $B$ ignores blocked value;
observed-ratio regression targets $E[A/B\mid\mathbf{x}]$ and depends on a
zero-denominator convention; unweighted regression on $B/(A+B)$ learns another
expectation of a ratio.
When $B>0$ almost surely and the inverse moments exist, observed-ratio
regression has the exact discrepancy, where $\operatorname{Cov}$ denotes conditional covariance,
\begin{align}
 E[A/B\mid\mathbf{x}]-\frac{E[A\mid\mathbf{x}]}{E[B\mid\mathbf{x}]}
 &=E[A\mid\mathbf{x}]
 \left(E[1/B\mid\mathbf{x}]-\frac1{E[B\mid\mathbf{x}]}\right) \nonumber\\
 &\quad+\operatorname{Cov}(A,1/B\mid\mathbf{x}),
 \label{eq:ratio-bias}
\end{align}
so the Jensen term is driven by denominator variation, with a covariance term
that can reinforce or offset it.

The instance weight is what turns the fractional label into the ratio of
conditional expectations.  Without the weight, fractional regression learns
\[
 q_{\rm frac}(\mathbf{x})=E[Q\mid\mathbf X=\mathbf{x}],
\]
an expectation of an observed fraction.  W-BCE instead learns
\begin{align*}
 q_{\rm W}(\mathbf{x})
 &=\frac{E[WQ\mid\mathbf{x}]}{E[W\mid\mathbf{x}]}
   =\frac{E[B\mid\mathbf{x}]}{E[B+A\mid\mathbf{x}]}\\
 &=\frac{1}{1+\Phi(\mathbf{x})},
\end{align*}
which preserves the yield order.

\paragraph{Training and serving simplicity.}
The preceding results rule out label-only reductions, so the operational
contrast is between explicit rankers and W-BCE.  An explicit ranker must
construct scores or comparisons from the value-per-loss target.  A plug-in
ranker does this by estimating $E[B\mid\mathbf{x}]$ and
$E[A\mid\mathbf{x}]$ separately and dividing at inference; pairwise or
listwise rankers must likewise encode the target order in their training
comparisons.  W-BCE recovers the same population order with one weighted
pointwise classifier, avoids inference-time division by an estimated
denominator, and gives thresholds the population interpretation
\[
 \hat y^*(\mathbf{x})=\frac{1}{1+\Phi(\mathbf{x})}
 \quad\Longleftrightarrow\quad
 \Phi(\mathbf{x})=\frac{1-\hat y^*(\mathbf{x})}{\hat y^*(\mathbf{x})}.
\]

\paragraph{Preprocessing effects on ranking.}
Ratio-based implementations, including plain W-BCE, raise preprocessing
questions.  In finite samples, we may clip, cap, or damp the soft
label, the weight, or the implied numerator and denominator.  The next
proposition, proved in Appendix~\ref{app:preprocessing-ranking}, shows that
preprocessing makes W-BCE optimize a transformed numerator and denominator.
Only a narrow linear class preserves the value-per-loss order for every
distribution.

\begin{proposition}[Preprocessing changes the ranking target]
\label{prop:preprocessing-ranking}
Let $Q'\in[0,1]$ and $W'>0$ be finite measurable transformations on the
full outcome cone $A,B\ge0$, $A+B>0$. Assume finite conditional
transformed means and positive transformed denominator. After replacing
$(Q,W)$ by $(Q',W')$, W-BCE ranks by
\[
 \frac{E[W'Q'\mid\mathbf X=\mathbf{x}]}
      {E[W'(1-Q')\mid\mathbf X=\mathbf{x}]}.
\]
On all finite-support distributions on that cone with positive outcome
means and positive transformed denominator, this order equals a fixed
strictly decreasing function of $\Phi$ for every such distribution
if and only if the transformed outcome sides satisfy
\[
 W'Q'=p_1B+p_2A,\qquad W'(1-Q')=r_1B+r_2A,
\]
for fixed nonnegative constants $p_1,p_2,r_1,r_2$ with
$p_1r_2-p_2r_1>0$.
In particular, nonlinear weight damping is not equivalent to raw W-BCE.
\end{proposition}

\begin{proof}[Proof sketch]
Put $U=W'Q'$ and $V=W'(1-Q')$, and let $H$ be the fixed strictly
decreasing ratio function in the hypothesis. For each $t>0$, two-point mixtures with
$E[A-tB]=0$ must also have $E[U-H(t)V]=0$. Varying mixture endpoints gives
$U-H(t)V=c_t(A-tB)$ pointwise for some scalar $c_t$. Two distinct values of $t$ express $U,V$
linearly in $A,B$. Nonnegativity on the outcome cone forces nonnegative
coefficients, and strict decrease of $H$ forces the positive determinant.
Conversely, the ratio is $(p_1+p_2\Phi)/(r_1+r_2\Phi)$, whose derivative has
the required negative sign. Appendix~\ref{app:preprocessing-ranking} gives the full argument.
\end{proof}

Thus clipping, logarithms,
square roots, and other nonlinear damping are finite-sample bias--robustness
choices, not equivalent implementations of W-BCE.

%% file: sections/remediable_adverse_impact.tex
\section{Account-Level Objections}
\label{sec:remediable}

\begin{figure}[t]
\centering
\includegraphics[width=\columnwidth]{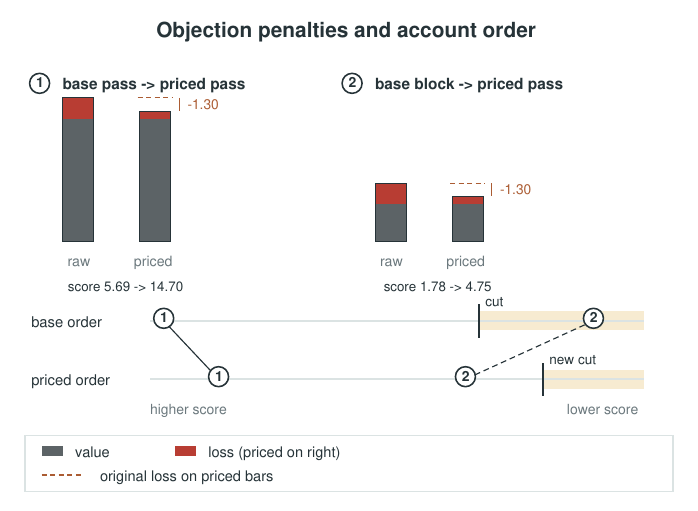}
\caption{Objection penalties can move a candidate across the blocking
cut. The candidates are from Figure~\ref{fig:greedy-overview}.
Subtracting a penalty from expected loss moves candidate 2 from block
to pass. Dashed marks show the loss before the penalty.}
\Description{Two panels show the original and adjusted order of the illustrated candidates. Subtracting an objection penalty changes candidate 2 from block to pass.}
\label{fig:objection-pricing}
\end{figure}

Two candidates may carry similar expected losses. The yield rule can
block the candidate with less expected value first. A rule that ranks by
harm alone may pass that candidate at the same budget. Figure~\ref{fig:objection-pricing}
shows how charging for these disagreements changes the blocking order.

\paragraph{Yield curves and criterion-based adverse impact.}
Write $a(x)=E[A\mid X=x]$ and $b(x)=E[B\mid X=x]$. We use expected harm
$b(x)$ and net harm $b(x)-\lambda_0 a(x)$ as reference scores, with
$\lambda_0$ converting value units to loss units. The empirical settings
below use $\lambda_0=1$ in their stated units, except for the video task,
where $\lambda_0\simeq1.6424\times10^{-4}$ loss units per second of play
time. A third possible score is the probability of positive loss,
$\Pr(B>0\mid X=x)$. Expected harm weights this probability by expected
severity conditional on a positive loss wherever that probability is
positive. For example, when
$A=z(1-Y)$ and $B=zY$ for a binary loss label $Y$ and a positive amount
$z$ known at decision time, the loss odds equal $b(x)/a(x)$.
The probability and yield orders then coincide.

For each accepted criterion $c\in\mathcal C$, let $\pi_c^\beta(x)$
denote its reference policy, which blocks high scores under the same
blocked-value budget $\beta$. Freeze these policies and their tie rules
before adjusting the yield policy. The reference scores are inputs to the
analysis. A disagreement identifies a block that one of those rules
would contest \cite{hardt2016equality,menon2018cost,agarwal2018reductions}.

\paragraph{Pairwise disagreement.}
Let $q_c(x)=1-\pi_c^\beta(x)$ be the probability that criterion $c$
passes a candidate. Its objection to policy $\pi$ is
\begin{equation}
 O_c(x;\pi)=q_c(x)\pi(x).
 \label{eq:criterion-objection}
\end{equation}
For fixed binary reference decisions, this is the rate at which $\pi$
blocks a candidate whom $c$ passes. With randomized reference decisions
it is the probability of that disagreement under independent draws.
The union indicator $q_{\mathcal C}(x)=\mathbf1(\exists c\in\mathcal C:
q_c(x)>0)$ marks any candidate whom at least one criterion can pass.
The union objection rate $O_{\mathcal C}(x;\pi)=q_{\mathcal C}(x)\pi(x)$
counts each contested block once. The boosted evaluation uses binary
harm and net-harm reference decisions.

Let $U_1=1$, $U_A=A$, and $U_B=B$. Multiplying an objection by $U_w$
measures it in count, value, or loss units. A block can contribute to
several criterion-specific totals when several criteria contest it.

\paragraph{Objection-penalized ranking.}
Write $C(\pi)=E[\pi B]$ for loss capture and $C_0=C(\pi_0)>0$ for
baseline capture. For fixed positive normalizers $N_c^w$ and nonnegative
weights $\omega_c^w$, put
\[
 W(\pi)=\sum_{c,w}\omega_c^w
       \frac{E[\pi q_c U_w]}{N_c^w}.
\]
For a fixed $\eta\geq0$, we minimize over $\pi$
\begin{equation}
 1-\frac{C(\pi)}{C_0}+\eta W(\pi),
 \qquad E[\pi A]\leq\beta.
 \label{eq:weighted-objection-objective}
\end{equation}
The weights trade among criteria and units. The factor $\eta$ trades
their combined objection cost against capture. For per-criterion
weights, a candidate contested by two criteria contributes to both
terms. Using $q_{\mathcal C}$ in place of the individual flags gives a
single union term per unit. Positive baseline masses provide one choice
of normalizers. A term omitted for zero baseline mass can still receive
pointwise protection.

With $u_w(x)=E[U_w\mid X=x]$, define
$m(x)=\eta C_0\sum_{c,w}\omega_c^w q_c(x)u_w(x)/N_c^w$.
Minimizing Eq.~\eqref{eq:weighted-objection-objective} is equivalent to
maximizing $E[\pi(B-m)]$ under the same budget. With unrestricted block
rates, this is objection-penalized ranking. A budget multiplier $\mu$
subtracts $\mu a(x)$ from the adjusted loss $b(x)-m(x)$, so the rule
blocks where the resulting coefficient is positive and passes where it
is negative. The next section uses the same coefficient to choose
within bounds on each contested block rate.

\paragraph{Algorithm.}
We fit the yield and criterion scores on training rows, then freeze the
criteria and candidate score cuts before policy calibration. The price
form of Eq.~\eqref{eq:weighted-objection-objective} varies the trade-off
at a fixed blocked-value budget. A capture-floor form instead minimizes
$W(\pi)$ subject to $C(\pi)\geq\rho C_0$ and the same budget. This
second form lets us compare policies at a specified capture requirement.
For the historical unrestricted penalty comparison, Appendix~\ref{app:objection-alg}
gives the fitted scores and validation procedure.

%% file: sections/group_fairness_conditions.tex
\section{Group-Level Disparity}
\label{sec:group-conditions}

A smaller difference between groups can come from more objections in the
group with fewer. We study which blocks a policy can release while ensuring
that every group's objection level falls or stays at its baseline level.

\paragraph{Group complaint disparity.}
\label{subsec:group_disparity}
Let $G$ be a group variable. A policy may use $G$ along with its other
features. Using the reference decisions from Section~\ref{sec:remediable},
write the union objection rate as
\begin{equation}
 O_{\mathcal C}(x,g;\pi)=\pi(x,g)q_{\mathcal C}(x).
 \label{eq:group-account-objection}
\end{equation}
For a group $g$ with a positive finite denominator, its level in unit $w$ is
\begin{equation}
 D_g^w(\pi)=
 \frac{E[O_{\mathcal C}(X,G;\pi)U_w\mid G=g]}
      {E[U_w\mid G=g]} .
 \label{eq:group-complaint-level}
\end{equation}
The denominator uses all of the group's count, value, or loss. For two
groups, the signed gap is
\begin{equation}
 \Delta_D^w(\pi)=D_1^w(\pi)-D_0^w(\pi).
 \label{eq:group-complaint-disparity}
\end{equation}
A penalty on the absolute gap rewards a rise in the lower level when
that rise brings the levels closer. A cap on each level can prevent
this for groups and units named in advance. The pointwise bound below
also covers groups named after fitting.

\paragraph{Complaint-disparity control.}
\label{subsec:group_control}
\input{sections/weighted_release}

\paragraph{Relation to outcome parity.}
Ordinary outcome rates average $\pi$ in each group. Objection levels
average $q_{\mathcal C}\pi$. The level factors into the group's share of
the measurement unit on contested candidates and its average block rate
within that share. Groups can therefore have the same rate among
contested candidates and different objection levels. Appendix~\ref{app:group-disparity-decomposition}
gives the factorization, illustrates the denominator in
Figure~\ref{fig:group-denominator-choice}, and derives the signed gap
penalty used in the comparison. Appendix~\ref{app:regularity} gives its
feasibility conditions.

%% file: sections/weighted_release.tex
\textsc{Pointwise protection.}
Write $q$ for the union of the criteria the policy protects and fix a
baseline $\pi_0$. Our experiments use $q=q_{\mathcal C}$.
On contested candidates, we require $q\pi\leq q\pi_0$. This leaves
nonobjector rates free. The common controller below keeps the
baseline nonobjector blocks and adds replacements where there is room.
A criterion can appear in the objective without being protected.
Protection applies to every $q_c$ whose support lies inside $q$.

\smallskip
\noindent\textsc{Proportional release.}
For $0\leq s_-\leq s_+\leq1$, require
\begin{equation}
 s_-q\pi_0\leq q\pi\leq s_+q\pi_0.
 \label{eq:objection-retention-band}
\end{equation}
The interval $[0,1]$ allows any release consistent with the pointwise
bound. Equal endpoints give a common contraction factor.

\begin{proposition}[Retention bands and subgroup levels]
\label{prop:retention-band}
Equation~\eqref{eq:objection-retention-band} holds almost surely if and
only if every measurable subgroup's union objection level stays
between $s_-$ and $s_+$ times its baseline level in every nonnegative
unit with a positive finite subgroup denominator. The same bounds
then hold for each constituent pass probability supported on $q$.
\end{proposition}
\begin{proof}
Multiply the pointwise inequalities by the unit and condition on
the subgroup. For a constituent criterion, first multiply by its
pass probability. For the converse, take the count unit on the set
where either inequality fails. A set of positive probability violates
the corresponding subgroup bound.
\end{proof}

Thus the pointwise cap characterizes protection of all subgroup levels.
For baseline levels $d_1$ and $d_0$, a band bounds the new absolute
gap by $s_+|d_1-d_0|+(s_+-s_-)\min(d_1,d_0)$. This follows by checking
both signed differences at the interval endpoints.
A common factor $s$ multiplies every gap by $s$ and preserves the
ratio of any two positive levels when $s>0$.

Selective release can widen a gap even when the baseline already
maximizes capture. Consider two equally weighted candidates per
group. Each group has one contested baseline block of value $1$.
These blocks capture losses $3$ in group 1 and $1$ in group 0. The
other candidates carry no loss, and total value is $2$ in group 1 and
$4$ in group 0. The value objection levels start at $(1/2,1/4)$.
To retain half of baseline capture with the least contested value,
keep two thirds of the first block and release all of the second.
The levels become $(1/3,0)$ and the gap grows from $1/4$ to $1/3$.
Common contraction at the same capture gives $(1/4,1/8)$ and a gap
of $1/8$.

\input{sections/objection_safe_method}

\smallskip
\noindent\textsc{Fitting weighted release.}
For a fixed band, write $\ell(x)$ and $u(x)$ for the allowed lower and
upper rates. The blocked-value budget admits a policy exactly when
$E[\ell A]\leq\beta$. Start at $\ell$. Among candidates with positive
expected value, allocate the remaining block probability in order of
$(b-m)/a$, stopping at the budget or when the adjusted loss becomes
nonpositive. Zero-value candidates with positive adjusted loss take
their upper rate. This is the fractional knapsack rule for a finite
population \cite{dantzig1957discrete}. In the population formulation, when
a finite budget multiplier $\mu\geq0$ attains the dual optimum, the rule
takes $u$ if $b-m-\mu a>0$ and $\ell$ if $b-m-\mu a<0$. At equality,
randomization can use the remaining budget. If $\mu>0$, the assigned
rates spend exactly $\beta$. A rule that uses group membership
conditions all the means on that information as well.

Let $\theta_w(x)=\eta C_0\sum_c\omega_c^w q_c(x)/N_c^w$.
For count, value, and loss penalties together, the coefficient is
\[
 (1-\theta_B)b-(\mu+\theta_A)a-\theta_1 .
\]
Binary reference flags give the same $\theta_w$ to candidates with
the same pattern of objections. With no count weight and
$1-\theta_B>0$, the rule uses one cut on $b/a$ within that pattern.
A count penalty also depends on the scale of value or loss. It can
change which candidates within a pattern are released. These weights
can move release across patterns while the pointwise guarantee
continues to hold. With a common factor, all protected terms instead
fall by the same fraction. Weights can change the chosen factor
through the total penalty, and the capture-floor optimum within this
class is the smallest feasible common factor when the objective has
positive protected mass and all priced terms lie inside the union.

Appendix~\ref{app:objection-safe-details} gives a converse for common
gap contraction and bounds the costs of protection and uniformity.
On new candidates, the guarantee follows from the frozen bounds.
Retained capture and blocked value remain empirical outcomes.

%% file: sections/objection_safe_method.tex
\smallskip
\noindent\textsc{Contraction and replacement.}
The common controller releases the same fraction of each contested
block and replaces some of the forfeited capture with nonobjector
blocks. With contraction factor $s$ and replacement fraction $h(x)$,
both in $[0,1]$, its rate is
\begin{equation}
 \pi_{s,h}
 =sq\pi_0+(1-q)\bigl[\pi_0+(1-\pi_0)h\bigr].
 \label{eq:objection-safe-policy}
\end{equation}
The term $1-\pi_0$ is the blocking probability still available for
replacement. When the baseline and criterion scores use no group
attribute, this rule needs none at decision time.

Each replacement cut draws from nonobjectors above a frozen yield
score threshold. We mix cuts to reach the chosen replacement cost within the
blocked-value budget. Fitting the mixture and $s$ under the capture floor is a
linear program. A mixture of at most two cuts on the value-capture
frontier suffices at the chosen replacement cost. When the floor
cannot be met, the fitted rule uses the best capture within its
budget-feasible class. If even the retained nonobjector blocks
exceed the budget, the class is infeasible.

Let $B_q=E[q\pi_0B]$ and let $H_B=E[(1-q)(1-\pi_0)B]$ be the
capture available for replacement before applying the budget.
The common rule gives up $(1-s)B_q$ and recovers at most $H_B$.
A capture floor of $\rho C_0$ therefore requires
\[
 1-s\leq\min\left\{1,\frac{(1-\rho)C_0+H_B}{B_q}\right\},
 \qquad B_q>0.
\]
A new protected criterion can reduce $H_B$ by contesting blocks
previously available for replacement. It can also make some
baseline blocks eligible for release. Both effects matter when
comparing fits under different criterion sets.

%% file: sections/experiments.tex
\section{Experiments}
\label{sec:experiments}

We evaluate estimation of the value-budgeted utility order, measurement of
criterion objections, and complaint control under shared value budgets and
capture requirements.

\subsection{Experimental Settings}
\label{subsec:exp-data}

\paragraph{Score comparison.}
We compare fixed raw W-BCE with ordinary BCE on ten public recommendation,
insurance, credit, and mortgage datasets. Value and loss represent engagement
and negative feedback, exposure and claims, or lending value and credit loss.
For cumulative loss capture $c(v)$ at blocked-value fraction $v$, yield Gini is
$2\int_0^1c(v)\,dv-1$ \cite{lorenz1905methods}; a yield Gini of zero is the
value of ordering at chance. Figure~\ref{fig:score-bce-wbce}
reports paired differences in yield Gini and capture at a 10\% value budget,
using ten splits.

\input{sections/wbce_score_figure}

Yield Gini summarizes the whole curve and needs no threshold. Capture is
measured at an operating point, so its protocol matters: four datasets
(Australian MTPL, Fannie Mae, Freddie Mac, KuaiRand) use a validation-frozen
threshold [V], while the remaining six interpolate the threshold on the test
curve [T] and are therefore a ranking measure rather than an attainable
operating point. Positive paired means occur on \ManuscriptRawGiniPositive{} of
\ManuscriptRawGiniTotal{} datasets for yield Gini and \ManuscriptRawCapturePositive{}
of \ManuscriptRawCaptureTotal{} for capture, but only the [V] datasets speak to a
reachable operating point, and among those the paired interval excludes zero
on Australian MTPL and Fannie Mae.

Ordinary BCE targets a different functional. Its negative yield Gini on
Australian MTPL illustrates that loss-probability ranking can reverse the
value-weighted order. W-BCE and a two-model plug-in estimator target the
same population ratio. On the heteroskedastic construction, the evaluated
observed-ratio and pairwise baselines perform worse
(Appendix~\ref{app:exp-protocol}). The operational form in
Section~\ref{sec:yield_model_form} estimates this ratio with one bounded
classifier score and no division by separately fitted outcome means.

\paragraph{Fairness comparison.}
The fairness panel contains ten public source families. Pairing a
dataset with a group definition gives $24$ settings. Ten seeds per
setting give $240$ cells. We fit boosted trees for the W-BCE score
and separate conditional means for harm and net harm. The blocked-value
budget is $10\%$ of total value and the capture floor is $95\%$ of
baseline capture. Earlier linear results appear in
Appendix~\ref{app:objection-safe-details}.

\subsection{Account-Level Frontiers}
\label{subsec:exp-account}

A blocked candidate can object when a harm or net-harm criterion would pass
it at the same value budget (Eq.~\eqref{eq:criterion-objection}).
Figure~\ref{fig:account-objection-pareto} and
Table~\ref{tab:objection-penalized} in
Appendix~\ref{app:score-account-results} retain the descriptive account
frontier and its endpoints. Those bank summaries supply context for the
selected-policy comparisons below.

\subsection{Objection Removal and Capture}
\label{subsec:exp-group}

\paragraph{Protocol.}
Training fits the predictors. Validation fixes the baseline, criterion
thresholds and replacement cuts. Calibration fits the common
contraction factor and replacement mixture. The final test split
$T$ fits nothing. The four policies in Table~\ref{tab:boosted-contraction}
use the same held-out rows. The two group-quota policies fit
group-specific release cuts, with and without a reserve on replacement.
The signed group penalty, shared penalty, and mixture of shared
penalties were selected on validation and evaluated separately on
the original test split. The final $T$ comparison is a continuation
on previously examined data.

\paragraph{Results.}
\textsc{Gap reduction and group levels.}
On the original test split, the signed group penalty raises a group's
value objection level in $202$ of $240$ cells. Of the $144$ cells in
which it narrows the gap, $137$ have a group whose level rises. The
shared penalty and its mixture raise a group's level in $22$ and $14$
cells. Common contraction raises neither group's level in any of the
$240$ cells of its final test split. Its guarantee follows from the
pointwise bound, including for groups formed after calibration.

\input{sections/boosted_contraction_table}

\noindent\textsc{Removal and capture.}
On $T$, common contraction removes about $48\%$ of objections in
each unit and retains $94.6\%$ of baseline capture on average.
Thinning removes $5\%$ and retains $95\%$ by construction.
The joint operating targets hold in $57$ cells for common contraction
and $183$ for thinning. Common contraction misses the capture floor
in $136$ cells. Group-quota cuts remove more value-weighted
objections on average and retain less capture.

\noindent\textsc{Proportionality.}
The three removal fractions agree to numerical precision in every
one of the $237$ cells in which all three are defined for common
contraction. Count, value, and loss weight different candidates, so
this agreement follows from the common factor. The group-quota
policies preserve nonincrease while allowing different removal
fractions. The benchmark evaluates common contraction and these
group-quota policies. The retention band gives a way to impose a
specified bound on those differences in future fits.

%% file: sections/wbce_score_figure.tex
\begin{figure}[t]
\centering
\includegraphics[width=\columnwidth]{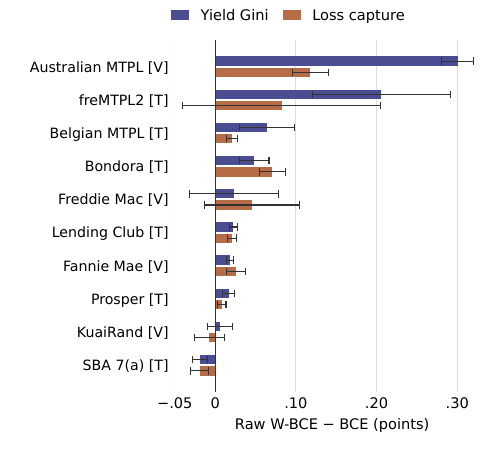}
\caption{Raw W-BCE minus BCE on ten datasets. Each pair shows yield-Gini
and loss-capture deltas; whiskers are 95\% paired Student $t$ intervals over
ten splits. Positive values favor W-BCE. Dataset suffixes describe capture:
[V] validation-frozen threshold; [T] retrospective test-curve interpolation.}
\Description{A grouped horizontal bar chart pairs yield-Gini and loss-capture differences for each of ten datasets. Both metrics retain their paired confidence intervals and use the same dataset order. Capture protocol labels distinguish validation-frozen thresholds from retrospective test-curve interpolation.}
\label{fig:score-bce-wbce}
\end{figure}

%% file: sections/boosted_contraction_table.tex
\begin{table}[t]
\centering
\small
\setlength{\tabcolsep}{3pt}
\begin{tabular}{@{}lrrrrr@{}}
\toprule
& & \multicolumn{3}{c}{Objections removed} & \\
Policy & Capture & Count & Value & Loss & Both targets \\
\midrule
Thinning & 95.00 & 5.00 & 5.00 & 5.00 & 183 \\
Common contraction & 94.62 & 48.03 & 47.59 & 48.12 & 57 \\
Group quota & 93.34 & 60.72 & 64.83 & 59.98 & 48 \\
Group quota with reserve & 91.33 & 60.72 & 64.83 & 59.98 & 49 \\
\bottomrule
\end{tabular}
\caption{Boosted-tree policies on the final test rows. Capture and removal
are mean percentages over cells with defined baseline ratios. Capture
and count removal use 240 cells, value removal uses 238, and loss removal
uses 239. The final column counts cells that reach both the 95\% capture
floor and 10\% blocked-value cap.}
\label{tab:boosted-contraction}
\end{table}

%% file: sections/discussion_limitations_and_conclusion.tex
\section{Related Work, Limitations, and Conclusion}
\label{sec:related}

\paragraph{Related work.}
Binns connects algorithmic fairness to political philosophy
\cite{binns2018fairness}; Lyons et al. study contestability
\cite{lyons2021contestability}. De Vos et al. constrain demographic parity
within relevant decision regions \cite{devos2025decision}, while H{\"o}ppner
et al. derive instance-dependent cost-sensitive fraud decisions
\cite{hoppner2020instance}. These works motivate normative, decision, and
cost-sensitive perspectives. Our target is disagreement between yield and
accepted review policies at a common blocked-value budget, aggregated into
group complaints and controlled through block probabilities. Fair-classification
thresholds and reductions \cite{hardt2016equality,kleinberg2017inherent,
chouldechova2017fair,corbettdavies2017algorithmic,menon2018cost,
agarwal2018reductions} and fair-ranking trade-offs
\cite{singh2018fairness,singh2026scoring} supply established tools.
Singh et al. show that individual scoring can miss optimal full-ranking
trade-offs when fairness depends on the query composition
\cite{singh2026scoring}. Our binary allocation uses fixed linear budget and
complaint constraints. Our common controller contracts objecting
block probabilities and fits additions along frozen score cuts. Its empirical
optimization covers this anchored class. The value-versus-loss objective and
objection units determine the ratio target and the additional information
required for resource and capture control.
Budgeted allocation, ratio ranking, and proper losses provide the learning
and optimization foundations \cite{neyman1933problem,dantzig1957discrete,
clemencon2008ranking,kalashnov2021,qiu2022treatment,lambert2008,savage1971,
reid2010composite,gneiting2011point,frongillo2015vector}; logged ranking and
partial identification inform observation limits
\cite{joachims2017unbiased,manski1990nonparametric}.

\paragraph{Normative scope.}
We take the admissible review criteria and their fixed pass indicators as
inputs. Disagreement identifies an objection under a specified criterion,
without establishing which criterion should govern.

\paragraph{Policy class.}
The policy class is binary pass/block under a value budget. Position-dependent
ranking or intervention intensity requires a richer action space. The primary
optimization covers common contraction and add-only score cuts; retaining
baseline nonobjecting blocks can make a tighter value budget infeasible.

\paragraph{Outcome proxies.}
The measured outcomes are proxies. Insurance exposure measures coverage rather
than profit, and paid claims require causal assumptions to represent
preventable loss. KuaiRand hate feedback measures observed responses, and its
historical batch-size override cannot be verified because the original run
arguments were not retained. CMS uses a provider potential-fraud flag, and
UCI and German credit convert binary labels into amount-weighted loss proxies;
none of these measure confirmed or preventable cash loss. Selective credit
outcomes also limit identification
\cite{joachims2017unbiased,manski1990nonparametric,fawkes2024fragility}.
These constructions also assign an amount to one side only, so value and
loss are disjoint across candidates in nine of the ten datasets and in
$99.9\%$ of KuaiRand. Two consequences follow for scope. These benchmarks mainly exercise W-BCE with hard labels and amount
weights. And a measurement unit selects which
candidates a policy can act on at all rather than only how its objections
are weighed, so unit-specific findings may not transfer to settings where a
candidate realizes value and loss together.

\paragraph{Scope of the guarantees.}
For fixed $q$, $\pi_0$, $0\leq s\leq1$, and unchanged nonnegative outcome weights,
$q\pi=sq\pi_0$ contracts total and group objections in every unit;
normalized levels require positive finite denominators. This guarantee also
covers each constituent criterion indicator contained in $q$. A shared
randomization couples realized objecting blocks as a subset of baseline
blocks, but exact gap scaling concerns probabilities and expectations.
The objection identity alone does not impose a hard value budget on unknown
new stakes or guarantee a fitted capture floor. The finite-cut calibration
bound requires independent identically distributed calibration data and
known domain bounds; the current unbuffered experiments do not apply it.

\paragraph{Scope of the empirical claims.}
The boosted benchmark uses harm and net harm and evaluates common
contraction and group-quota cuts. The loss-occurrence criterion and
general weighted bands specify additional policies for evaluation.
Capture and value targets can fail when calibration is reapplied
to new candidates. The retrospective public outcomes are proxies
for the loss and value a served system would realize.

\paragraph{Group measurement.}
Binary sex or gender fields cannot establish causal discrimination or
represent intersectional identities.

\paragraph{Conclusion.}
Weighted binary cross-entropy estimates the yield order with one score.
A pointwise bound on contested rates lets us optimize weighted objections
and loss capture while preserving objection nonincrease in any subgroup
under the fixed criteria. A retention band controls how differently
those groups can improve. Common contraction gives every subgroup the
same proportional reduction, and the boosted-tree evaluation shows
substantial objection removal with uneven capture and blocked-value
target attainment.

%% file: sections/ethical_considerations.tex
\section*{Ethical Considerations}

Errors in predicted value or loss can wrongly block accounts, and changes
in blocking can shift harm between groups. The common controller applies
the same factor and pooled backfill cuts across groups. Its upstream score
or criterion indicator may use group membership; when they do, that use
requires justified criteria, privacy controls, subgroup review, and appeal
paths for consequential blocks. Objection reductions under fixed criteria
and outcome proxies establish neither causal harm reduction nor legal
compliance. Our
experiments use public retrospective data, with no production use and no new
personal-data collection.

%% file: sections/experiment_protocol_appendix.tex
\section{Experiment Protocol Details}
\label{app:exp-protocol}

Public score-learning and candidate-log runners, with input and dependency
instructions, are available at \url{https://github.com/atavory/yielding_fairness}.
The \href{https://github.com/atavory/yielding_fairness/blob/b1c2cd7d0ad4cac8d8d01398f653aadfd2bab5a0/artifacts/frozen_paper_evidence_20261011.zip}{original evidence archive}
contains the frozen table and figure inputs for the earlier experiments.
The \href{https://github.com/atavory/yielding_fairness/blob/b1c2cd7d0ad4cac8d8d01398f653aadfd2bab5a0/artifacts/\%62andha_xgboost_evidence_20261113.zip}{boosted-tree evidence archive}
contains the frozen test outputs for Table~\ref{tab:boosted-contraction},
the reproduction script, and its input checksums.

\subsection{Datasets, cohorts, and model settings}
\input{sections/experimental_settings_details}

\paragraph{Encoding and reporting.}
Categorical variables are one-hot encoded and
numeric variables are standardized using training rows only.  Policy
thresholds or penalties are selected on validation rows, and reported policy
results use untouched test rows.  A paper-table aggregator in the data artifact
generates manuscript table rows and confidence intervals from frozen seed-level
CSVs.  Tables
report split means with 95\% confidence intervals where space permits; compact
slash-entry tables have per-entry intervals in the generated CSVs.

For existing protected-group metrics, we report loss capture and absolute gaps
in counted blocks, blocked value, and protected loss.  For a measurement unit
$w\in\{1,A,B\}$, the group rate is
\[
 H_g^w(\pi)=
 \frac{E[\pi(\mathbf X,G)w\mid G=g]}{E[w\mid G=g]},
\]
and the reported gap is $\max_g H_g^w(\pi)-\min_g H_g^w(\pi)$.

\paragraph{Historical group-complaint ablation.}
The original-bank diagnostics below precede the normalized and enlarged
comparison in Appendix~\ref{app:normalized-prices}. Its original numbers are
retained as provenance; they are not the corrected main results.

For the group-complaint selected-policy table, each policy family is selected
only on validation rows.  The baseline is the unpriced W-BCE value-budget rule.
The group-blind account-price family subtracts the same complaint price from
all objection-bearing accounts.  The group-aware family uses the same lambda
grid but applies opposite per-group multipliers to reduce the validation
complaint gap.  The fixed family consists of the global rule and the
validation-fitted blocked-value, block-count, and protected-loss parity
postprocessors.  Within each split, family, unit, and capture floor, we
keep policies whose validation loss capture is at least the requested fraction
of the baseline validation capture, choose the smallest absolute validation
complaint gap, and score that one policy on held-out test rows.  Source-sign
tests pool rows by independent source before assigning signs: freMPL1--freMPL9
count as one French portfolio, and Belgian MTPL is counted once across
overlapping evidence families.  The cluster test in
the group-complaint selected-policy table uses the same source pooling but
retains magnitudes: each source contributes its median held-out lift, and a
one-sample t test evaluates whether the mean source lift is positive.  We
report the two-sided t p-value in the table, the directional p-value in text,
and exact sign-flip and source-bootstrap checks because there are only seven
source clusters.

\paragraph{Fixed-parity split.}
The pooled fixed-family comparison is not the headline claim because the family
contains targets with different theoretical status.  We therefore also compare
the selected group-aware price with each fixed postprocessor separately, using
the same validation-capture floor and the same held-out closure metric.  At the
$.95$ floor, raw value-unit closure, and materiality $.01$, the split is:

\begin{center}
\input{sections/fixed_parity_table}
\end{center}

These postprocessors constrain protected loss, block count, and blocked value,
respectively. Their comparisons measure performance against those distinct
targets. EU health remains in the
seven-source group-aware-versus-group-blind table, but its fixed parity
postprocessors do not clear the $.95$ validation-capture floor, so it is
recorded in the eligibility table rather than in these paired rows.

\paragraph{Attainable-ceiling diagnostic.}
For each split, we compute the best held-out value-unit closure among
group-aware rows satisfying the validation capture floor. This descriptive
ceiling diagnoses available headroom and is not a performance estimate.
At the $.95$ floor, Norwegian auto has median ceiling $.6748$ but median
selected closure zero. For freMPL8, the corresponding medians are $.3455$
and $.1450$. Thus validation transfer can leave headroom even when complaint
reduction is attainable within the evaluated bank. The artifact retains the
cohort-level ceilings, selected closures, and eligibility rows.

\paragraph{Frozen cohorts and groups.}
The headline analysis retains seven source families. Each cohort contributes
up to ten split cells, indexed by seeds 0--9, and infeasible cells are kept in
the eligibility ledgers rather than silently paired. At the primary $.95$
floor, the 180 paired cells comprise 20 Australian, 30 Belgian, 90 French,
and ten cells each for Bondora, EU health, Norwegian auto, and Swedish
motorcycle.
Australian and Belgian cached group-aware and
group-blind cohorts are separate fitted-score experiments, not additional
independent sources. French freMPL1--freMPL9 are nine datasets within one
portfolio source. The public insurance cohorts use the recorded sex or gender
field (\texttt{sex} for beMTPL97, \texttt{Gender} for freMPL1--freMPL9 and
swmotorcycle, and \texttt{Male} for norauto); cached Australian, Belgian, and
Bondora banks retain their recorded Female/Male labels. EU health uses
\texttt{gender}, restricted to F/M. KuaiRand activity cohorts and constructed
Australian stress cohorts are diagnostics outside this seven-source analysis.
These cohort identifiers and their frozen row assignments are retained in the
selection CSVs; overlapping cohorts are pooled at source level.

\paragraph{Data construction and splits.}
Fairness models use nested 80/20 splits followed by an 80/20 split of the
training portion, giving 64/16/20 training/validation/test proportions;
loss-occurrence stratification is used where the outcome permits it. These
are separate experiments from score-learning tasks with 80/0/20 splits.
For the public CASdatasets fairness cohorts, exposure is retained on no-claim rows
and loss is the claim amount. EU-health loss sums nonnegative
\texttt{amt\_} fields and value is \texttt{policy\_years} on no-claim rows;
only positive policy-years and F/M group labels are retained. The group field
is excluded from the public outcome-model predictors. Training rows determine
numeric scaling and categorical encodings. The executable artifact records
raw-file hashes, cohort filters, sample sizes, feature lists, split identities,
and model parameters; these records distinguish fitted banks that share a
source name.

\paragraph{Original policy-bank construction and selection.}
Policies are constructed from frozen fitted scores, without refitting models
for each complaint price. The promoted public-auto and cached value-unit banks
use a dense grid from $0$ to $.300$ in steps of $.005$, followed by
$\{.35,.4,.45,.5,.6,.75,1,1.5,2,3,5,10\}$.
The lambda recorded on each frozen row identifies its actual available bank.
Pricing uses the W-BCE-implied loss side
$\widetilde b=\widehat a\exp(\operatorname{clip}(q,-50,50))$, where $q$ is
the fitted W-BCE logit. Both price families share this loss side; their
zero-price order reproduces fitted W-BCE up to numerical clipping.
Separate raw-target value and loss regressions construct the harm and net-harm
objection predicates and measurement weights. They do not replace the W-BCE
loss side in this group-pricing comparison.

With $z(x)=q_{\mathcal C}(x)\widehat u_w(x)$, the common-price score is
$(\widetilde b-\lambda z)/\widehat a$ on its positive loss side.
The group-aware score replaces $\lambda z$ by
$s_g\lambda z/R_g$, where $s_g=1$ for the validation high-complaint group
and $s_g=-1$ for the low-complaint group, and
$R_g=n_{\rm val}^{-1}\sum_{i:G_i=g}\widehat u_{w,i}$.
This mass includes the group frequency required by
the complaint-disparity postprocessor proposition. Zero price is
included. Thus the families share the price axis; this is not a claim that they
have identical expressive capacity.

Scores are sorted with row index as a deterministic tie-breaker. A validation
threshold is frozen and applied to test scores. If the resulting validation
blocked-value fraction $v$ exceeds $\beta=.10$, independent uniform thinning
with probability $\alpha=\min(1,\beta/v)$ defines the evaluated randomized
policy. Metrics are its expectations: blocked value, captured loss, and
objection masses are multiplied by $\alpha$. The same frozen $\alpha$ is used
on test rows and applied separately to every family, including the baseline
and fixed postprocessors. Test budgets are reported outcomes, not constraints
used to select or adjust policies.

For $\rho\in\{.90,.95,.98,1\}$, retain only validation-budget-feasible policies
with capture at least $\rho$ times the adjusted baseline validation capture.
Choose the smallest absolute validation complaint gap, breaking ties by higher
validation capture, smaller price, and deterministic policy order. The fixed
family searches its available parity postprocessors by the same eligibility
and gap rules. A family with no eligible policy is recorded as ineligible;
it does not enter a paired comparison. All comparisons use common eligible
cohort/split cells and retain the baseline and family thinning factors.

\paragraph{Closure, aggregation, and uncertainty.}
Let $d_0$ be the unpriced baseline's absolute test complaint gap and $d_P$ the
selected policy's gap, using the same measurement unit and group denominators.
Raw closure is $C_P=1-d_P/d_0$ for $d_0>0$, and comparison lift is
$C_{\rm group}-C_{\rm comparator}$. Negative closure denotes a gap increase;
raw values are not clipped. At $d_0=0$, the implemented convention gives
closure one if $d_P=0$ and zero otherwise; such cells must be identifiable
separately because relative improvement is undefined there. Clipped closure
$\max(0,\min(1,C_P))$ is a separately labeled diagnostic.
The materiality threshold .01 classifies lift signs; it does not remove cells
merely because their baseline gap is small.

Each source contributes its median paired lift across its retained cells.
The headline cell median and the mean of seven source medians are distinct
summaries. The source-cluster statistic is the one-sample t statistic on those
seven medians, with six degrees of freedom. Exact sign flips enumerate all
$2^7$ source-sign assignments. The source bootstrap resamples seven source
medians with replacement 50,000 times and reports percentile endpoints; the
implementation uses base seed 20261007 with deterministic analysis offsets.
The artifact retains per-cell gaps, capture, budgets, eligibility, and source
pooling, so these summaries can be recomputed without rerunning model fitting.

%% file: sections/experimental_settings_details.tex
\paragraph{Score setting.}
The score-learning check uses public recommendation, insurance, credit, and
mortgage datasets: KuaiRand-Pure randomized short-video
exposures~\cite{gao2022kuairand}; motor third-party liability (MTPL) insurance data from
CASdatasets~\cite{casdatasets}; Lending Club~\cite{lendingClubData},
Prosper~\cite{prosperLoanData}, Small Business Administration (SBA)
7(a)~\cite{sba7aFoiaData}, and
Bondora~\cite{bondoraLoanData} credit portfolios; and Fannie Mae
\cite{fannieMaeHarpData} and Freddie Mac~\cite{freddieMacCrtData} mortgage
samples.
Loss $B$ is the hate indicator for videos, paid claims for insurance, and
realized credit loss for loans. Value $A$ is video play time, insurance
exposure, or realized lending return; the mortgage tasks use original
principal on non-loss loans. Australian insurance masks exposure on claim
rows; Belgian and French score tasks retain it. Lending Club, Prosper, and
Bondora mask return on loss rows; SBA retains remaining principal. Thus
the task constructions need not make $A$ and $B$ mutually exclusive.
For a descending score order, let $c(v)$ be cumulative loss capture at
cumulative value fraction $v$, with linear interpolation between accounts.
Yield Gini is $2\int_0^1 c(v)\,dv-1$, evaluated by trapezoidal integration
\cite{lorenz1905methods}. We compare fixed raw W-BCE with BCE;
log-damped W-BCE is a separate robustness comparison, never selected using
test outcomes. Positive paired mean differences occur on
\ManuscriptRawGiniPositive{} of \ManuscriptRawGiniTotal{} datasets for yield Gini
and \ManuscriptRawCapturePositive{} of \ManuscriptRawCaptureTotal{} for capture;
these counts do not establish significance for every dataset.

All score comparisons use ten paired splits and the same two-hidden-layer
multilayer perceptron (MLP; 64/32 units with rectified linear unit
(ReLU) activations), Adam with learning rate $10^{-3}$ and 20 epochs.
The archived KuaiRand runner defaults to batch size 512, but its original
run arguments were not retained; we cannot verify whether that default was
overridden. Numeric standardization and categorical encoding are
fit on training rows. Figure~\ref{fig:score-bce-wbce} uses [V] for
validation-frozen capture thresholds and [T] for retrospective test-curve
interpolation. Table~\ref{tab:score-protocol} reports the frozen sample
sizes and training/validation/test proportions. In the four 64/16/20
evaluations, the 10\% value threshold is chosen on validation and frozen on
test. The six 80/0/20 evaluations report retrospective $c(.10)$ on the
test curve, which is a retrospective ranking measure.
The freMTPL2 comparison was rerun with training-only preprocessing;
its artifact stores split identities, predictions, and transformation
parameters for independent metric replay.
Ordinary BCE uses the binary loss-occurrence label with unit weights.

\begin{table}[t]
\centering
\caption{Score samples and training/validation/test percentages. Each dataset
uses ten paired splits; SBA denotes the Small Business Administration.}
\label{tab:score-protocol}
\input{sections/score_protocol_table}
\end{table}

\paragraph{Fairness setting.}
The account-objection frontier uses one natural public cohort from each of
Australian MTPL, Belgian MTPL, and KuaiRand.  The group-complaint evaluation uses
public group-bearing evaluations from seven independent sources: Australian
MTPL, Belgian MTPL, Bondora, EU health insurance, the French
motor-insurance dataset family freMPL, Norwegian auto, and Swedish motorcycle.  The
freMPL1--freMPL9 panels are parts of one portfolio and are not counted as
independent sources. Unless
otherwise stated, policies share a 10\% aggregate value budget.
Fairness tasks use 64/16/20 training/validation/test splits and
training-fitted preprocessing. Scores use linear weighted-logistic W-BCE and BCE fit by full-batch
Adam (learning rate $.05$, $\ell_2$ (squared Euclidean norm) regularization coefficient $10^{-4}$). Separate raw-scale linear
regressions supply nonnegative value and loss proxies
(learning rate $.03$, $\ell_2$ coefficient $10^{-4}$). These predictors are fixed during
postprocessor comparisons; only thresholds and complaint prices change.
Appendix~\ref{app:exp-protocol} gives split, learner, preprocessing,
aggregation, and reporting details.

\begin{table}[t]
\centering
\caption{Group-evaluation cohorts and held-out rows per split. Each fitted
cohort has ten paired seeds; multiple fitted banks from one source remain
one source for inference. F/M means Female/Male; these labels include their recorded case variants.}
\label{tab:fairness-cohorts}
\begin{tabular}{lrrl}
\toprule
Source & Cohorts & Test rows & Groups \\
\midrule
Australian MTPL & 2 & 13,571 & Female/Male \\
Belgian MTPL & 3 & 10,000--32,642 & Female/Male \\
French portfolio & 9 & 3,200--7,660 & Female/Male \\
Bondora & 1 & 10,006 & Female/Male \\
EU health & 1 & 35,207 & F/M \\
Norwegian auto & 1 & 10,000 & 0/1 \\
Swedish motorcycle & 1 & 10,000 & Female/Male \\
\bottomrule
\end{tabular}
\end{table}

\paragraph{Group evaluation units.}
Table~\ref{tab:fairness-cohorts} records the held-out sizes and groups.
A cell is one fitted cohort and split seed; a policy family is the common-price,
signed group-price, or fixed-parity candidate set evaluated on that cell.
Seeds are $0$--$9$. The $.95$ comparison has 180 paired cells: 20 Australian,
30 Belgian, 90 French, and ten each for Bondora, EU health, Norwegian auto,
and Swedish motorcycle. Cached Australian and Belgian names denote separate
fitted banks, not independent sources; the French panels form one portfolio.
Within each source we take the median paired test lift, then average the seven
source medians and apply source-level inference. A family is eligible only if
it has a validation-budget-feasible policy clearing the capture floor; both
families must be eligible for a paired cell.
For the seven-source scope, grouping fields are recorded binary sex/gender (Female/Male or F/M);
Norwegian auto uses \texttt{Male} coded 0/1. The public insurance cohorts use
exposure on no-claim rows and paid claims; EU health uses positive policy-years
on no-claim rows and the sum of nonnegative claim-amount fields. Bondora uses
the lending value/loss construction described above. Public outcome predictors
exclude the group field. Appendix~\ref{app:exp-protocol} documents cohort construction; the public artifact records per-cohort filters and feature lists.

The full ten-source sensitivity adds two grouping cohorts per source for
Centers for Medicare \& Medicaid Services (CMS) provider fraud,
University of California, Irvine (UCI) credit, and German credit, giving 240 cells per floor.
These use $A=M(1-Y)$ and $B=MY$: $M$ is reimbursed claim amount and $Y$ the
provider's potential-fraud flag in CMS; credit limit and next-month default
in UCI; and credit amount and the bad-credit label in German credit.
They are label-based outcome proxies, not measured preventable losses.
CMS uses recorded gender and race code 1 versus other codes; UCI uses sex
and age below 40 versus at least 40; German credit uses sex derived from the
recorded personal-status code and age below 35 versus at least 35.
The two grouping attributes are pooled within their source for inference.

%% file: sections/score_protocol_table.tex
\begin{tabular}{lrr}
\toprule
Dataset & Sample size & Split \\
\midrule
KuaiRand & 1,186,059 & 64/16/20 \\
Australian MTPL & 67,856 & 64/16/20 \\
freMTPL2 & 100,000 & 80/0/20 \\
Belgian MTPL & 100,000 & 80/0/20 \\
Lending Club & 100,000 & 80/0/20 \\
Prosper & 55,051 & 80/0/20 \\
SBA 7(a) & 100,000 & 80/0/20 \\
Bondora & 52,793 & 80/0/20 \\
Fannie Mae & 300,000 & 64/16/20 \\
Freddie Mac & 300,000 & 64/16/20 \\
\bottomrule
\end{tabular}

%% file: sections/fixed_parity_table.tex
\begin{tabular}{lrrr}
\toprule
Fixed postprocessor & Median lift & Source signs & Sign $p$ \\
\midrule
Protected-loss parity & $+1.1738$ & $6/0/0$ & $0.0156$ \\
Block-count parity & $+0.1615$ & $4/2/0$ & $0.3438$ \\
Blocked-value parity & $+0.1491$ & $5/1/0$ & $0.1094$ \\
\bottomrule
\end{tabular}

%% file: sections/preprocessing_appendix.tex
\section{Preprocessing and Ranking}
\label{app:preprocessing-ranking}

We prove the preprocessing proposition and give the exact
condition under which weight damping preserves the raw W-BCE target.

\begin{theorem}[Outcome-side characterization]
\label{thm:app-characterization}
Let the outcome domain be the full cone
$\mathcal Z=\{(B,A):B,A\geq0,\ B+A>0\}$, and let
$U=W'Q'$ and $V=W'(1-Q')$ be finite measurable transformations on this domain,
where $W'>0$ and $Q'\in[0,1]$. Quantify over all finite-support distributions
on $\mathcal Z$ with positive means of $A$ and $B$, and assume the transformed
denominator has positive mean for each such distribution. The induced
proper-loss ranking
\[
\frac{E[U\mid\mathbf X=\mathbf{x}]}
     {E[V\mid\mathbf X=\mathbf{x}]}
\]
is a fixed strictly decreasing function of $\Phi(\mathbf{x})$ for all these
distributions if and only if
\begin{align}
 \begin{bmatrix}U\\V\end{bmatrix}
 &=
 \underbrace{\begin{bmatrix}p_1&p_2\\r_1&r_2\end{bmatrix}}_{M}
 \begin{bmatrix}B\\A\end{bmatrix},
 & M&\geq0, & \det(M)&=p_1r_2-p_2r_1>0.       \label{eq:app-linear-family}
\end{align}
\end{theorem}

Thus the universally valid reductions are exactly invertible nonnegative
linear recombinations of the outcomes; W-BCE is the canonical $(U,V)=(B,A)$.
The necessity claim uses the full outcome cone and arbitrary mixtures;
preservation on a restricted dataset support need not imply this characterization.
The same sufficiency holds for arbitrary distributions with finite outcome
means and positive conditional means of $A$ and $B$.
\begin{proof}
Let $\rho_M(\mathbf{x})$ denote $E[U\mid\mathbf X=\mathbf{x}]/
E[V\mid\mathbf X=\mathbf{x}]$. Linearity of expectation gives
\[
 \rho_M(\mathbf{x})
 =\frac{p_1+p_2\Phi(\mathbf{x})}
        {r_1+r_2\Phi(\mathbf{x})}
\]
and
$d\rho_M/d\Phi=-(p_1r_2-p_2r_1)/(r_1+r_2\Phi)^2<0$, proving sufficiency.  For
necessity, let $H$ be the fixed decreasing
function in the premise and fix a ratio $t$.  A conditional distribution satisfies
$E[A]/E[B]=t$ exactly when $E[A-tB]=0$.  For the transformed outcomes, the
same distribution must satisfy $E[U-H(t)V]=0$; because $H$ is one-to-one, the converse
also holds.

Now consider two outcome points $z_1,z_2$ for which
$G_t(z)=A(z)-tB(z)$ has opposite signs.  There is a unique mixture weight
$\alpha$ with
$\alpha G_t(z_1)+(1-\alpha)G_t(z_2)=0$.  The same mixture must also make
$F_t(z)=U(z)-H(t)V(z)$ have zero expectation, so
$F_t(z_1)/G_t(z_1)=F_t(z_2)/G_t(z_2)$.  Since the two points were arbitrary,
$F_t=c_tG_t$ pointwise for some scalar $c_t$ (points with $G_t=0$ follow by
mixing with any point off the hyperplane).  Thus
\[
 U-H(t)V=c_t(A-tB).
\]
Writing this identity for two distinct values of $t$ and solving the two
linear equations shows that both $U$ and $V$ are linear combinations of $B$
and $A$.  Their nonnegativity gives $M\geq0$, and the strict decrease of $H$
gives $\det(M)>0$.
\end{proof}

The first claim in the preprocessing proposition follows from the W-BCE theorem
after replacing the original outcome sides by
$U=W'Q'$ and $V=W'(1-Q')$.  The universal-preservation claim follows from
Theorem~\ref{thm:app-characterization}.

\begin{proposition}[Exact effect of weight damping]
\label{prop:app-damping}
Suppose $W>0$ and replace $W$ by $g(W)>0$ while retaining $Q=B/W$.  Put
$h(W)=g(W)/W$, and assume both conditional outcome means are positive.  The
damped optimum has inverse ratio
\[
 R(\mathbf{x})=
 \frac{E[h(W)A\mid\mathbf X=\mathbf{x}]}
      {E[h(W)B\mid\mathbf X=\mathbf{x}]}.
\]
This equals $\Phi(\mathbf{x})$ exactly when
\begin{equation}
 \operatorname{Cov}\!\left(
 h(W),\frac{A}{E[A\mid\mathbf{x}]}-
       \frac{B}{E[B\mid\mathbf{x}]}
 \;\middle|\;\mathbf X=\mathbf{x}\right)=0.
 \label{eq:app-damping-covariance}
\end{equation}
Therefore only linear $g$ preserves the target for every conditional
distribution.
\end{proposition}

\begin{proof}
The covariance term equals
\[
\frac{E[h(W)A\mid\mathbf{x}]}{E[A\mid\mathbf{x}]}
-
\frac{E[h(W)B\mid\mathbf{x}]}{E[B\mid\mathbf{x}]}.
\]
Setting it to zero is equivalent to $R(\mathbf{x})=\Phi(\mathbf{x})$.  A
nonconstant $h$ can be correlated with the centered bracket on a two-point
distribution, so universal preservation forces $h$ to be constant.
\end{proof}

%% file: sections/regularity_appendix.tex
\section{Prediction Domains and Frontier Regularity}
\label{app:regularity}

\paragraph{Weighted proper-loss boundary cases.}
The population statement for W-BCE optimizes over the closed
prediction interval $[0,1]$, with each loss extended to its proper boundary
values. Assume finite conditional means and
$0<E[A+B\mid\mathbf X=\mathbf{x}]<\infty$. Write
$q=E[B\mid\mathbf{x}]/E[A+B\mid\mathbf{x}]$. For $0<q<1$, strict properness
gives the unique minimizer $q$. At $q=0$ or $q=1$, the optimum is the corresponding
endpoint. For binary cross-entropy, use $0\log0=0$ in the weighted conditional
risk. A finite-logit sigmoid takes values only in $(0,1)$: at an endpoint its
risk has an infimum approached as the logit tends to $-\infty$ or $+\infty$,
rather than an attained finite-logit minimizer. A region with zero conditional
total stake is indifferent. These population facts do not assert consistency
of a restricted finite-sample model family.

\paragraph{Nonempty and closed attainable sets.}
For the frontier feasibility proposition, assume integrable
nonnegative $A,B$, positive finite group denominators, and integrable
coefficients in the complaint functional. Policies range over all measurable
$\pi(\mathbf X,G)\in[0,1]$, and the budget/capture feasible set is nonempty.
A sufficient certificate for a fractional floor $0\leq\rho\leq1$ is a baseline
$\pi_0$ that satisfies the value budget, with $\eta=\rho E[\pi_0B]$.

The order interval $0\leq\pi\leq1$ is weak-* compact in $L^\infty$.
Integration against the stated $L^1$ coefficients is weak-* continuous, so
budget and capture inequalities define a closed subset. Its linear complaint
image is a nonempty compact interval. Hence if zero is outside that interval,
its distance from zero is strictly positive and attained. If no policy meets
the capture and budget requirements, the attainable set is empty; no policy
or finite residual-disparity optimum is asserted. For a restricted finite
candidate bank, the corresponding statement concerns its convex mixtures,
not all measurable policies.

\paragraph{Duality and the supported policy form.}
For the complaint-disparity postprocessor proposition, retain the preceding
integrability assumptions and fix the outcome proxies and objection indicator
before optimization. In a finite empirical problem, randomized allocations
form a finite-dimensional polytope. Feasibility and boundedness give linear
programming strong duality and attained dual multipliers. In the population
problem, additionally assume no duality gap and attainment by finite
multipliers; a suitable constraint qualification, such as Slater feasibility
for the inequality constraints after accounting for equality constraints,
is sufficient. A nonempty feasible set alone does not establish this
population dual-attainment condition.

For a capture-maximizing supported point satisfying those conditions, let
$\mu\geq0$ be the budget multiplier and let the signed disparity multiplier
be $\nu=\alpha_+-\alpha_-$. Conditioning the Lagrangian on $(\mathbf X,G)$
gives coefficient
\[
 b_g(\mathbf{x})-\mu a_g(\mathbf{x})
 -\kappa_g q_{\mathcal C}(\mathbf{x})u_w(\mathbf{x},g)/U_g^w,
 \qquad \kappa_1=\nu/P_1,\quad\kappa_0=-\nu/P_0.
\]
An optimal policy blocks on its positive region, passes on its negative
region, and randomizes on ties to satisfy active constraints. This establishes
the stated policy form under dual attainment. A finite grid of prices followed
by uniform thinning is a candidate-policy evaluation, not a proof that the
selected grid policy attains this population optimum.

\paragraph{Complaint frontier feasibility.}
At a required loss-capture level, exact complaint parity may be infeasible.
Let $\pi_0$ be a budget-feasible reference policy. Fix a minimum capture requirement $E[\pi B]\geq\eta$, for example
$\eta=0.9E[\pi_0B]$ for 90\% of the baseline policy's capture.  Among all
policies that satisfy this capture requirement and the value budget, collect
the attainable complaint gaps
$\Delta_D^w(\pi)=D_1^w(\pi)-D_0^w(\pi)$.
Proposition~\ref{prop:complaint-frontier-feasibility} shows that these
attainable gap values form an interval.  Exact parity is feasible if and only
if that interval contains zero; otherwise its closest endpoint gives the
unavoidable residual gap.

\begin{proposition}[Frontier feasibility]
\label{prop:complaint-frontier-feasibility}
Fix an absolute protected-loss capture floor $\eta$, in the same units as
$E[\pi B]$, and define
\[
 \mathcal G_\eta=
 \{D_1^w(\pi)-D_0^w(\pi):
  0\leq\pi\leq1,\ E[\pi A]\leq\beta,\ E[\pi B]\geq\eta\}.
\]
For a fractional floor $\rho$ relative to a budget-feasible baseline
policy $\pi_0$, set
$\eta=\rho E[\pi_0B]$.
If this feasible set is nonempty, $A,B$ and complaint coefficients are
integrable, and group denominators are positive and finite,
$\mathcal G_\eta$ is a compact interval.  Exact complaint parity at that
operating point is feasible if and only if $0\in\mathcal G_\eta$; otherwise
the smallest achievable absolute disparity is
$\inf_{\Delta\in\mathcal G_\eta}|\Delta|>0$.
\end{proposition}

\begin{proof}
The randomized-policy feasible set is convex, and
$D_1^w(\pi)-D_0^w(\pi)$ is linear in $\pi$.  A linear image of a convex set in
$\mathbb R$ is an interval.  The order interval $0\leq\pi\leq1$ is weak-* compact in $L^\infty$.
Integration against the integrable objective and constraint coefficients is
weak-* continuous, so the feasible set is closed and compact. Its complaint
image is therefore a compact interval; the remaining claims follow by taking
its distance from zero.

\end{proof}

Together, Propositions~\ref{prop:complaint-disparity-frontier}
and~\ref{prop:complaint-frontier-feasibility} identify the population
allocation and feasibility targets. \ifdefined\ManuscriptMainPaper In Section~\ref{subsec:exp-group},
we approximate them with\else The empirical comparison approximates them with\fi fixed score families and mixtures of frozen
policy banks.

At exactly maximum capture, the budget dual coefficient fixes every
decision outside the threshold tie set. Corollary~\ref{cor:max-capture-ties}
shows that a budget-feasible reallocation within this set is the only way
to change complaint disparity without capture loss. When the budget
multiplier $\mu$ is positive,
such a reallocation must preserve expected blocked value. Without tie mass,
the maximum-capture policy is unique almost surely. These statements
concern the coefficients actually optimized; held-out gains over a fitted
baseline need not preserve the population optimum.

%% file: sections/objection_algorithm_appendix.tex
\section{Account-Objection Validation Algorithm}
\label{app:objection-alg}

The account-objection frontier figure uses ten paired seeds for each
natural Australian MTPL, Belgian MTPL, and 300K KuaiRand cohort. Models are
fit on training rows. Within a split, every policy uses the same validation
and test accounts, observed value and loss, and frozen harm and net-harm
criterion decisions. The latter are block indicators $H$ and $N$; hence
$F_h=1-H$, $F_n=1-N$, and the union objection is
$F_{\rm any}=F_h\lor F_n$. A block satisfying both objections contributes
once to the plotted count-objection mass. The W-BCE reference is not an
objecting criterion.

Let $q$ and $q_b$ be fitted W-BCE and BCE logits, and let
$\hat a,\hat b$ be the nonnegative raw-target value and loss predictions.
The three unpriced orders rank descending by $q$, $q_b$, and
$\hat b/\hat a$, respectively. Negative fitted logits remain eligible.
The plug-in union-price order ranks by
$(\hat b-\lambda F_{\rm any})/\hat a$ on its positive loss side.
A second union-price family replaces $\hat b$ by $\hat a\exp(q)$,
so its zero-price order reproduces fitted W-BCE. The separate-price family
uses $(\hat b-\lambda_h F_h-\lambda_n F_n)/\hat a$; its penalty can count
both criteria while the plotted union mass still counts each account once.
The executable checks reproduce the unpriced masks and thinning factors at
zero price and recover the stored BCE decisions.

Prices include the legacy absolute grids and validation-scaled values.
For each loss-side family, positive predicted loss-side values among
validation accounts with $F_{\rm any}=1$ determine their median and the
90th, 95th, 99th percentiles and maximum. The grid includes those quantiles
and the median times
\[\{.01,.02,.05,.1,.2,.5,1,2,5,10,20,50\}.\]
Separate harm and net-harm prices use pairs from the plug-in grid.
Across seeds, summaries join the same relative price rule, retaining its
seed-specific absolute prices in the policy bank.

For every family, a validation threshold targets 10\% blocked value.
Threshold ties can exceed that amount; uniform thinning with probability
$\alpha=\min(1,.10/v_{\rm val})$ corrects the resulting validation value
fraction $v_{\rm val}$. The threshold, absolute prices, and $\alpha$ are
then frozen on test. Reported loss, value, and objection masses are
expectations under thinning, and test value spend is recorded separately.

For policy $P$ in each test split, write $L_P$ for loss capture and $O_P$
for union objection count divided by the number of accounts. The plot
averages $L_P/L_0$ and $1-O_P/O_0$ over the ten paired splits, where zero
denotes fitted W-BCE. All reference captures and objection shares are
positive in these three cohorts. Intervals are paired Student t intervals.
Nondominance is computed after averaging; the frontier and its zero-objection
endpoints are descriptive summaries of evaluated policies. They do not
select a deployable policy using held-out outcomes. The group experiment
uses validation-only family selection instead.

The data artifact's common-frontier and endpoint-table scripts generate the
policy bank, summaries, and endpoint table. Their provenance
records source and candidate-input checksums. KuaiRand's recovered BCE
logits use the accepted raw-target runner: sample identities, splits, raw
outcomes, and W-BCE logits match the earlier frozen bank; raw-target outcome
proxies and their derived criterion decisions supersede its old proxy lineage.

%% file: sections/score_account_results_appendix.tex
\section{Score Estimation and Account-Level Endpoints}
\label{app:score-account-results}

\paragraph{Score estimation.}
We report the full score comparison under the protocols of
Appendix~\ref{app:exp-protocol}. Yield Gini summarizes the ordering;
the [V] capture points use validation-frozen thresholds, and the [T]
points summarize retrospective test curves.

\ifdefined\ManuscriptMainPaper\else
\input{sections/wbce_score_figure}
\fi

\begin{figure*}[t]
\centering
\includegraphics[width=\textwidth]{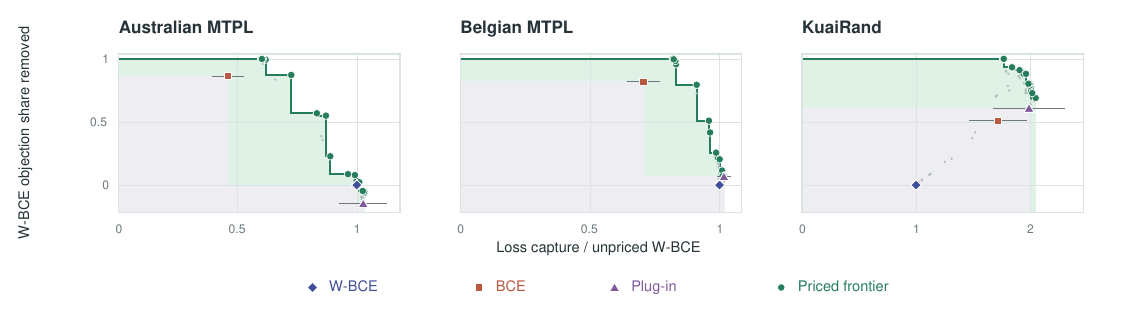}
\caption{Account-objection trade-offs relative to fitted W-BCE. Green points
are nondominated priced policies. Gray shading is dominated by unpriced
baselines; green shading shows the additional dominated region. Markers show
W-BCE, BCE, and plug-in ratios with 95\% paired-split intervals. Policies use
validation-frozen thresholds and thinning at a 10\% value cap. The test-mean
frontier is descriptive; higher and farther right is better.}
\Description{Three panels compare loss capture and account-objection reduction relative to fitted W-BCE across the evaluated policy bank. The envelope uses held-out means. All baseline points are visible, including KuaiRand capture ratios above one.}
\label{fig:account-objection-pareto}
\end{figure*}

\paragraph{Descriptive account frontier.}
A blocked account can object when a harm or net-harm criterion would pass
it at the same value budget \ifdefined\ManuscriptMainPaper
(Eq.~\eqref{eq:criterion-objection})\fi.
Figure~\ref{fig:account-objection-pareto} compares fitted W-BCE, BCE, and
unpriced plug-in ratios with union- and separate-objection price policies
on the same accounts. A union-price policy charges one penalty when either
norm would pass the account; a separate-price policy sums the harm and
net-harm penalties and can charge both. Union prices apply to plug-in loss
and fitted W-BCE implied loss; separate prices use plug-in loss.
Appendix~\ref{app:objection-alg} gives the scores,
validation-scaled price grids, and common budget correction.

This figure is a descriptive envelope, not a selected result. Nondominance is
computed on held-out means over the evaluated bank, the bank contains the
unpriced baseline as its zero-price member, and the baseline is also the
normalizer, so it sits at the origin of the plotted axes with no
spread. The envelope therefore describes which operating points the evaluated
bank reached; it does not select a policy using held-out outcomes and does not
establish an improvement over any marker. Complete objection removal retains
about 60\% of W-BCE capture in Australian MTPL and 82\% in Belgian MTPL
(Appendix~\ref{app:score-account-results}); KuaiRand has zero-objection policies
with greater capture than fitted W-BCE, which reflects a weak fitted baseline
on that cohort. \ifdefined\ManuscriptMainPaper Section~\ref{subsec:exp-group} supplies the selected
comparison.\else The main paper supplies the selected comparison.\fi

\paragraph{Account-level endpoints.}
\begin{table}[h]
\centering
\scriptsize
\setlength{\tabcolsep}{3pt}
\caption{Natural-cohort endpoints from the same corrected policy bank as the
account-objection frontier figure. Entries are means $\pm$ 95\%
paired-split intervals. $L/L_0$ is loss capture relative to fitted W-BCE,
$1-O/O_0$ is its count-objection share removed, and $V$ is held-out blocked
value divided by total value. Each priced endpoint has zero objection in
every test split and maximizes mean test capture among such evaluated
policies. This endpoint selection is descriptive, not validation-selected
performance. All families use the same validation cap and frozen thinning.}
\label{tab:objection-penalized}
\resizebox{\columnwidth}{!}{%
\input{sections/figure3_endpoint_table}
}
\end{table}

%% file: sections/figure3_endpoint_table.tex
\begin{tabular}{llccc}
\toprule
Cohort & Policy & $L/L_0$ & $1-O/O_0$ & $V$ \\
\midrule
Australian MTPL & W-BCE & $1.000\pm0.000$ & $0.000\pm0.000$ & $0.100\pm0.002$ \\
 & Priced endpoint & $0.599\pm0.125$ & $1.000\pm0.000$ & $0.098\pm0.004$ \\
\addlinespace
Belgian MTPL & W-BCE & $1.000\pm0.000$ & $0.000\pm0.000$ & $0.101\pm0.001$ \\
 & Priced endpoint & $0.820\pm0.074$ & $1.000\pm0.000$ & $0.100\pm0.002$ \\
\addlinespace
KuaiRand & W-BCE & $1.000\pm0.000$ & $0.000\pm0.000$ & $0.101\pm0.002$ \\
 & Priced endpoint & $1.769\pm0.258$ & $1.000\pm0.000$ & $0.086\pm0.004$ \\
\addlinespace
\bottomrule
\end{tabular}

%% file: sections/group_disparity_decomposition_appendix.tex
\section{Group Disparity Decomposition}
\label{app:group-disparity-decomposition}

We give the denominator decomposition used in the main group-disparity section
and in the empirical analysis.

Denote the main text's group account-objection indicator by
$q_{\mathcal C}(\mathbf{x})$, so
\begin{align}
 O_{\mathcal C}(\mathbf X,G;\pi)
 &=
 \pi(\mathbf X,G)q_{\mathcal C}(\mathbf X).
 \label{eq:app-group-objection-factor}
\end{align}
For groups with $E[q_{\mathcal C}(\mathbf X)U_w\mid G=g]>0$, define
\begin{align}
 e_g^w
 &=
 \frac{E[q_{\mathcal C}(\mathbf X)U_w\mid G=g]}
      {E[U_w\mid G=g]},
 \label{eq:group-exposure}\\
 R_g^w(\pi)
 &=
 \frac{E[\pi(\mathbf X,G)q_{\mathcal C}(\mathbf X)U_w\mid G=g]}
      {E[q_{\mathcal C}(\mathbf X)U_w\mid G=g]} .
 \label{eq:group-treatment}
\end{align}
Here $e_g^w$ is fixed by the group distribution and objecting criteria, while
$R_g^w(\pi)$ is chosen by the policy.

\begin{proposition}[Association--treatment decomposition]
\label{prop:complaint-factorization}
For each group with positive exposure,
\[
 D_g^w(\pi)=e_g^wR_g^w(\pi).
\]
For two such groups, if treatment parity holds with
$R_1^w(\pi)=R_0^w(\pi)=r$, then
\[
 D_1^w(\pi)-D_0^w(\pi)=r(e_1^w-e_0^w).
\]
If the treatment rates are positive, exact complaint parity
$D_1^w(\pi)=D_0^w(\pi)$ is equivalent to
\[
 \frac{R_1^w(\pi)}{R_0^w(\pi)}
 =
 \frac{e_0^w}{e_1^w}.
\]
\end{proposition}

\begin{proof}
For each group with positive exposure,
\begin{align*}
 e_g^wR_g^w(\pi)
 &=
 \frac{E[q_{\mathcal C}(\mathbf X)U_w\mid G=g]}
      {E[U_w\mid G=g]}
 \cdot
 \frac{E[\pi(\mathbf X,G)q_{\mathcal C}(\mathbf X)U_w\mid G=g]}
      {E[q_{\mathcal C}(\mathbf X)U_w\mid G=g]}\\
 &=
 \frac{E[\pi(\mathbf X,G)q_{\mathcal C}(\mathbf X)U_w\mid G=g]}
      {E[U_w\mid G=g]}
 =
 D_g^w(\pi),
\end{align*}
using Eq.~\eqref{eq:app-group-objection-factor}.  If
$R_1^w(\pi)=R_0^w(\pi)=r$, then
\begin{align*}
 D_1^w(\pi)-D_0^w(\pi)
 &=
 e_1^wR_1^w(\pi)-e_0^wR_0^w(\pi)\\
 &=
 r(e_1^w-e_0^w).
\end{align*}
The last claim follows by rearranging
$e_1^wR_1^w(\pi)=e_0^wR_0^w(\pi)$.
\end{proof}

\begin{figure*}[t]
\centering
\includegraphics[width=\textwidth]{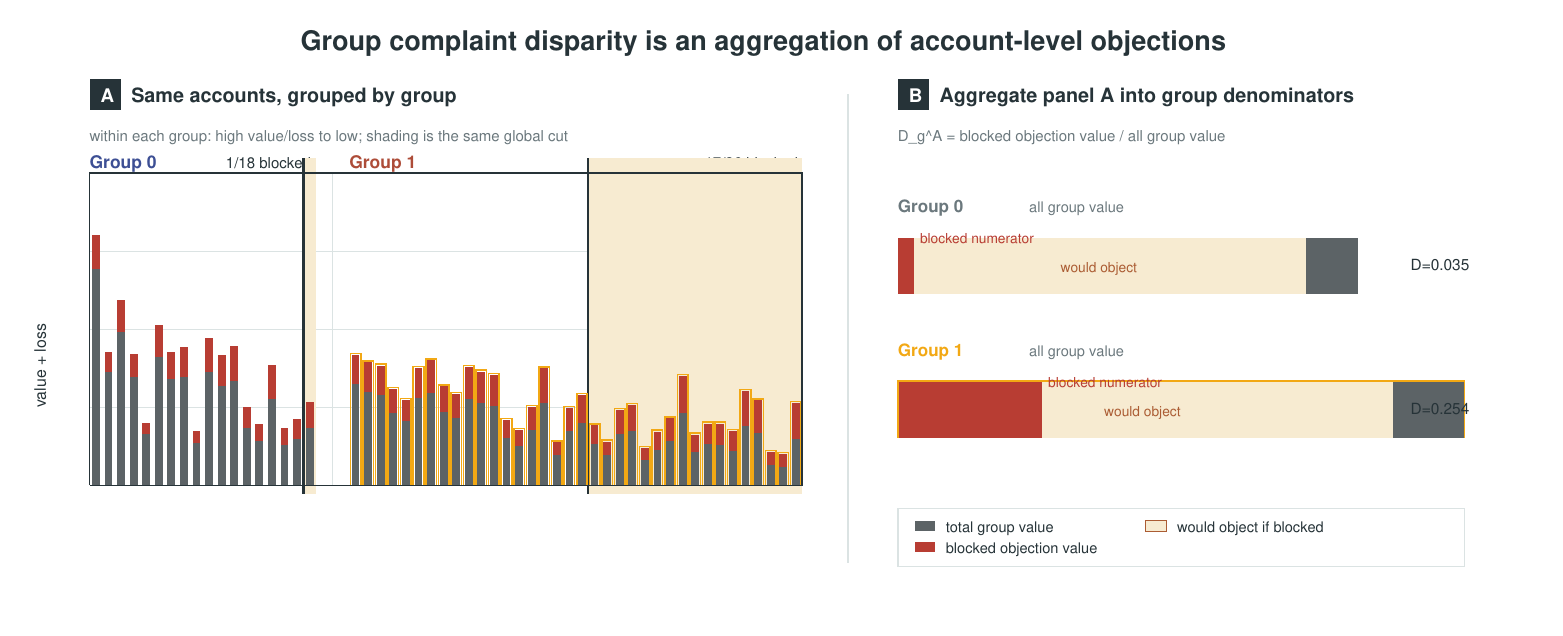}
\caption{Denominator choice for group complaint disparity.  The orange outline
marks group 1, as in
\ifdefined\ManuscriptMainPaper Figure~\ref{fig:greedy-overview}\else
the running example in the main paper\fi. Panel A shows the same
global value-per-loss threshold applied
to both groups.  Panel B aggregates the resulting objections in value units.
$D_g^A$ counts blocked objection-bearing value and divides by all value in
group $g$.  Dividing instead by all objection-bearing value would report only
the conditional blocking rate among accounts that could object.}
\Description{An illustrative account table grouped by protected group, followed by bars for total group value, objection-bearing value, and blocked objection-bearing value. The complaint rate divides the last quantity by total group value.}
\label{fig:group-denominator-choice}
\end{figure*}

\paragraph{Score information by objection unit.}
We use the objection units from the main account-level definition and fix
the norm indicator $q_{\mathcal C}$ before adjusting the policy.  Write
$a_g(\mathbf{x})=E[A\mid\mathbf X=\mathbf{x},G=g]$,
$b_g(\mathbf{x})=E[B\mid\mathbf X=\mathbf{x},G=g]$, and
$u_w(\mathbf{x},g)=E[U_w\mid\mathbf X=\mathbf{x},G=g]$.
Let $U_g^w=E[U_w\mid G=g]$ be the positive finite group denominator,
and let $\kappa_g$ be a fixed finite group coefficient.

\begin{proposition}[Score information by objection unit]
\label{prop:unit-score-information}
At a covariate value with $0<a_g(\mathbf{x})<\infty$ and
$0\leq b_g(\mathbf{x})<\infty$, define
$r_g(\mathbf{x})=b_g(\mathbf{x})/a_g(\mathbf{x})$ and
\[
 S_g^w(\mathbf{x})=
 \frac{b_g(\mathbf{x})-
 \kappa_gq_{\mathcal C}(\mathbf{x})u_w(\mathbf{x},g)/U_g^w}
 {a_g(\mathbf{x})}.
\]
Then
\begin{align}
 S_g^A(\mathbf{x})
 &=r_g(\mathbf{x})-
   \frac{\kappa_gq_{\mathcal C}(\mathbf{x})}{U_g^A},\nonumber\\
 S_g^B(\mathbf{x})
 &=r_g(\mathbf{x})
   \left(1-\frac{\kappa_gq_{\mathcal C}(\mathbf{x})}{U_g^B}\right),\nonumber\\
 S_g^1(\mathbf{x})
 &=r_g(\mathbf{x})-
   \frac{\kappa_gq_{\mathcal C}(\mathbf{x})}{a_g(\mathbf{x})}.
 \label{eq:unit-adjusted-scores}
\end{align}
For fixed norm indicator, group, denominators, and coefficients, $r_g$
determines $S_g^A$ and $S_g^B$.  When
$\kappa_gq_{\mathcal C}(\mathbf{x})\ne0$, no function of those fixed
quantities and $r_g$ alone determines $S_g^1$ for all positive conditional
values $a_g$.
\end{proposition}

\begin{proof}
Substitute $u_A=a_g$, $u_B=b_g$, and $u_1=U_g^1=1$ into the score.
For the count claim, consider two covariate values in the same group with
$q_{\mathcal C}=1$, the same $r>0$, and conditional means
$(a_g,b_g)=(t,rt)$ and $(2t,2rt)$, where $t>0$.
For any fixed $\kappa_g\ne0$, their count scores are
$r-\kappa_g/t$ and $r-\kappa_g/(2t)$.  These scores differ although the
ratio, indicator, and group quantities agree.
\end{proof}

This calculation concerns score adjustment after we fix the auxiliary
quantities.  Estimating the norm indicator and group denominators, and
calibrating the blocked-value budget, may still require auxiliary outcome
predictions.  A nonzero count penalty also needs conditional value scale, or
equivalent information such as conditional total stake.  For $a_g=0$, we use the
undivided decision coefficient rather than a ratio.  Signed penalties can
produce negative adjusted scores; these remain ranking scores and do not
require negative W-BCE training weights.

\paragraph{Maximum-capture alternatives.}
We compare policies using the same outcome coefficients and blocked-value
budget.  Write $a=a_G(\mathbf X)$ and $b=b_G(\mathbf X)$, and let
$z_+=\max(z,0)$.  The following consequence of the budget optimality
conditions identifies the candidates on which exact maximum capture permits
different complaint outcomes.

\begin{corollary}[Maximum-capture tie set]
\label{cor:max-capture-ties}
Let $a,b$ be nonnegative integrable coefficients and $\beta\geq0$, and define
\[
 L^\star=\max_{0\leq\pi\leq1,\ E[\pi a]\leq\beta}E[\pi b].
\]
Assume primal attainment, no duality gap, and a finite attained budget
multiplier $\mu\geq0$, so
$L^\star=\mu\beta+E[(b-\mu a)_+]$.
Set $c=b-\mu a$ and $T=\{c=0\}$.
For every budget-feasible policy $\pi$,
\begin{align}
 L^\star-E[\pi b]
 &=\mu\bigl(\beta-E[\pi a]\bigr)\nonumber\\
 &\quad+E[c_+(1-\pi)+(-c)_+\pi].
 \label{eq:max-capture-dual-gap}
\end{align}
A budget-feasible policy attains $L^\star$ if and only if it satisfies
$\pi=1$ on $(c>0)$ and $\pi=0$ on $(c<0)$ almost surely, and
$\mu(\beta-E[\pi a])=0$.
All maximizing policies agree outside $T$.  If $\Pr(T)=0$, the maximizing
policy is unique almost surely, and every maximizing policy has the same
complaint disparity for fixed integrable complaint coefficients, positive
group probabilities, and positive finite group denominators.
\end{corollary}

\begin{proof}
Substitute $b=c+\mu a$ into the attained dual value.  The identity
$c_+-\pi c=c_+(1-\pi)+(-c)_+\pi$ gives
Eq.~\eqref{eq:max-capture-dual-gap}.  Both terms on its right are
nonnegative.  Their sum vanishes exactly under the stated budget condition
and policy values outside $T$.  Thus maximizing policies can differ only
on $T$.  A null tie set gives almost-sure uniqueness; integration against
the fixed complaint coefficients then gives the same disparity.
\end{proof}

When $\mu>0$, every maximizing policy spends the full blocked-value budget.
Holding its decisions outside $T$ fixed, we can change complaint outcomes
without reducing capture only by reallocating blocks within $T$ while
preserving its expected blocked value.  The tie set's complaint mass alone
therefore does not determine the attainable disparity changes.
When $\mu=0$, $T$ consists of candidates with zero conditional loss,
including candidates with positive conditional value; tie choices need only
respect the remaining budget.  Candidates with $a=b=0$ contribute no
value- or loss-unit complaint mass, but they can still affect count
complaints.  We apply the corollary to exact maximum capture under the fixed
coefficients.  Nearly tied candidates may permit changes at arbitrarily small
capture cost, and held-out outcomes need not preserve the fitted optimum.

\paragraph{Constrained complaint allocation.}
Given $\Delta_D^w$, group complaint-disparity control is a constrained
allocation problem.  With two groups (for simplicity), we maximize
prevented loss subject to the blocked-value budget and a complaint-disparity
target with tolerance $\epsilon\geq0$,
\begin{equation}
 \max_{0\leq\pi\leq1} E[\pi B]
 \quad\text{s.t.}\quad
 E[\pi A]\leq\beta,\qquad
 |\Delta_D^w(\pi)|\leq\epsilon .
 \label{eq:group-complaint-objective}
\end{equation}
Sweeping the complaint tolerance or its multiplier gives the capture--disparity
frontier.  With more than two groups, the scalar complaint multiplier becomes a
vector of group coefficients and the same Lagrange form applies.

Let
\[
 u_w(\mathbf{x},g)=
 \begin{cases}
  1, & w=1,\\
  E[A\mid\mathbf X=\mathbf{x},G=g], & w=A,\\
  E[B\mid\mathbf X=\mathbf{x},G=g], & w=B,
 \end{cases}
\]
and use validation rows to estimate each group denominator.

Let $P_g=\Pr(G=g)>0$. Attach nonnegative multipliers $\alpha_+$ and
$\alpha_-$ to $\Delta_D^w\leq\epsilon$ and $-\Delta_D^w\leq\epsilon$,
respectively, and set $\nu=\alpha_+-\alpha_-$.
The conditional coefficients use $\kappa_1=\nu/P_1$ and
$\kappa_0=-\nu/P_0$, because the capture objective averages groups with
weights $P_g$ while each complaint rate conditions on its own group.
Write $\kappa=(\kappa_0,\kappa_1)$ for this group coefficient vector.
Let $\mu\geq0$ be the blocked-value budget multiplier.

\begin{proposition}[Optimal complaint-disparity postprocessor]
\label{prop:complaint-disparity-frontier}
Fix a measurement unit $w$ and the accepted criterion indicator
$q_{\mathcal C}$. Let
$U_g^w=E[u_w(\mathbf X,g)\mid G=g]>0$ be the group denominator.  Define $a_g(\mathbf{x})=E[A\mid\mathbf X=\mathbf{x},G=g]$ and
$b_g(\mathbf{x})=E[B\mid\mathbf X=\mathbf{x},G=g]$.
Assume integrable outcome and complaint coefficients, finite positive group
denominators, no duality gap, and attainment by finite multipliers.
At a capture-maximizing supported point on the linear blocked-value and
complaint-disparity frontier, an
optimal randomized binary policy has the form
\begin{equation}
\pi_{\mu,\kappa}(\mathbf{x},g)
 =
 \mathbf1(
 b_g(\mathbf{x})-\mu a_g(\mathbf{x})
 -\kappa_g q_{\mathcal C}(\mathbf{x})u_w(\mathbf{x},g)/U_g^w>0
 ),
 \label{eq:group-complaint-tilt}
\end{equation}
with randomization on ties.
\end{proposition}

\begin{proof}
In the finite empirical problem, the randomized allocation set is a polytope;
Appendix~\ref{app:regularity} gives the population conditions. The objective
and constraints in Eq.~\eqref{eq:group-complaint-objective} are linear in
$\pi$.  Attach a multiplier $\mu$ to the blocked-value budget and group
multipliers to the complaint constraints.  After conditioning on
$(\mathbf X,G)=(\mathbf{x},g)$, the Lagrangian is pointwise linear in
$\pi(\mathbf{x},g)$ with coefficient
$b_g(\mathbf{x})-\mu a_g(\mathbf{x})
-\kappa_g q_{\mathcal C}(\mathbf{x})u_w(\mathbf{x},g)/U_g^w$, giving
Eq.~\eqref{eq:group-complaint-tilt}; ties may be randomized to hit exact
constraints.
\end{proof}

The same algebra applies to a finite fitted allocation problem when its
objective, budget, unit weights, and group denominators all use fixed
coefficients. Such a fitted optimum need not maximize population capture.

Proposition~\ref{prop:complaint-disparity-frontier} gives a
tilted value/loss rule.  The multiplier $\mu$ sets the blocked-value operating
point, while a signed group-penalty contrast tilts decisions against
$\Delta_D^w$.  Because Eq.~\eqref{eq:group-complaint-objective} constrains
$|\Delta_D^w|$, the active side fixes the sign: if $D_1^w>D_0^w$, the tilt
penalizes group-1 objections relative to group 0; if $D_1^w<D_0^w$, it reverses.
For the fixed allocation coefficients, varying the supported multipliers
and choosing $\mu$ gives the capture--complaint frontier. The finite
empirical bank samples these score families.

For $a_g(\mathbf{x})>0$, let
$r_g(\mathbf{x})=b_g(\mathbf{x})/a_g(\mathbf{x})$.
The score adjustment in Eq.~\eqref{eq:group-complaint-tilt} is
\[
 r_g-\kappa_gq_{\mathcal C}
 \frac{u_w}{a_gU_g^w}.
\]
Value-unit objections give $r_g-\kappa_gq_{\mathcal C}/U_g^A$;
loss-unit objections give $r_g(1-\kappa_gq_{\mathcal C}/U_g^B)$;
count-unit objections give $r_g-\kappa_gq_{\mathcal C}/a_g$.
Proposition~\ref{prop:unit-score-information} shows that, after fixing the
auxiliary indicators and normalizers, the ratio determines the value- and
loss-unit adjusted scores, while count adjustments generally need absolute
outcome scale. Applied with the same conditioning variables,
\ifdefined\ManuscriptMainPaper Theorem~\ref{thm:wbce} gives\else
the weighted proper-loss target gives\fi
$q_{{\rm W},g}(\mathbf{x})=
E[B\mid\mathbf X=\mathbf{x},G=g]/E[A+B\mid\mathbf X=\mathbf{x},G=g]$
and $r_g=q_{{\rm W},g}/(1-q_{{\rm W},g})$.
Estimating the criterion indicators, normalizers, and value budget still
requires auxiliary information. The multiplicative loss-unit identity
requires $u_B=b_g$. In the empirical score construction, the auxiliary
loss weight need not equal the W-BCE implied loss coefficient; those scores
use the general adjustment above
(Appendix~\ref{app:normalized-prices}).

%% file: sections/normalized_price_sensitivity_appendix.tex
\section{Normalized Bank and Group Outcomes}
\label{app:normalized-prices}

The original price bank used the same numerical axis for penalties expressed
in different units. A common-price policy subtracted $\lambda z$, whereas a
group-price policy subtracted $s_g\lambda z/R_g$. Consequently, matching
numerical prices did not match their effective score shifts. Validation
selections at the largest available price also indicated truncated coverage.
We therefore report a post-review reanalysis on the same frozen candidate
logs. This analysis is not a new, independently collected test evaluation.

\paragraph{Empirical score construction.}
For the empirical scores, let $\widehat a$ be the nonnegative value proxy and
$\widetilde b=\widehat a\exp(\operatorname{clip}(s_\theta,-50,50))$
the fitted W-BCE logit's implied loss side. The auxiliary loss regression
constructs objections and loss-unit weights; it does not replace
$\widetilde b$. Let $\widehat u_w$ be $1$, $\widehat a$, or the nonnegative auxiliary
loss proxy for count, value, or loss units, respectively.
Here $\lambda\ge0$ is a scalar complaint price, and $n_{\rm val}$ is the
number of validation accounts. With $z=q_{\mathcal C}\widehat u_w$, define
$R_g=n_{\rm val}^{-1}\sum_{i:G_i=g}\widehat u_{w,i}$ and
$R_{\rm all}=R_0+R_1$. The normalized scores are
\[
 s_{\rm blind}=\frac{\widetilde b-\lambda z/R_{\rm all}}{\widehat a},
 \qquad
 s_{\rm group}=\frac{\widetilde b-s_g\lambda z/R_g}{\widehat a},
\]
where $s_g=1$ for the validation high-complaint group and $-1$ for the low
group. Positive adjusted loss is eligible; zero predicted value retains the
original score convention. These are plug-in policies, with finite-model and
clipping error, rather than guarantees of population-optimal capture.

\paragraph{Common price units and preservation.}
Write $z=q_{\mathcal C}\widehat u_w$, with the estimated count, value, or loss
weight used by the existing experiment, and
$R_{\rm all}=n_{\rm val}^{-1}\sum_i\widehat u_{w,i}=R_0+R_1$.
The normalized common-price policy scores
$(\widetilde b-\lambda z/R_{\rm all})/\widehat a$;
the group-price policy scores
$(\widetilde b-s_g\lambda z/R_g)/\widehat a$.
Only positive adjusted loss is eligible for blocking, with the existing
zero-predicted-value convention retained. Normalization puts the scalar
multiplier in comparable aggregate units; it does not make the two families'
pointwise score shifts equal or make the families nested.

For every cohort, seed, and measurement unit, let $\mathcal G$ be the original
scalar grid. Both families search the same union
$\mathcal G\cup R_{\rm all}\mathcal G$. Original group policies occur at
$\lambda\in\mathcal G$, and original common-price physical policies occur at
$\lambda=R_{\rm all}\lambda_{\rm old}$. Thus the corrected candidate sets
retain both original families' policies.

\paragraph{Validation-only coverage and freezing.}
Starting from the union's maximum, we add geometric doublings to both families.
We continue until no selected policy at any of the four capture floors is at
the largest price and two further extensions leave all selected validation
policies unchanged. This is a finite coverage diagnostic, not an optimality
guarantee over continuous prices. All thresholds, thinning probabilities,
selected policies, and LP mixture weights are written to a validation freeze
before test-policy evaluation. Predictors, outcome proxies, objection
predicates, splits, the $.10$ value cap, and deterministic ties are unchanged.
The LP uses the full enlarged common-price bank, minimizing validation absolute
signed gap, then maximizing capture and minimizing spend within numerical
tolerances; its original constraints are checked after solution.

\paragraph{Source scope and aggregation.}
The seven insurance-dominated sources remain one reported scope. The extended
scope includes all recorded applicable binary-group sources, adding CMS
provider-fraud, UCI credit, and German credit cohorts. Multiple grouping
attributes or fitted cohorts from one source are pooled before source-level
inference. Neither negative nor zero effects are removed. Count and loss
units are reported alongside value units. Source medians and source means are
separate estimands; each averages one summary per source. Frozen input hashes,
coverage records, eligibility, selected component group masses, and all
source-level results are available in the public reproduction artifact.

\paragraph{Additional aggregation checks.}
At the $.95$ floor, the seven-source cell median lift is $+.2453$ and
its mean of source medians is $+.2185$ (two-sided $p=.3071$).
Across all ten recorded sources, these are $+.0803$ and $+.0863$
($p=.6007$). Neither comparison establishes a statistically significant
average group-aware advantage. The seven-source mean of source means is
$+.0977$ ($p=.6997$); the ten-source mean is $-.3979$ ($p=.3097$).
Raw cell means are $-.2274$ and $-.5592$, respectively. These summaries
measure different effects and retain the negative cells rather than clipping
them. French and Norwegian source medians are negative in the seven-source
scope; CMS and UCI credit also have negative medians in the extended scope.
The seven-source exact sign-flip directional $p$ is $.1094$.
An independent run with a common global axis extended through $512000$
gives ten-source mean source-median lift $+.0202$ ($p=.8990$), likewise
not establishing an average advantage.

\paragraph{Independent global-grid sensitivity.}
A separate replay extends a common global base axis to $512000$ before
applying the same original-policy preservation mapping to every cell.
All 240 value-unit cells are eligible at each of the four capture floors;
selections are unchanged at maxima $128000$, $256000$, and $512000$.
At the $.95$ floor, its ten-source mean of source medians is $+.0202$
(two-sided source $t$-test $p=.8990$); its mean of source means is $-.6209$
($p=.1207$). This wider bank is a separate sensitivity, with the same frozen
predictors and units. It supports the same absence of a significant average
advantage; the LP table uses the per-cell bank defined above.

\paragraph{Historical comparison.}
Before normalization and extension, the $.95$ seven-source comparison had cell
median lift $+.5447$, mean source-median lift $+.4458$, and two-sided source
$t$-test $p=.002938$. Its mean of source means was $+.3014$ with $p=.1482$;
the cell mean was $-.0085$. Adding all ten recorded sources to that historical
bank gave mean source-median lift $+.2444$ with $p=.1009$. These historical
numbers describe the original finite bank, rather than the corrected
comparison reported in the main text.

\paragraph{LP comparator and source-level results.}
We also compare with a linear program (LP) over the full frozen
group-blind account-price bank; $C_{\rm LP}$ denotes its mixture's closure.
The LP chooses nonnegative mixture weights
summing to one, minimizing the absolute signed validation complaint gap
subject to the same capture floor and expected blocked-value cap.
Among gap-optimal mixtures it maximizes capture, then minimizes spend;
each stage is checked against the original constraints. Component thinning
and mixture weights are frozen on test, where we report expected outcomes.
The comparator optimizes the available policy bank, not all possible
candidate-level decisions.

\begin{table}[t]
\centering
\caption{Group-aware prices versus linear-program (LP) mixtures of the
group-blind account-price bank at the $.95$ capture floor. Gaps and lift
$C_{\rm group}-C_{\rm LP}$ are separately computed source medians over
held-out cells; $C_{\rm LP}$ is the mixture complaint closure and positive
lift favors group pricing. Both families use the
same validation constraints and the enlarged normalized price axis. All ten
recorded sources are retained.}
\label{tab:matched-complaint-lp}
\input{sections/normalized_lp_table}
\end{table}

\begin{table*}[t]
\centering
\caption{Corrected closure-lift sensitivity across measurement units and capture
floors. Each cell averages one source summary: medians or means. The seven-source
and full ten-source scopes are distinct. Negative results are retained; all
source test statistics and exact sign-flip results are in the frozen artifact.}
\label{tab:normalized-currencies}
\input{sections/normalized_currency_table}
\end{table*}

\paragraph{Per-group outcomes and rate caps.}
We reconstruct the unpriced policy from its frozen score threshold and
thinning probability and combine the selected policies' group amounts
using their frozen weights. We keep zero-gap cells and report each group's
normalized objections, absolute gaps, capture, and blocked-value spend.
These quantities identify gap reductions that increase another group's
objections.

For the additional capped comparison, let $L_k$, $V_k$, and $D_{g,k}^A$
be the frozen validation capture fraction, blocked-value fraction, and
group complaint level of bank policy $k$. We use every common and signed
bank policy and choose mixture weights through
\[
\begin{aligned}
 \min_{\omega_k\geq0}\quad&
 \left|\sum_k\omega_k(D_{1,k}^A-D_{0,k}^A)\right|\\
 \text{s.t.}\quad&
 \sum_k\omega_k=1,\qquad
 \sum_k\omega_kL_k\geq.95L_0,\\
 &\sum_k\omega_kV_k\leq.10,\qquad
 \sum_k\omega_kD_{g,k}^A\leq D_g^A(\pi_0)
 \quad(g=0,1).
\end{aligned}
\]
The baseline belongs to the bank, so the validation constraints are
feasible. Among gap-optimal mixtures we maximize capture and then minimize
spend. We check the original constraints after every stage and freeze all
weights before reading test outcomes. The solver uses feasibility
tolerance $10^{-10}$, objective tie allowance $10^{-9}$, and explicit
constraint-check tolerance $10^{-8}$. Held-out classifications use
$10^{-10}$ for both the existing and capped policies.

We also evaluate fixed baseline thinning, $.95\pi_0$, independently of
the LP basis. It reduces capture and each complaint rate by the same
factor, so any improvement beyond its capture--gap comparison requires
more than automatic gap reduction from blocking less.
The capped mixture improves that comparison in \ManuscriptNWImprovesUniform{}
of \ManuscriptFWCellTotal{} cells. In
\ManuscriptNWImprovesUniformWithConstraints{} cells, both policies also meet
the held-out group, capture, and value constraints. This comparison uses
the previously examined test rows; it is an exploratory frozen-bank
assessment, not independent confirmation or a population guarantee.

\begin{table*}[t]
\centering
\caption{Capped mixtures at the value-unit $.95$ floor. Caps counts cells
where neither group exceeds its baseline test complaint level; Joint also
requires the test capture floor and blocked-value cap. Retained capture
and gap gain are within-source cell medians. Gap gain subtracts the
mixture's absolute gap from fixed baseline thinning's gap, in native
fraction units. Every source remains in the comparison.}
\label{tab:capped-source-outcomes}
\input{sections/no_worsening_source_table}
\end{table*}

\paragraph{Matched-bank calibration protocol.}
The corrected study uses identical components in the constrained and
unconstrained variants of each common or common-plus-signed bank. An
all-pass component makes uniform baseline thinning available in both.
The relative validation group caps use margins $0$, $.02$, $.05$, $.10$,
$.20$, and $.35$; infeasible margins retain a flagged baseline fallback.
We keep all cells and use the same C acceptance conditions for every arm.

The C/T assignment uses immutable source/account identifiers where
available and verified within-seed ordinals otherwise. It precedes numeric
outcomes and is shared across all floors, caps, margins, and aligned
model variants. C accepts only the already selected feasible proposal,
using both group-rate checks, the capture floor, and the value cap.
Rejection retains the exact baseline; its budget feasibility is measured,
not assumed. No C statistic changes the policy coefficients or thresholds.
The acceptance freeze precedes T outcomes.

On T we evaluate the frozen proposal and the applied-or-baseline policy
on the same rows. Every selector and cell remains in the denominator.
The complete original test split is a separate proposal diagnostic and
must not be confused with this final-partition comparison.

\begin{table*}[t]
\centering
\caption{All declared value-unit margins at the $.95$ floor. $V$ counts feasible
proposals, $C$ acceptance, $N$ non-increase, and $J$ joint targets;
$P$ denotes the proposal and $E$ the applied policy. $U_E$ counts
qualified gains over uniform thinning. $P$ includes exact-baseline returns
for validation-infeasible arms; $E$ also includes C-rejection fallback.
Higher non-increase with fewer accepted proposals
does not establish larger useful gains. Every row retains all cells.}
\label{tab:corrected-margin-coverage}
\input{sections/corrected_margin_table}
\end{table*}

\begin{table*}[t]
\centering
\caption{Source-specific zero-margin capped outcomes on final T.
$C$ counts accepted proposals, $N$ non-increase, $J$ joint targets, and
$U_E$ qualified gains beyond uniform thinning. Subscripts $P$ and $E$
denote proposal and applied policies. All sources remain in the study.}
\label{tab:corrected-source-calibration}
\input{sections/corrected_source_table}
\end{table*}

\paragraph{Earlier matched selector controls.}
These controls minimize validation gap under the capture floor and group caps,
then maximize capture among gap-optimal policies. Their matched bases and
separate calibration provide the reference for direct target selection.

\paragraph{Matched banks and separate calibration.}
We isolate the group-cap effect within each common or common-plus-signed
bank, adding an explicit all-pass component to both. We retain all
\ManuscriptCorrectedCells{} cells at each of the four capture floors and all
declared relative-margin arms. Validation choices precede a separate C
check; acceptance or exact-baseline fallback is frozen before final T
outcomes. Source identities keep repeated accounts together where available;
legacy inputs use disclosed within-seed row partitions. The reference
rules, score coefficients, thresholds, and thinning remain fixed.

At the $.95$ floor, adding group caps within the union bank increases
proposal non-increase on T from
\ManuscriptCorrectedUnionGapTProposalNoIncrease{} to
\ManuscriptCorrectedUnionCapsTProposalNoIncrease{} cells. The common-bank
zero-margin summaries coincide. The C gate accepts
\ManuscriptCorrectedUnionCapsCAccepted{} union-cap proposals and retains
baseline otherwise; after that fallback,
\ManuscriptCorrectedUnionCapsTAppliedNoIncrease{} cells show neither group
increasing. Qualified gains beyond uniform thinning survive in
\ManuscriptCorrectedUnionCapsTAppliedQualifiedGain{} union-cap cells and
\ManuscriptCorrectedCommonCapsTAppliedQualifiedGain{} common-cap cells.
Thus higher applied non-increase must be read with acceptance coverage
and useful gains, not as a confidence guarantee.

\begin{table*}[t]
\centering
\caption{Corrected value-unit matched comparisons at the $.95$ floor, retaining all
\ManuscriptCorrectedCells{} cells. $V$ counts feasible validation proposals,
$C$ accepted proposals, $N$ non-increase in both group rates, and $J$ joint
group, capture, and value targets. Subscripts $P$ and $E$ denote proposals
and applied policies on the same T rows. $P$ uses baseline if validation
selection is infeasible; $E$ also uses baseline after C rejection.
$U_E$ counts smaller gaps than uniform thinning without lower capture,
with joint targets met by both policies. These are descriptive counts.}
\label{tab:corrected-matched-calibration}
\input{sections/corrected_primary_table}
\end{table*}

Both C and T use previously examined data. C evaluates paired empirical
means; it does not provide a probability bound or independent replication.
Appendix~\ref{app:normalized-prices} gives all margins and source outcomes.

\paragraph{Earlier comparison settings.}
Account-level frontiers use Australian MTPL, Belgian MTPL, and KuaiRand.
Group evaluation uses seven independent source families (180 cohort--seed cells),
with an expanded ten-source sensitivity (240 cells), adding Centers for
Medicare \& Medicaid Services (CMS) provider fraud, University of California,
Irvine (UCI) credit, and German credit. A cell is one fitted
cohort and seed; repeated cohorts and French portfolio panels are pooled within
their source. We average source-median paired effects for source-level inference.
Policies share a 10\% value budget. Predictors are fitted on training rows and
held fixed during pricing; thresholds and prices use validation rows, and
selected policies are evaluated on held-out test rows. The primary capture
floor is $.95$ of unpriced W-BCE capture. Appendix~\ref{app:exp-protocol}
gives datasets, outcome constructions, sample sizes, groups, splits, learners,
and preprocessing; the bank-selection protocol below gives the comparison.

\paragraph{Penalty families.}
Proposition~\ref{prop:complaint-disparity-frontier} derives signed group
coefficients from an explicit complaint-disparity constraint. A common
penalty targets aggregate objections. Both adjustments can change which
candidates cross the blocking threshold, and therefore both can change
group complaint levels and their difference. Their held-out ordering
depends on the outcome distribution and fitted policy families.
We compare them using the same value cap, validation capture requirement,
penalty normalization, and coverage procedure.

\paragraph{Protocol.}
For each split, every family must clear the same validation capture floor;
each family then selects the feasible policy with the smallest validation
complaint gap and is scored once on held-out test rows. The main ablation is
within-method: a group-blind account-price family uses one complaint price
for all accounts, while the group-aware family uses an opposite signed group
contrast. The common price is normalized by pooled validation mass, while the
signed group contrast uses each group's validation mass. Both families search a
common axis that retains every original physical policy and extends
geometrically using validation only. We require no selected policy at the
largest price and two further extensions with unchanged selections.
Appendix~\ref{app:normalized-prices} gives the formulas, preservation mapping,
coverage rule, and historical results.

Both price families use the fitted W-BCE implied loss side; auxiliary
regressions construct objections and measurement weights.
Appendix~\ref{app:normalized-prices} specifies their scores. We retain the
$.95$ validation capture floor as the primary operating point and report the
full $.90$--$1.00$ grid.

Before selection, let $v_{\rm val}$ be a policy's blocked value divided by
total validation value. We apply independent thinning with probability
$\alpha=\min(1,.10/v_{\rm val})$, taking $\alpha=1$ when
$v_{\rm val}=0$, to every family and the unpriced baseline.
Capture floors use the corrected baseline. We freeze this probability,
the selected penalties, and thresholds on test, and report expected metrics
under thinning.

Let $\pi_0$ be unpriced W-BCE and $L_P=E_{\rm test}[\pi_P B]$ the
loss capture of policy $P$, including frozen thinning; $L_0=E_{\rm test}[\pi_0B]$.
For test complaint gaps $d_P=|\Delta_D^w(\pi_P)|$ and $d_0=|\Delta_D^w(\pi_0)|$,
closure is $C_P=1-d_P/d_0$. Lift is
$C_{\rm group}-C_{\rm comparator}$, the selected group-aware policy's
closure minus the comparator's closure, without clipping. When $d_0=0$,
closure is one if $d_P=0$ and zero otherwise; relative improvement is
undefined in these cells and the cell is dropped.

\paragraph{Attainable complaint gaps.}
As an outcome-informed finite-sample diagnostic, two linear programs per
cohort, unit, and seed minimize and maximize the signed complaint gap on
the frozen test rows. They use realized value and loss, allow independent
row decisions, and use the baseline's realized test spend and a $.95$
capture floor. These intervals describe an optimistic sample relaxation;
they do not establish population feasibility or attainable performance of
a learned policy. The reference masks remain fixed at their original
validation operating point. A second pair adds
$D_g^w(\pi)\leq D_g^w(\pi_0)$ for both groups, forbidding a gap reduction
that raises either group's objection level.

\begin{table}[t]
\centering
\caption{Attainable signed complaint gaps at the $.95$ capture floor over
\ManuscriptAttainTotalCells{} cohort, unit and seed cells. Parity feasible counts
cells whose attainable interval contains zero; no-worse-off repeats that count
after requiring $D_g^w(\pi)\leq D_g^w(\pi_0)$ for both groups. Width is the
median interval length. These are outcome-informed descriptive bounds on
the reported eligible frozen rows, using each baseline's realized test spend.}
\label{tab:attainable-disparity}
\input{sections/attainable_disparity_table}
\end{table}

Table~\ref{tab:attainable-disparity} reports both. The optimistic rowwise
relaxation contains parity in every one of its
\ManuscriptAttainTotalCells{} reported eligible cells. This result does not
show that feasibility is nonbinding for the validation-selected policy
families or for population conditional means.
Forbidding either group to worsen changes that only in the protected-loss
unit, which retains parity in \ManuscriptAttainLossNoWorseOff{} of
\ManuscriptAttainLossCells{} cells and loses a median
\ManuscriptAttainBindingShare{}\% of the baseline gap where it binds, while
blocked value retains parity in all \ManuscriptAttainValueCells{}.
Proposition~\ref{prop:unit-score-information} characterizes the additional
information needed by the adjusted score after the reference indicators
and normalizers are fixed. It does not identify a uniquely appropriate
unit or establish that these hindsight row decisions can be served from
the utility score.

\begin{figure*}[t]
\centering
\includegraphics[width=\textwidth]{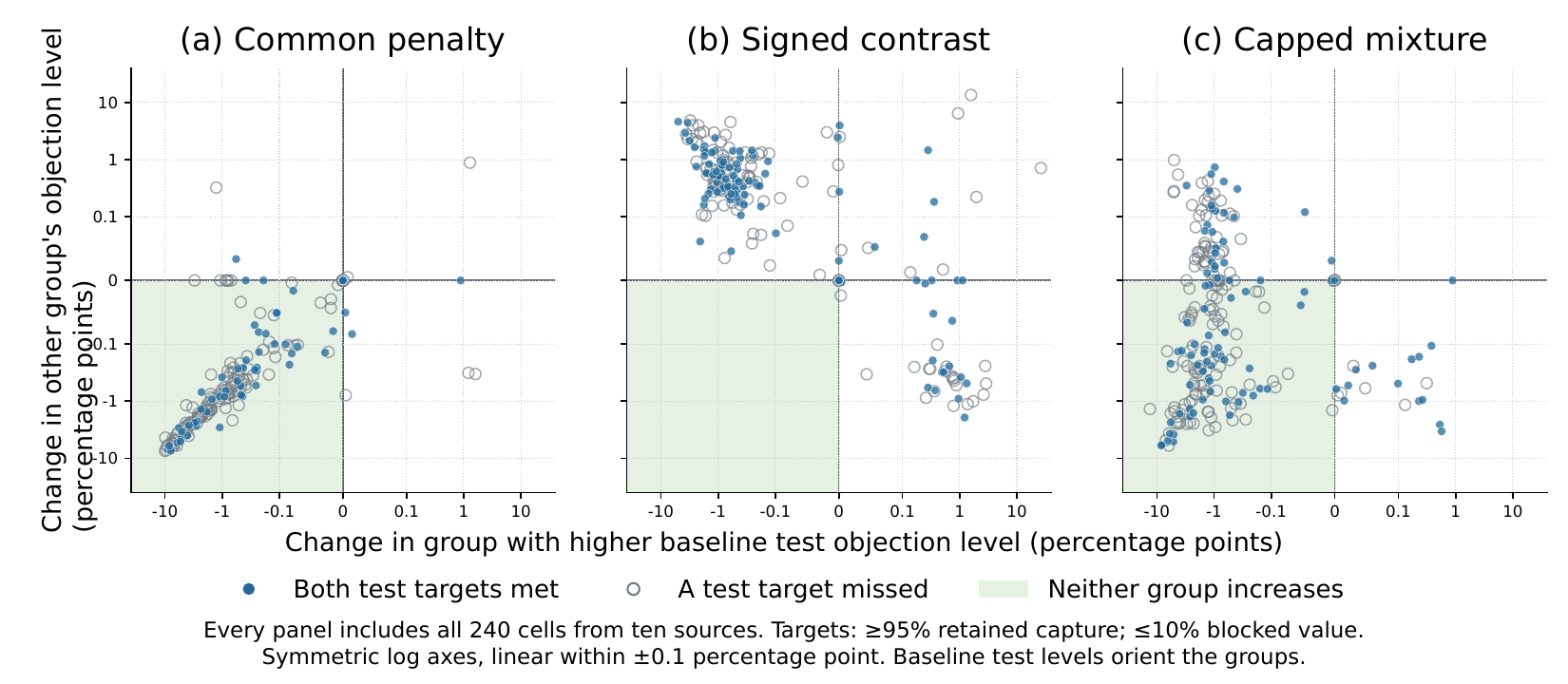}
\caption{Expected value-unit complaint changes relative to unpriced W-BCE.
Each point is a fitted cohort--seed cell. Axes orient the group with higher
baseline test complaints against the other group. Green shading marks
non-increase in both rates; filled points meet the test capture floor and
blocked-value cap, and hollow points miss a target. Every cell remains
visible on shared symmetric-log axes, linear within $\pm.1$ percentage point.}
\Description{Three panels compare a common penalty, a signed group contrast,
and a capped mixture. The signed contrasts commonly reduce one group's
complaints while increasing the other's. The other families more often
reduce both rates, and hollow markers retain the failed business targets.}
\label{fig:group-outcome-changes}
\end{figure*}

\paragraph{Group outcomes and rate caps.}
A smaller complaint gap can accompany higher objections in another group.
We reconstruct each group's outcomes from the frozen candidates and
selected components. At the primary value-unit $.95$ floor, signed
contrasts reduce the held-out gap in \ManuscriptFWSignedGapReduced{} of
\ManuscriptFWCellTotal{} cells; \ManuscriptFWSignedGapReducedWithIncrease{} of those
reductions increase at least one group's complaint level.
Figure~\ref{fig:group-outcome-changes} shows both changes, and
Table~\ref{tab:group-outcome-constraints} separates these reductions from
ones that preserve both group levels and the business constraints.

\begin{table}[t]
\centering
\caption{Value-unit outcomes at the $.95$ validation floor over
\ManuscriptFWCellTotal{} cells. $G$ counts smaller test gaps, $N$ the subset
without either group increasing, and $F$ the subset also meeting test
capture and value constraints. Counts are descriptive and include
repeated cohorts and seeds. Baseline thinning multiplies $\pi_0$ by $.95$.}
\label{tab:group-outcome-constraints}
\input{sections/fairness_group_outcomes_table}
\end{table}

We also minimize validation gap over mixtures of both frozen banks,
adding $D_g^A(\pi)\leq D_g^A(\pi_0)$ for each group to the original
capture and value constraints. The selected mixture retains a median
\ManuscriptNWMedianRetainedPercent{}\% of baseline test capture;
\ManuscriptNWAllConstraintsMet{} of \ManuscriptFWCellTotal{} cells meet all
three test constraints. It improves the capture--gap comparison over
fixed baseline thinning in \ManuscriptNWImprovesUniform{} cells, including
\ManuscriptNWImprovesUniformWithConstraints{} where both policies meet the
test constraints. These post-review outcomes retain every source and
do not constitute independent replication.
Appendix~\ref{app:normalized-prices} gives the selection problem and
source-specific results.

\paragraph{Normalized-bank comparison.}
We report each family's outcome after validation-only selection. The
per-source results and all three measurement units retain the variation
in held-out effects. This selected-policy comparison does not characterize
all attainable points of either family. The frozen bank and its LP
mixtures define the empirical comparison scope.

\begin{table}[t]
\centering
\caption{Validation-selected blocked-value comparison on the normalized,
extended bank. Each mean averages source medians of paired held-out closure
lift. The seven-source scope and all ten sources are reported separately;
$p$ is a two-sided source $t$ test.}
\label{tab:group-complaint-selected}
\input{sections/normalized_headline_table}
\end{table}

At the $.95$ floor, mean source-median lift is $+.2185$ across seven sources
(two-sided $p=.3071$) and $+.0863$ across ten ($p=.6007$), so this bank does
not establish an average group-aware advantage. Raw cell means are $-.2274$
and $-.5592$; these measure different effects, and the gap between them
reflects small-gap cells in which closure divides by a near-zero baseline.
Count-unit effects are mixed, and protected-loss average lifts are negative at
every floor on this bank. Linear-program mixtures of the full group-blind bank
give a mean ten-source median lift of $-.0521$ ($p=.7339$)
(Table~\ref{tab:matched-complaint-lp}). Fixed block-, value-, and loss-parity
postprocessors constrain quantities other than complaint disparity and are
reported as diagnostics. Appendix~\ref{app:normalized-prices} gives the bank
construction, the aggregation and floor sensitivities, and the historical
truncated-bank result.

\input{sections/normalized_source_lifts}

%% file: sections/normalized_lp_table.tex
\begin{tabular}{lrrr}
\toprule
Source & Group gap & LP gap & Lift \\
\midrule
Australian MTPL & 0.0043 & 0.0132 & +0.3291 \\
Belgian MTPL & 0.0019 & 0.0055 & +0.1903 \\
Bondora & 0.0276 & 0.0508 & +0.4322 \\
CMS provider fraud & 0.0264 & 0.0093 & -0.7152 \\
EU health & 0.0026 & 0.0068 & +0.3010 \\
French portfolio & 0.0079 & 0.0060 & -0.0373 \\
German credit & 0.0509 & 0.0323 & -0.2659 \\
Norwegian auto & 0.0097 & 0.0014 & -0.9302 \\
Swedish motorcycle & 0.0058 & 0.0128 & +0.3428 \\
UCI credit & 0.0087 & 0.0070 & -0.1684 \\
\bottomrule
\end{tabular}

%% file: sections/normalized_currency_table.tex
\begin{tabular}{llrrrr}
\toprule
Unit & Floor & Seven medians & Seven means & Ten medians & Ten means \\
\midrule
Value & 0.90 & +0.0391 & -0.1533 & -0.0372 & -0.5979 \\
Value & 0.95 & +0.2185 & +0.0977 & +0.0863 & -0.3979 \\
Value & 0.98 & +0.1969 & +0.1127 & +0.0735 & -0.3605 \\
Value & 1.00 & +0.1093 & +0.0158 & +0.0472 & -0.4358 \\
Count & 0.90 & -0.0514 & -0.2949 & -0.0699 & -0.2966 \\
Count & 0.95 & +0.0995 & -0.2123 & +0.0321 & -0.2080 \\
Count & 0.98 & +0.0580 & -0.2176 & +0.0406 & -0.1419 \\
Count & 1.00 & +0.0892 & -0.1877 & +0.0624 & -0.1058 \\
Loss & 0.90 & -0.1702 & -0.6889 & -0.2130 & -0.9735 \\
Loss & 0.95 & -0.0375 & -0.4740 & -0.1121 & -0.7918 \\
Loss & 0.98 & -0.0748 & -0.3661 & -0.0758 & -0.6242 \\
Loss & 1.00 & -0.1367 & -0.4901 & -0.1021 & -0.3450 \\
\bottomrule
\end{tabular}

%% file: sections/no_worsening_source_table.tex
\begin{tabular}{lrrrrr}
\toprule
Source & Cells & Caps & Joint & Retained capture (\%) & Gap gain \\
\midrule
Australian MTPL & 20 & 11 & 5 & 97.88 & +0.0224 \\
Belgian MTPL & 30 & 13 & 3 & 97.77 & +0.0092 \\
Bondora & 10 & 6 & 1 & 90.32 & +0.0122 \\
CMS provider fraud & 20 & 15 & 8 & 101.18 & -0.0043 \\
EU health insurance & 10 & 7 & 0 & 93.59 & +0.0079 \\
French portfolio & 90 & 61 & 23 & 99.32 & +0.0037 \\
German credit & 20 & 11 & 3 & 94.57 & +0.0021 \\
Norwegian auto & 10 & 10 & 3 & 95.40 & +0.0062 \\
Swedish motorcycle & 10 & 6 & 3 & 99.37 & +0.0107 \\
UCI credit & 20 & 11 & 4 & 98.56 & -0.0019 \\
\bottomrule
\end{tabular}

%% file: sections/corrected_margin_table.tex
\begin{tabular}{llrrrrrrr}
\toprule
Basis & $m$ & $V$ & $C$ & $N_P$ & $N_E$ & $J_P$ & $J_E$ & $U_E$ \\
\midrule
Common & 0\% & 240 & 75 & 231 & 238 & 80 & 125 & 22 \\
Common & 2\% & 240 & 76 & 234 & 240 & 82 & 126 & 22 \\
Common & 5\% & 240 & 74 & 235 & 240 & 77 & 123 & 25 \\
Common & 10\% & 211 & 53 & 234 & 239 & 74 & 119 & 19 \\
Common & 20\% & 164 & 32 & 236 & 240 & 82 & 120 & 13 \\
Common & 35\% & 118 & 24 & 236 & 240 & 90 & 124 & 12 \\
Common + signed & 0\% & 240 & 42 & 144 & 229 & 53 & 119 & 10 \\
Common + signed & 2\% & 240 & 52 & 187 & 231 & 67 & 118 & 11 \\
Common + signed & 5\% & 240 & 63 & 214 & 236 & 79 & 124 & 19 \\
Common + signed & 10\% & 222 & 60 & 223 & 236 & 81 & 120 & 21 \\
Common + signed & 20\% & 170 & 38 & 232 & 239 & 85 & 121 & 12 \\
Common + signed & 35\% & 123 & 24 & 235 & 239 & 90 & 123 & 9 \\
\bottomrule
\end{tabular}

%% file: sections/corrected_source_table.tex
\begin{tabular}{llrrrrrr}
\toprule
Source & Basis & Cells & $C$ & $N_P$ & $N_E$ & $J_E$ & $U_E$ \\
\midrule
Australian MTPL & Common & 20 & 4 & 20 & 20 & 10 & 1 \\
Australian MTPL & Common + signed & 20 & 5 & 13 & 19 & 10 & 2 \\
Belgian MTPL & Common & 30 & 5 & 30 & 30 & 11 & 0 \\
Belgian MTPL & Common + signed & 30 & 2 & 11 & 29 & 10 & 0 \\
Bondora & Common & 10 & 3 & 10 & 10 & 4 & 1 \\
Bondora & Common + signed & 10 & 2 & 4 & 9 & 3 & 0 \\
CMS provider fraud & Common & 20 & 8 & 19 & 20 & 9 & 7 \\
CMS provider fraud & Common + signed & 20 & 6 & 14 & 18 & 8 & 2 \\
EU health insurance & Common & 10 & 1 & 10 & 10 & 6 & 0 \\
EU health insurance & Common + signed & 10 & 0 & 5 & 10 & 7 & 0 \\
French portfolio & Common & 90 & 32 & 90 & 90 & 49 & 6 \\
French portfolio & Common + signed & 90 & 14 & 60 & 87 & 50 & 2 \\
German credit & Common & 20 & 3 & 13 & 19 & 14 & 0 \\
German credit & Common + signed & 20 & 2 & 10 & 19 & 14 & 0 \\
Norwegian auto & Common & 10 & 4 & 10 & 10 & 6 & 2 \\
Norwegian auto & Common + signed & 10 & 4 & 10 & 10 & 6 & 2 \\
Swedish motorcycle & Common & 10 & 4 & 10 & 10 & 4 & 2 \\
Swedish motorcycle & Common + signed & 10 & 2 & 6 & 10 & 3 & 1 \\
UCI credit & Common & 20 & 11 & 19 & 19 & 12 & 3 \\
UCI credit & Common + signed & 20 & 5 & 11 & 18 & 8 & 1 \\
\bottomrule
\end{tabular}

%% file: sections/corrected_primary_table.tex
\begin{tabular}{llrrrrrrr}
\toprule
Basis & Caps & $V$ & $C$ & $N_P$ & $N_E$ & $J_P$ & $J_E$ & $U_E$ \\
\midrule
Common & Gap only & 240 & 75 & 231 & 238 & 80 & 125 & 22 \\
Common & Both group caps & 240 & 75 & 231 & 238 & 80 & 125 & 22 \\
Common + signed & Gap only & 240 & 15 & 60 & 238 & 19 & 127 & 4 \\
Common + signed & Both group caps & 240 & 42 & 144 & 229 & 53 & 119 & 10 \\
\bottomrule
\end{tabular}

%% file: sections/attainable_disparity_table.tex
\begin{tabular}{lrrrr}
\toprule
Unit & Cells & Parity feasible & No-worse-off & Median width \\
\midrule
Candidate count & 150 & 150/150 & 146/150 & 1.016 \\
Blocked value & 150 & 150/150 & 150/150 & 0.484 \\
Protected loss & 150 & 150/150 & 90/150 & 1.752 \\
\bottomrule
\end{tabular}

%% file: sections/fairness_group_outcomes_table.tex
\begin{tabular}{lrrr}
\toprule
Policy & $G$ & $N$ & $F$ \\
\midrule
Common penalty & 148 & 146 & 37 \\
Signed contrast & 151 & 1 & 1 \\
Common mixture & 189 & 187 & 48 \\
Capped mixture & 174 & 116 & 40 \\
Baseline thinning & 240 & 240 & 193 \\
\bottomrule
\end{tabular}

%% file: sections/normalized_headline_table.tex
\begin{tabular}{rrrrr}
\toprule
Floor & Seven mean & $p$ & Ten mean & $p$ \\
\midrule
0.90 & +0.0391 & 0.8426 & -0.0372 & 0.8008 \\
0.95 & +0.2185 & 0.3071 & +0.0863 & 0.6007 \\
0.98 & +0.1969 & 0.3214 & +0.0735 & 0.6333 \\
1.00 & +0.1093 & 0.5482 & +0.0472 & 0.7127 \\
\bottomrule
\end{tabular}

%% file: sections/normalized_source_lifts.tex
\begin{table}[t]
\centering
\caption{Corrected value-unit group-aware versus group-blind complaint
closure lift at the $.95$ capture floor. Each row is one source's median
paired held-out lift. Negative and zero effects are retained.}
\label{tab:normalized-source-lifts}
\begin{tabular}{lr}
\toprule
Source & Median lift \\
\midrule
Australian MTPL & +0.5393 \\
Belgian MTPL & +0.4448 \\
Bondora & +0.4892 \\
CMS provider fraud & -0.6403 \\
EU health insurance & +0.4899 \\
French portfolio & -0.0027 \\
German credit & +0.0000 \\
Norwegian auto & -0.8800 \\
Swedish motorcycle & +0.4489 \\
UCI credit & -0.0257 \\
\bottomrule
\end{tabular}
\end{table}

%% file: sections/direct_selector_details.tex
\section{Direct Selector Details}
\label{app:direct-selector-details}

\paragraph{Comparison objectives.}
\input{sections/direct_selector_method}

Fix conditional outcome means $a_g(\mathbf x)=E[A\mid\mathbf X=\mathbf x,G=g]$
and $b_g(\mathbf x)=E[B\mid\mathbf X=\mathbf x,G=g]$, a binary reference
indicator $q_{\mathcal C}$, and a conditional unit coefficient $u_w(\mathbf x,g)$.
Let $P_g=\Pr(G=g)$ and $U_g^w=E[u_w(\mathbf X,g)\mid G=g]$.
The group rate is
$D_g^w(\pi)=E[\pi q_{\mathcal C}u_w\mid G=g]/U_g^w$.
We maximize capture subject to the value budget, a gap tolerance, and
predeclared group caps. Attach nonnegative multipliers $\mu$, $\alpha_+$,
$\alpha_-$, $\eta_0$, and $\eta_1$ to those five inequalities. Set
$\nu=\alpha_+-\alpha_-$ and $\sigma_0=-1,\ \sigma_1=1$.

\begin{proposition}[Direct group-capped selector]
\label{prop:direct-group-capped-selector}
Fix the reference indicator and measurement coefficients. Assume
integrability, positive group probabilities and finite positive group
denominators, no duality gap, and attainment by finite multipliers.
An optimal randomized capture-maximizing policy under the value budget,
gap tolerance, and group caps blocks where
\begin{equation}
 b_g(\mathbf{x})-\mu a_g(\mathbf{x})
 -\frac{(\eta_g+\sigma_g\nu)q_{\mathcal C}(\mathbf{x})
         u_w(\mathbf{x},g)}{P_gU_g^w}>0,
 \label{eq:direct-group-capped-coefficient}
\end{equation}
passes where this coefficient is negative, and randomizes on its zero set
to satisfy the active constraints.
\end{proposition}

\begin{proof}
Conditioning on $(\mathbf X,G)=(\mathbf{x},g)$ gives the coefficient
on the left of Eq.~\eqref{eq:direct-group-capped-coefficient}.
Each conditional group constraint contributes its coefficient divided
by $P_g$, while the two gap multipliers contribute the signed term
$\sigma_g\nu$. Maximizing the resulting linear function over $[0,1]$
gives the stated policy.
\end{proof}

When $\eta_0=\eta_1=0$, the two group coefficients reduce to the signed
gap contrast. Active group caps add separate coefficients to that contrast.

\paragraph{Normalized dual recovery.}
Fix validation coefficients $a_i=\widehat a_i\geq0$,
$r_i=\exp(\operatorname{clip}(s_{\theta,i},-50,50))$,
$b_i=a_i r_i$, binary reference indicators $q_i$, and $G_i\in\{0,1\}$.
Here $s_{\theta,i}$ is the fitted classifier logit.
Let $\pi_0$ be the fixed reference block-probability vector.
Let $A_\Sigma=\sum_i a_i$, $B_\Sigma=\sum_i b_i$, and
$A_g=\sum_{i:G_i=g}a_i$ be positive and finite. Define
\[
 c_{g,i}=\frac{q_i a_i\mathbf1(G_i=g)}{A_g},
 \qquad D_g(x)=c_g^\top x,\qquad
 \ell_g=(1-m_g)D_g(\pi_0).
\]
The block probabilities $x_i\in[0,1]$ solve the minimization LP
\[
\begin{aligned}
 \min_x\quad&-b^\top x/B_\Sigma\\
 \text{s.t.}\quad&a^\top x/A_\Sigma\leq v,\\
 &(c_1-c_0)^\top x\leq\epsilon,\quad
 (c_0-c_1)^\top x\leq\epsilon,\\
 &c_g^\top x\leq\ell_g\quad(g=0,1).
\end{aligned}
\]
There is no simplex equality on these block probabilities.
For $\epsilon\geq0$, $v\geq0$, and $m_g\in[0,1]$, all-pass is feasible.
For numerical stability, the implementation multiplies the objective,
all five inequality rows, and their right-hand sides by the same positive
factor. This scaling preserves the primal solution and inequality multipliers;
we undo it when checking the normalized objective and bound marginals.

For these exact row scalings, negate the minimization solver's inequality
RHS marginals to obtain
$(\lambda_v,\alpha_+,\alpha_-,\eta_0,\eta_1)\geq0$, in the row order above.
Set $\nu=\alpha_+-\alpha_-$ and recover
\begin{equation}
 \tau=\lambda_v\frac{B_\Sigma}{A_\Sigma},\qquad
 \vartheta_0=(\eta_0-\nu)\frac{B_\Sigma}{A_0},\qquad
 \vartheta_1=(\eta_1+\nu)\frac{B_\Sigma}{A_1}.
 \label{eq:direct-normalized-dual}
\end{equation}
The maximization reduced coefficient equals
\[
 \frac{b_i}{B_\Sigma}
 -\lambda_v\frac{a_i}{A_\Sigma}
 -(\eta_0-\nu)c_{0,i}-(\eta_1+\nu)c_{1,i}
 =\frac{a_i}{B_\Sigma}(r_i-\tau-q_i\vartheta_{G_i}).
\]
At an optimal primal--dual pair, complementary slackness gives
\[
\begin{aligned}
 \lambda_v(v-a^\top x/A_\Sigma)&=0,\\
 \alpha_+(\epsilon-D_1(x)+D_0(x))&=0,\\
 \alpha_-(\epsilon+D_1(x)-D_0(x))&=0,\\
 \eta_g(\ell_g-D_g(x))&=0\quad(g=0,1).
\end{aligned}
\]
We check these relations and the reduced coefficients using the actual
LP row scalings. The budget RHS marginal alone is not the odds threshold.

\paragraph{Portable boundary probabilities.}
For exact fitted boundary strata, define
\[
 T_{gq}=\{i:G_i=g,\ q_i=q,\ a_i>0,\
                  r_i-\tau-q_i\vartheta_g=0\}.
\]
For positive stratum value, set
\[
 p_{gq}=\frac{\sum_{i\in T_{gq}}a_i x_i}
              {\sum_{i\in T_{gq}}a_i}.
\]
We set $p_{gq}=0$ when the stratum has no validation value, fixing the
convention for new candidates on an unseen boundary.

\begin{lemma}[Value-weighted boundary replacement]
\label{lem:direct-portable-ties}
Let $x$ and the multipliers above form an optimal fitted primal--dual pair.
Replace $x_i$ by $p_{gq}$ on each $T_{gq}$, retain its decisions
outside these strata, and set $x_i=0$ where $a_i=0$.
The resulting threshold policy preserves the fitted objective, budget,
and both group levels, and remains optimal.
\end{lemma}

\begin{proof}
Within $T_{gq}$, the ratio is the constant
$r_i=\tau+q\vartheta_g$. Thus $b_i=a_i(\tau+q\vartheta_g)$, and
the objective, budget, and each group-level coefficient are constant
multiples of $a_i$ within that stratum. The definition of $p_{gq}$
preserves $\sum_{i\in T_{gq}}a_i x_i$, and therefore preserves all these
quantities. Off the boundary, primal--dual optimality already fixes $x_i$
to the threshold decision. At $a_i=0$, both $b_i$ and complaint
coefficients vanish, so passing changes none of the fitted quantities.
\end{proof}

We freeze the boundary probabilities and the three score parameters
before evaluating new candidates. With no group-cap multipliers,
the offsets form a mass-normalized signed contrast. With active caps,
they can contain both a common adjustment and a signed contrast;
this score representation does not establish equivalence to a finite
bank of separate policies or its mixtures. Parameter count alone does
not establish a generalization guarantee.

\paragraph{Numeric replay and target scope.}
In the implementation, each group/indicator stratum has a validation-fitted
ratio band around $t_{gq}=\tau+q\vartheta_g$. We choose its width to cover
material fractional primal points and numeric conflicts between the primal
bound state and the score sign, with a fixed minimum numeric tolerance.
Within that band, we use the value-weighted probability; outside it, we use
the score decision. We freeze all four widths before evaluation. They do not
depend on a new candidate's positive predicted value. Because the budget and
group coefficients are proportional to value within each stratum, band
averaging preserves those amounts. Varying ratios within a numeric band can
change capture, so we check it independently.

We fix numeric boundary tolerances before evaluation and verify the
serialized rule's original budget, gap, group caps, and fitted capture.
Let $p=b/B_\Sigma$, and write the five normalized inequalities as
$Mx\leq h$. Any nonnegative multiplier vector $\lambda$ gives the bound
\[
 p^\top x\leq
 U(\lambda)=\lambda^\top h+
 \sum_i[p_i-(M^\top\lambda)_i]_+
 \qquad(0\leq x\leq1,\ Mx\leq h).
\]
This follows by maximizing each term of
$p^\top x=\lambda^\top Mx+(p-M^\top\lambda)^\top x$ over $[0,1]$.
We require feasibility within its numeric tolerance and a two-sided
numeric check of $U(\lambda)-p^\top\widehat\pi$. We keep differences from
the solver's primal amounts as diagnostics; distinct nearly optimal
allocations can differ on inactive constraints. The exact boundary lemma
still applies, while approximate ties require the feasibility and objective
check. Additional thinning of the direct rule changes its attained objective.
The optimization has no capture floor. If its fitted maximum falls below
a prespecified fraction of baseline fitted capture, that floor is
infeasible under these fitted constraints. This conclusion does not
establish population infeasibility. We report the reference baseline's
fitted spend separately. Observed group, gap, capture, and value targets
require separate evaluation; baseline fallback does not generally
satisfy a stricter gap target or the observed value budget.

\paragraph{Fitted-coefficient and bank controls.}
The controls maximize capture at the same declared gap fraction and group
margin. The bank controls use observed validation amounts; the fitted dual
uses predicted value and W-BCE-implied loss. Their absolute target estimates
and policy classes differ. We retain every original cohort--seed cell.

\begin{table*}[t]
\centering\small
\caption{Ungated maximum-capture controls at the half-gap target and zero
margin over \ManuscriptDirectCells{} original-test cells. $G$ counts smaller gaps,
$N$ also requires no group increase, and $F$ also requires the capture and
value targets. $N_{all}$ and $J$ omit the gap-reduction requirement. $U$ denotes
joint-qualified gains over uniform thinning. $Q_{rel}$ and $Q_{fix}$ distinguish
relative held-out targets from frozen validation limits; $K$ counts fitted
capture-floor feasibility. Retained capture is the test median.}
\label{tab:direct-control-outcomes}
\input{sections/direct_selector_primary_table}
\end{table*}

\begin{table*}[t]
\centering\small
\caption{Separate T outcomes for the same controls. $C$ counts proposals
accepted by the empirical group/capture/value gate; subscripts $P$ and $E$
identify proposed and accepted-or-baseline policies. $N$ counts group
non-increase, $J$ joint targets, and $U$ qualified gains over thinning.
C acceptance does not approve every gap target.}
\label{tab:direct-control-calibration}
\input{sections/direct_selector_calibration_table}
\end{table*}

\begin{table*}[t]
\centering\scriptsize
\caption{Prespecified gap fractions and group margins for the fitted-dual
and frozen-bank controls. All cells and fallback statuses remain in each arm.}
\label{tab:direct-control-sensitivity}
\input{sections/direct_selector_sensitivity_table}
\end{table*}

%% file: sections/direct_selector_method.tex
We now constrain both group levels while maximizing capture. Fix the
reference policy $\pi_0$, a gap tolerance $\epsilon\geq0$, and
predeclared margins $m_g\in[0,1]$. Write
$d_g^{\max}=(1-m_g)D_g^w(\pi_0)$. The direct selector solves
\begin{equation}
\begin{aligned}
 \max_{0\leq\pi\leq1}\quad&E[\pi B]\\
 \text{s.t.}\quad&E[\pi A]\leq\beta,\qquad
 |D_1^w(\pi)-D_0^w(\pi)|\leq\epsilon,\\
 &D_g^w(\pi)\leq d_g^{\max}\quad(g=0,1).
\end{aligned}
\label{eq:direct-group-capped-objective}
\end{equation}
This objective maximizes capture at a specified complaint target. It has
no minimum-capture constraint; the attained maximum determines whether
an additional capture requirement is feasible for those coefficients.

Group caps add nonnegative multipliers to the signed complaint contrast
in Proposition~\ref{prop:complaint-disparity-frontier}.
Proposition~\ref{prop:direct-group-capped-selector} in
Appendix~\ref{app:direct-selector-details} gives the resulting coefficient.
The direct fit optimizes a candidate-level allocation and recovers a
threshold rule.

\input{sections/empirical_cut_method}

%% file: sections/empirical_cut_method.tex
For value-unit objections, we learn constrained score cuts directly
from observed validation amounts. We keep the yield score $r_i$ and the
criterion indicator $q_i$ fixed, and let $a_i,b_i\geq0$ be observed value
and loss. We partition candidates into three buckets,
\[
 \mathcal I_\star=\{i:q_i=0\},\qquad
 \mathcal I_g=\{i:q_i=1,\ G_i=g\}\quad(g=0,1).
\]
The nonobjecting bucket pools both groups. In each bucket, we evaluate
all distinct score cuts, including all-pass and full-block endpoints.
A cut blocks the whole equal-score batch. Its validation point
$(C_{kj},L_{kj})$ gives blocked value and captured loss.

We take the upper concave hull of these points over the bucket's full
cost interval. At equal cost we retain the highest capture, while keeping
the endpoint needed to represent a flat interval. Let $\mathcal H_k$
index the hull vertices and
$C_k(\omega)=\sum_{j\in\mathcal H_k}\omega_{kj}C_{kj}$.
The empirical selector solves
\begin{equation}
\begin{aligned}
 \max_{\omega\geq0}\quad&
       \sum_k\sum_{j\in\mathcal H_k}\omega_{kj}L_{kj}\\
 \text{s.t.}\quad&
       \sum_{j\in\mathcal H_k}\omega_{kj}=1\quad(k=\star,0,1),\\
 &\sum_k C_k(\omega)\leq vA_\Sigma,\\
 &\left|C_1(\omega)/A_1-C_0(\omega)/A_0\right|\leq\epsilon,\\
 &C_g(\omega)/A_g\leq(1-m_g)D_g^A(\pi_0)\quad(g=0,1),
\end{aligned}
\label{eq:empirical-cut-objective}
\end{equation}
where $A_\Sigma=\sum_i a_i$ and $A_g=\sum_{i:G_i=g}a_i>0$.
The denominator includes all group value, including nonobjecting value.
Only the objecting bucket's cost contributes to that group's complaints.

At the selected cost in each bucket, two adjacent hull vertices suffice.
We freeze their score cutoffs $t_k^-,t_k^+$ and a mixture weight
$\lambda_k\in[0,1]$. A new candidate in bucket $k$ has block probability
\[
 \pi_{\rm cut}(\mathbf{x},g)=
 (1-\lambda_k)\mathbf1(r(\mathbf{x},g)\geq t_k^-)
 +\lambda_k\mathbf1(r(\mathbf{x},g)\geq t_k^+).
\]
Degenerate buckets use one cut. The resulting rule uses at most six
cutoffs and three mixture weights, and depends on $r$, $G$, and
$q_{\mathcal C}$ at decision time. Observed validation amounts determine
the cuts and weights; new candidates' realized value and loss do not enter
the decision. Auxiliary models remain inputs to the fixed review criteria.

Proposition~\ref{prop:empirical-cut-optimality} proves optimality within
the convexified bucket score-cut family. The reduction uses the value-unit
identity $D_g^A=C_g/A_g$: at a fixed bucket cost, higher capture changes
none of the fairness constraints. It does not prove optimality over all
candidate policies or population outcomes. We retain all-pass separately
as a control when a zero-cost, positive-capture point dominates it on
validation. We specify the gap and margin targets before evaluation,
and freeze all cuts and weights before the separate C/T checks.

%% file: sections/direct_selector_primary_table.tex
\begin{tabular}{lrrrrrrrrrr}
\toprule
Method & $G$ & $N$ & $F$ & $N_{all}$ & $J$ & $U$ & $Q_{rel}$ & $Q_{fix}$ & $K$ & Retained (\%) \\
\midrule
Common bank & 205 & 203 & 37 & 233 & 45 & 35 & 17 & 20 & 126 & 91.74 \\
Common + signed bank & 175 & 107 & 36 & 141 & 49 & 34 & 20 & 20 & 222 & 99.08 \\
Direct selector & 171 & 72 & 27 & 102 & 36 & 26 & 7 & 16 & 208 & 99.83 \\
\bottomrule
\end{tabular}

%% file: sections/direct_selector_calibration_table.tex
\begin{tabular}{lrrrrrrrrr}
\toprule
Method & $C$ & $N_P$ & $N_E$ & $J_P$ & $J_E$ & $U_P$ & $U_E$ & $Q_{rel,E}$ & $Q_{fix,E}$ \\
\midrule
Common bank & 48 & 232 & 238 & 50 & 121 & 38 & 16 & 8 & 21 \\
Common + signed bank & 45 & 150 & 228 & 63 & 119 & 40 & 15 & 9 & 19 \\
Direct selector & 38 & 80 & 223 & 29 & 124 & 21 & 7 & 0 & 12 \\
\bottomrule
\end{tabular}

%% file: sections/direct_selector_sensitivity_table.tex
\begin{tabular}{lrrrrrrrrrrr}
\toprule
Method & $\gamma$ & $m$ & $V$ & $C$ & $G$ & $F$ & $N_{all}$ & $J$ & $U$ & $Q_{rel}$ & $Q_{fix}$ \\
\midrule
Common bank & 0 & 0 & 240 & 23 & 224 & 21 & 235 & 28 & 18 & 7 & 7 \\
Common bank & 0 & 0.05 & 240 & 23 & 228 & 25 & 236 & 28 & 22 & 7 & 7 \\
Common bank & 0.25 & 0 & 240 & 35 & 213 & 31 & 235 & 40 & 26 & 10 & 16 \\
Common bank & 0.25 & 0.05 & 240 & 35 & 218 & 35 & 236 & 41 & 29 & 10 & 16 \\
Common bank & 0.5 & 0 & 240 & 48 & 205 & 37 & 233 & 45 & 35 & 17 & 20 \\
Common bank & 0.5 & 0.05 & 240 & 48 & 208 & 39 & 234 & 46 & 37 & 16 & 17 \\
Common bank & 0.75 & 0 & 240 & 64 & 196 & 49 & 231 & 64 & 41 & 22 & 26 \\
Common bank & 0.75 & 0.05 & 240 & 64 & 201 & 53 & 233 & 65 & 48 & 22 & 25 \\
Common bank & 1 & 0 & 240 & 92 & 101 & 37 & 231 & 103 & 23 & 88 & 39 \\
Common bank & 1 & 0.05 & 240 & 100 & 186 & 83 & 233 & 106 & 51 & 52 & 38 \\
Common + signed bank & 0 & 0 & 240 & 38 & 174 & 38 & 150 & 49 & 33 & 2 & 2 \\
Common + signed bank & 0 & 0.05 & 240 & 58 & 179 & 55 & 219 & 75 & 51 & 2 & 2 \\
Common + signed bank & 0.25 & 0 & 240 & 40 & 177 & 40 & 145 & 52 & 35 & 10 & 15 \\
Common + signed bank & 0.25 & 0.05 & 240 & 62 & 180 & 54 & 219 & 78 & 50 & 9 & 10 \\
Common + signed bank & 0.5 & 0 & 240 & 45 & 175 & 36 & 141 & 49 & 34 & 20 & 20 \\
Common + signed bank & 0.5 & 0.05 & 240 & 68 & 179 & 58 & 215 & 82 & 54 & 18 & 16 \\
Common + signed bank & 0.75 & 0 & 240 & 47 & 176 & 38 & 140 & 52 & 36 & 24 & 25 \\
Common + signed bank & 0.75 & 0.05 & 240 & 74 & 183 & 61 & 216 & 84 & 56 & 24 & 23 \\
Common + signed bank & 1 & 0 & 240 & 62 & 134 & 29 & 168 & 73 & 25 & 58 & 38 \\
Common + signed bank & 1 & 0.05 & 240 & 88 & 173 & 70 & 222 & 98 & 53 & 34 & 35 \\
Direct selector & 0 & 0 & 240 & 27 & 165 & 25 & 97 & 31 & 24 & 2 & 2 \\
Direct selector & 0 & 0.05 & 240 & 62 & 171 & 55 & 209 & 66 & 54 & 2 & 2 \\
Direct selector & 0.25 & 0 & 240 & 32 & 170 & 26 & 101 & 35 & 26 & 6 & 15 \\
Direct selector & 0.25 & 0.05 & 240 & 74 & 172 & 49 & 208 & 61 & 49 & 4 & 11 \\
Direct selector & 0.5 & 0 & 240 & 38 & 171 & 27 & 102 & 36 & 26 & 7 & 16 \\
Direct selector & 0.5 & 0.05 & 240 & 76 & 173 & 58 & 206 & 68 & 56 & 9 & 15 \\
Direct selector & 0.75 & 0 & 240 & 37 & 170 & 34 & 100 & 44 & 32 & 14 & 22 \\
Direct selector & 0.75 & 0.05 & 240 & 78 & 171 & 56 & 207 & 71 & 53 & 12 & 16 \\
Direct selector & 1 & 0 & 240 & 40 & 115 & 34 & 92 & 41 & 11 & 34 & 25 \\
Direct selector & 1 & 0.05 & 240 & 83 & 138 & 55 & 202 & 79 & 37 & 20 & 24 \\
\bottomrule
\end{tabular}

%% file: sections/empirical_cut_details.tex
\section{Empirical Cut Selector Details}
\label{app:empirical-cut-details}

\paragraph{Score cuts and the full cost interval.}
Fix $r_i$, binary $q_i$, group $G_i\in\{0,1\}$, and observed validation
amounts $a_i,b_i\geq0$. Let $\mathcal I_\star$ contain $q_i=0$ and
$\mathcal I_g$ contain $q_i=1,G_i=g$.
For each bucket $k$, a score cut $j$ gives
\[
 C_{kj}=\sum_{i\in\mathcal I_k}a_i\mathbf1(r_i\geq t_{kj}),
 \qquad
 L_{kj}=\sum_{i\in\mathcal I_k}b_i\mathbf1(r_i\geq t_{kj}).
\]
We include all-pass and full-block endpoints and advance cuts only after
complete equal-score batches. Per-row cuts inside such a batch would
require information beyond the score and bucket at decision time.

For $0\leq c\leq A_k^{\rm bucket}=\sum_{i\in\mathcal I_k}a_i$, define
\[
 H_k(c)=\max_{\omega\geq0}
 \left\{\sum_j\omega_jL_{kj}:
       \sum_j\omega_j=1,\quad\sum_j\omega_jC_{kj}=c\right\}.
\]
This finite upper boundary is concave and piecewise linear. Equal-cost
points contribute only their largest capture to this boundary.
We keep the full cost interval, including a flat final segment.
A larger cost at unchanged capture can still affect a cross-group gap
constraint, so capture--cost dominance alone does not justify removing
the right endpoint.

\paragraph{Empirical optimality and compact policies.}
Let $A_\Sigma=\sum_i a_i$ and $A_g=\sum_{i:G_i=g}a_i>0$.
For independent bucket mixtures, write $C_k$ and $L_k$ for their expected
blocked value and captured loss on validation.
The value budget depends on $C_\star+C_0+C_1$, while the group rates
are $D_g^A=C_g/A_g$. Fix $\epsilon\geq0$, $m_g\in[0,1]$, and reference
levels $D_g^A(\pi_0)$.

\begin{proposition}[Empirical score-cut optimality]
\label{prop:empirical-cut-optimality}
For the value budget, gap bound, and baseline group caps above, the LP
over upper-hull vertices attains the same maximum validation capture
as independent convex mixtures of all bucket score cuts.
An optimal policy uses at most two adjacent upper-hull cuts in each bucket.
\end{proposition}

\begin{proof}
Take any feasible mixtures of all cuts, producing costs $C_k$ and
captures $L_k$. Replace each bucket's point by $(C_k,H_k(C_k))$.
This weakly increases capture and preserves every constraint, because
the constraints depend only on the three bucket costs.
The full cost interval makes each replacement available.
Conversely, every hull vertex represents an original score cut, so
hull mixtures belong to the convexified cut family.
Both optimization problems therefore have the same optimum.
Every point on a piecewise-linear upper boundary lies between two
adjacent vertices. At an optimum we can replace each bucket mixture
by those two vertices at the same cost, preserving the optimum.
\end{proof}

For costs $C_k^-\leq C_k\leq C_k^+$ on such a segment, the higher-cost
vertex has weight
\[
 \lambda_k=\frac{C_k-C_k^-}{C_k^+-C_k^-}.
\]
Endpoint and one-vertex cases use the corresponding single cut.
We freeze these weights; new observations do not recalibrate them.

\paragraph{Zero-cost points and controls.}
A cut can capture positive validation loss while blocking zero observed
value. Its zero-cost point can dominate all-pass on the upper boundary.
We retain the actual all-pass policy independently as a control.
A nonempty bucket with total value zero uses its maximum-capture
zero-cost cut; an empty bucket uses all-pass.
These reductions concern the observed validation programme.
They do not imply that the retained cut has zero cost or dominates
another cut on new candidates.

Equal-cost highest-capture pruning uses the value-unit structure.
Count- or loss-unit complaint constraints can depend on additional
moments, so the same two-dimensional reduction need not preserve them.

\paragraph{Numerical validation.}
Let $p_{kj}$ be the LP's normalized capture coefficient, let $M$ and $h$
be its inequality matrix and RHS, and let $\xi\geq0$ be inequality
multipliers. The three bucket mixtures have separate simplex equalities.
For any feasible mixture, the maximization dual upper bound is
\[
 U=\xi^\top h+
   \sum_k\max_{j\in\mathcal H_k}
       \left(p_{kj}-(M^\top\xi)_{kj}\right).
\]
Each simplex averages coefficients by nonnegative weights summing to
one, giving its maximum as the support bound.
We verify every budget, gap, and group-cap constraint after compact
replacement and serialization, together with a two-sided numerical
gap to this upper bound. Inactive moments need not reproduce an
arbitrary numerical optimizer. The resulting guarantee concerns the
empirical convexified score-cut family; future constraints and capture
require separate evaluation.

\paragraph{Empirical cut outcomes.}
We retain all \ManuscriptEmpiricalCells{} cells at every prespecified gap
fraction and group margin. The primary full-test comparison uses the ungated
validation proposal. Applied full-test outcomes include the C sample and
remain descriptive. C acceptance checks group non-increase, capture, and
value; frozen and relative gap targets remain separate outcomes.

\begin{table*}[t]
\centering\small
\caption{Empirical score-cut outcomes at the half-gap target and zero group
margin on the full original test. $G/N/F$, $N_{all}/J$, $U$, $Q_{rel}/Q_{fix}$,
and $K$ follow the definitions in Table~\ref{tab:direct-control-outcomes}.}
\label{tab:empirical-cut-outcomes}
\input{sections/empirical_cut_primary_table}
\end{table*}

\begin{table*}[t]
\centering\small
\caption{Separate T outcomes after frozen C acceptance or exact baseline
fallback. Subscripts $P/E$ identify proposals and applied policies; high
non-increase after fallback must be read with acceptance and useful gains.}
\label{tab:empirical-cut-calibration}
\input{sections/empirical_cut_calibration_table}
\end{table*}

\begin{table*}[t]
\centering\scriptsize
\caption{All prespecified gap and group-margin targets for empirical cuts.
Relative held-out goals and the frozen numerical programme remain distinct.}
\label{tab:empirical-cut-sensitivity}
\input{sections/empirical_cut_sensitivity_table}
\end{table*}

\begin{table*}[t]
\centering\scriptsize
\caption{Primary empirical score-cut results by source. Each source retains
its original cohorts and seeds; these descriptive cell counts are not
independent inferential observations.}
\label{tab:empirical-cut-sources}
\input{sections/empirical_cut_source_table}
\end{table*}

\begin{table*}[t]
\centering\scriptsize
\caption{Relative and frozen full-target outcomes for empirical cuts. Frozen
programme feasibility excludes the capture floor; full targets include it.}
\label{tab:empirical-cut-full-targets}
\input{sections/empirical_cut_full_target_table}
\end{table*}

\paragraph{Direct constrained selection.}
We solve Eq.~\eqref{eq:direct-group-capped-objective} at a specified gap
fraction and per-group margin, with maximum capture as the objective.
The primary target halves the validation baseline gap and permits no
increase in either group. We retain the fixed gap and margin sensitivities.
The empirical selector learns capture--value curves from observed validation
amounts in three score strata and freezes at most two adjacent cuts per
stratum. We compare it with maximum-capture selectors over the common and
common-plus-signed banks and with the predicted-coefficient dual rule.
The latter uses the original value estimates and W-BCE-implied loss;
its fitted coefficients differ from the observed validation amounts.

\begin{table}[t]
\centering
\small
\caption{Maximum-capture selectors at the half-gap target and zero group
margin over \ManuscriptEmpiricalCells{} cells. $K$ counts validation optima
retaining at least $.95$ of reference capture under each selector's own
coefficients. $Q_{rel}$ counts full-test attainment of the half-gap,
no-increase, capture, and value targets. Retained capture is the test median.
$U$ counts smaller gaps than $.95$ baseline thinning without lower capture,
with both policies meeting the joint targets. Counts are descriptive.}
\label{tab:direct-constrained-selection}
\input{sections/empirical_cut_comparison_table}
\end{table}

The empirical cut learner reaches the fitted capture requirement in
\ManuscriptEmpiricalFullSelectionFloor{} cells and retains a median
\ManuscriptEmpiricalFullProposalRetentionPercent{}\% of test capture.
It meets the full relative target in \ManuscriptEmpiricalFullProposalQ{} cells
and improves the qualified comparison with thinning in
\ManuscriptEmpiricalFullProposalU{}; these outcomes do not establish a general
advantage over the common-bank control.

Every validation proposal precedes a separate C check and final T evaluation.
We also evaluate the ungated proposals on the full original test split.
The C gate checks group non-increase, retained capture, and value spend;
it does not refit a threshold or choose another target. Its acceptance
is distinct from attaining the requested gap and group margins.
Earlier minimum-gap matched controls appear in
Appendix~\ref{app:normalized-prices}.
The C gate accepts \ManuscriptEmpiricalTAccepted{} primary proposals;
the applied policy meets the relative full target in
\ManuscriptEmpiricalTAppliedQ{} T cells and retains qualified gains in
\ManuscriptEmpiricalTAppliedU{}.
Appendix~\ref{app:empirical-cut-details} gives all group outcomes, sources,
target sensitivities, and frozen numerical versus relative constraints.

%% file: sections/empirical_cut_primary_table.tex
\begin{tabular}{lrrrrrrrrrr}
\toprule
Method & $G$ & $N$ & $F$ & $N_{all}$ & $J$ & $U$ & $Q_{rel}$ & $Q_{fix}$ & $K$ & Retained (\%) \\
\midrule
Empirical score cuts & 186 & 87 & 24 & 100 & 27 & 23 & 15 & 15 & 234 & 97.86 \\
\bottomrule
\end{tabular}

%% file: sections/empirical_cut_calibration_table.tex
\begin{tabular}{lrrrrrrrrr}
\toprule
Method & $C$ & $N_P$ & $N_E$ & $J_P$ & $J_E$ & $U_P$ & $U_E$ & $Q_{rel,E}$ & $Q_{fix,E}$ \\
\midrule
Empirical score cuts & 27 & 113 & 233 & 29 & 123 & 23 & 9 & 4 & 13 \\
\bottomrule
\end{tabular}

%% file: sections/empirical_cut_sensitivity_table.tex
\begin{tabular}{rrrrrrrrrrr}
\toprule
$\gamma$ & $m$ & $V$ & $C$ & $G$ & $N$ & $F$ & $J$ & $U$ & $Q_{rel}$ & $Q_{fix}$ \\
\midrule
0 & 0 & 240 & 24 & 180 & 88 & 20 & 26 & 18 & 2 & 2 \\
0 & 0.05 & 240 & 42 & 180 & 139 & 30 & 44 & 29 & 2 & 2 \\
0.25 & 0 & 240 & 22 & 190 & 91 & 22 & 26 & 21 & 13 & 13 \\
0.25 & 0.05 & 240 & 39 & 190 & 149 & 37 & 47 & 34 & 11 & 12 \\
0.5 & 0 & 240 & 27 & 186 & 87 & 24 & 27 & 23 & 15 & 15 \\
0.5 & 0.05 & 240 & 46 & 189 & 143 & 37 & 46 & 36 & 13 & 13 \\
0.75 & 0 & 240 & 25 & 160 & 71 & 21 & 27 & 17 & 14 & 18 \\
0.75 & 0.05 & 240 & 42 & 166 & 122 & 33 & 47 & 30 & 13 & 16 \\
1 & 0 & 240 & 15 & 120 & 43 & 10 & 15 & 9 & 10 & 20 \\
1 & 0.05 & 240 & 37 & 133 & 94 & 28 & 43 & 23 & 11 & 20 \\
\bottomrule
\end{tabular}

%% file: sections/empirical_cut_source_table.tex
\begin{tabular}{lrrrrrrrr}
\toprule
Source & Cells & $C$ & $N_P$ & $N_E$ & $J_E$ & $U_E$ & $Q_{rel,E}$ & $Q_{fix,E}$ \\
\midrule
Australian MTPL & 20 & 2 & 11 & 19 & 11 & 1 & 0 & 0 \\
Belgian MTPL & 30 & 0 & 14 & 30 & 12 & 0 & 0 & 0 \\
Bondora & 10 & 1 & 5 & 10 & 4 & 0 & 0 & 0 \\
CMS provider fraud & 20 & 1 & 4 & 19 & 9 & 0 & 0 & 1 \\
EU health insurance & 10 & 2 & 4 & 8 & 5 & 0 & 0 & 0 \\
French portfolio & 90 & 11 & 50 & 88 & 49 & 1 & 1 & 7 \\
German credit & 20 & 0 & 1 & 20 & 15 & 0 & 0 & 2 \\
Norwegian auto & 10 & 5 & 10 & 10 & 6 & 4 & 2 & 1 \\
Swedish motorcycle & 10 & 1 & 8 & 10 & 2 & 0 & 0 & 0 \\
UCI credit & 20 & 4 & 6 & 19 & 10 & 3 & 1 & 2 \\
\bottomrule
\end{tabular}

%% file: sections/empirical_cut_full_target_table.tex
\begin{tabular}{rrrrrrrrr}
\toprule
$\gamma$ & $m$ & $P_{fix,P}$ & $P_{fix,E}$ & $Q_{fix,P}$ & $Q_{fix,E}$ & $Q_{rel,P}$ & $Q_{rel,E}$ & $U_{fix,E}$ \\
\midrule
0 & 0 & 8 & 2 & 3 & 2 & 3 & 2 & 2 \\
0 & 0.05 & 8 & 2 & 3 & 2 & 3 & 2 & 2 \\
0.25 & 0 & 25 & 12 & 12 & 10 & 8 & 2 & 3 \\
0.25 & 0.05 & 27 & 11 & 13 & 9 & 7 & 2 & 5 \\
0.5 & 0 & 33 & 16 & 19 & 13 & 11 & 4 & 3 \\
0.5 & 0.05 & 33 & 15 & 18 & 12 & 10 & 5 & 5 \\
0.75 & 0 & 32 & 22 & 20 & 19 & 14 & 4 & 3 \\
0.75 & 0.05 & 32 & 18 & 19 & 15 & 13 & 5 & 4 \\
1 & 0 & 35 & 36 & 23 & 34 & 14 & 124 & 1 \\
1 & 0.05 & 37 & 25 & 22 & 22 & 11 & 4 & 3 \\
\bottomrule
\end{tabular}

%% file: sections/empirical_cut_comparison_table.tex
\begin{tabular}{lrrrr}
\toprule
Method & $K$ & $Q_{rel}$ & Retained (\%) & $U$ \\
\midrule
Common bank & 126 & 17 & 91.74 & 35 \\
Common + signed bank & 222 & 20 & 99.08 & 34 \\
Predicted-coefficient dual & 208 & 7 & 99.83 & 26 \\
Empirical score cuts & 234 & 15 & 97.86 & 23 \\
\bottomrule
\end{tabular}

%% file: sections/objection_safe_details.tex
\section{Objection-Safe Selector Details}
\label{app:objection-safe-details}

\paragraph{Pointwise probabilities and total objections.}
Fix $0\leq\pi_0\leq1$, binary $q$, and unchanged nonnegative outcome
weights. For $0\leq s,h\leq1$, use
$\pi_{s,h}=sq\pi_0+(1-q)(\pi_0+(1-\pi_0)h)$.

\begin{proposition}[Objection-safe scaling]
\label{prop:objection-safe-scaling}
If $q\pi\leq q\pi_0$ pointwise, total objection mass and each group level
cannot increase for any nonnegative unit weight with positive finite
denominators. For $\pi_{s,h}$,
\[
 E[q\pi_{s,h}U]=sE[q\pi_0U],\qquad
 \frac{E[q\pi_{s,h}U\mid G=g]}{E[U\mid G=g]}
 =s\frac{E[q\pi_0U\mid G=g]}{E[U\mid G=g]}.
\]
Its normalized global level and each signed or absolute gap scale by $s$.
\end{proposition}

\begin{proof}
Multiply the pointwise inequality by $U\geq0$, integrate, and divide
by the fixed positive denominators. Since $q$ is binary,
$q\pi_{s,h}=sq\pi_0$, giving equality for the stated quantities.
Subtract group levels and use $s\geq0$ for absolute gaps.
\end{proof}

The guarantee concerns probabilities and expected metrics. A comparison
of realized block draws needs a specified coupling; exact gap scaling
does not hold for every independent draw.

\begin{corollary}[Constituent criterion contraction]
\label{cor:objection-safe-constituent-norms}
Fix each constituent criterion indicator $q_t\leq q$.
For $\pi_{s,h}$, $q_t\pi_{s,h}=sq_t\pi_0$ pointwise.
For any fixed subgroup event $H$ with $\Pr(H)>0$ and any unchanged
nonnegative weight $U$ with $0<E[U\mid H]<\infty$,
\[
 E[q_t\pi_{s,h}U\mid H]=sE[q_t\pi_0U\mid H].
\]
Each corresponding normalized objection level scales by $s$.
\end{corollary}

\begin{proof}
Since $q$ is binary and $q_t\leq q$, $q_tq=q_t$ and
$q_t(1-q)=0$. Multiply the policy formula by $q_t$, then integrate
against $U$ on $H$ and divide by its fixed positive denominator.
\end{proof}

The constituent indicators describe the frozen reference criteria whose
union defines $q$. The common factor applies across groups; its ratio
and criterion indicators may use group membership upstream.

\paragraph{Coupled block draws.}
For fixed probabilities and stakes, draw independent uniforms $\xi_i$
and use the same draw for each candidate's two policies:
\[
 Z_{0,i}=\mathbf1(\xi_i\leq\pi_{0,i}),\qquad
 Z_i=\mathbf1(\xi_i\leq\pi_{s,h,i}).
\]
Objecting thresholds are nested downward, giving
$q_iZ_i\leq q_iZ_{0,i}$ for every draw. Nonobjecting thresholds
are nested upward, so $(1-q_i)Z_i\geq(1-q_i)Z_{0,i}$.
Consequently, for any unchanged nonnegative stakes and any subgroup,
the realized objection sum cannot increase. The same comparison holds
for each constituent indicator $q_t\leq q$.
This coupling implements release of baseline objecting blocks and
addition of nonobjecting blocks. It does not impose exact proportional
gap scaling on each draw, and a realized gap may grow. Independently
randomizing the two policies does not guarantee the nested comparison.

\paragraph{Selective anchors and capture.}
Let $\pi_s=\pi_{s,0}$, $L_0=E[\pi_0B]$, $V_0=E[\pi_0A]$,
$L_{\rm obj}=E[q\pi_0B]$, and $V_{\rm obj}=E[q\pi_0A]$.

\begin{corollary}[Anchor capture condition]
\label{cor:objection-safe-anchor}
The anchor satisfies
\[
 L(\pi_s)=L_0-(1-s)L_{\rm obj},\qquad
 V(\pi_s)=V_0-(1-s)V_{\rm obj}\leq V_0.
\]
If $L_0>0$, $V_0\leq\beta$, and $0\leq\gamma,\rho\leq1$,
some anchor with $s\leq\gamma$ meets $L(\pi_s)\geq\rho L_0$
if and only if
\[
 (1-\gamma)\theta_B\leq1-\rho,\qquad
 \theta_B=L_{\rm obj}/L_0.
\]
\end{corollary}

\begin{proof}
Expand $\pi_s=\pi_0-(1-s)q\pi_0$. Capture increases with $s$,
so its maximum over $s\leq\gamma$ occurs at $\gamma$.
Every such anchor respects the baseline-feasible value budget.
\end{proof}

For fixed $s$, uniform $s\pi_0$ has the same objections and at most
the anchor's capture. Uniform thinning at
$u=1-(1-s)\theta_B$ instead matches the anchor's capture. Its relative
objection removal is $(1-s)\theta_B$, versus $1-s$ for the anchor,
where baseline objection mass is positive. Both policies have spend
at most $V_0$; their spends need not coincide.
If baseline captured loss is zero, $\theta_B$ is undefined and the
relative capture floor is vacuous.

\paragraph{Minimum keep factor and capture floors.}
Let $H_{\leq}(c)$ be the maximum additional nonobjecting capture from
the frozen score-cut mixture family at additional value at most $c$.
For $c\geq0$, this concave nondecreasing function saturates after all
available additional value; negative $c$ is resource-infeasible.
At a feasible keep factor, the maximum class capture is
\[
 \Lambda(s)=L_{\rm non}+sL_{\rm obj}
 +H_{\leq}(\beta-V_{\rm non}-sV_{\rm obj}).
\]
It is concave on the resource-feasible interval. Thus the factors
meeting $\Lambda(s)\geq\rho L_0$ form a closed interval, possibly empty.
We find its minimum by LP rather than a ratio or parameter grid.

If this programme is feasible, its optimum satisfies $s\leq\rho$.
For a feasible $s>\rho$, the anchor at $\rho$ has cost at most that
policy's cost because added value is nonnegative. Its capture is
$L_{\rm non}+\rho L_{\rm obj}\geq\rho L_0$, so it is feasible at a
smaller keep factor. Thus a feasible optimum removes at least $1-\rho$
of each positive baseline objection level.

With $V_0\leq\beta$ and $\theta_B>0$, the selective anchor
$s_a=\max(0,1-(1-\rho)/\theta_B)$ is feasible. Hence the optimized
removal is at least $\min(1,(1-\rho)/\theta_B)\geq1-\rho$.
For $\theta_B=0$, full removal is feasible. Capture-infeasible and
resource-infeasible cases retain their own fallback flags.
An empirical feasible minimum $s$ does not guarantee capture or
resource targets on new data; objection scaling still holds there.

\paragraph{Finite-cut calibration scope.}
Freeze $\pi_0$, $q$, and $K\geq1$ nonobjecting cut functions $h_j$
before calibration. The common class equals the convex hull of
the at most $2K$ policies
\[
 \pi_{\ell,j}=\ell q\pi_0
 +(1-q)\bigl(\pi_0+(1-\pi_0)h_j\bigr),
 \qquad \ell\in\{0,1\},\quad j=1,\ldots,K.
\]
For any $(s,\omega)$, use weights $(1-s)\omega_j$ and $s\omega_j$
at its two endpoints. Conversely, endpoint mixture weights give their
objecting endpoint marginal $s$ and cut marginal $\omega$.

\begin{proposition}[Finite-cut calibration bound]
\label{prop:objection-safe-calibration-bound}
Conditional on these fixed functions, let C contain $n\geq1$ independent
identically distributed observations from the target population,
independent of the functions' fitting data. Assume known domain bounds
$0\leq A\leq M_A<\infty$ and $0\leq B\leq M_B<\infty$.
Fix $v,\rho\in[0,1]$, $\delta\in(0,1)$, and put
\[
 t=\sqrt{\frac{\log(4K/\delta)}{2n}},\qquad
 \varepsilon_A=M_At,\qquad
 \varepsilon_B=(1+\rho)M_Bt.
\]
Writing $\widehat E_C$ for the sample mean, with probability at least
$1-\delta$, every policy $\pi$ in the common convex hull satisfies
\[
\begin{aligned}
 E[(\pi-v)A]&\leq\widehat E_C[(\pi-v)A]+\varepsilon_A,\\
 E[(\pi-\rho\pi_0)B]&\geq
 \widehat E_C[(\pi-\rho\pi_0)B]-\varepsilon_B.
\end{aligned}
\]
On this event, any C-selected policy satisfying
$\widehat E_C[(\pi-v)A]\leq-\varepsilon_A$ and
$\widehat E_C[(\pi-\rho\pi_0)B]\geq\varepsilon_B$
has $E[\pi A]\leq vE[A]$ and $E[\pi B]\geq\rho E[\pi_0B]$.
\end{proposition}

\begin{proof}
For each vertex, $(\pi-v)A$ has range length at most $M_A$,
and $(\pi-\rho\pi_0)B$ has range length at most $(1+\rho)M_B$.
Apply the relevant one-sided Hoeffding inequality to each moment.
A union bound over two moments and at most $2K$ vertices gives total
failure probability at most $\delta$. Both moments depend affinely on
$\pi$, so each one-sided bound extends to every convex mixture.
The buffered constraints then imply the population inequalities.
\end{proof}

Here $v$ is a value-budget fraction; its population absolute budget is
$\beta=vE[A]$. The bound uses the full V-frozen cut set, including cuts
subsequently removed by a C-dependent hull. It concerns population
expectations and does not impose a hard realized batch budget.
The current unbuffered experiments do not apply this bound. Their data
were previously examined, and we have not established these sampling
or known-domain-bound assumptions. Sample maxima do not supply the
required domain bounds.

\paragraph{Constant group scales and uncertain baseline levels.}
Let $I_0,I_1$ be specified intervals of nonnegative baseline levels,
and fix $\gamma\in[0,1]$.

\begin{proposition}[Robust constant scaling]
\label{prop:objection-safe-robust-scales}
Suppose $I_0\cap I_1$ contains $d>0$.
If constants $s_0,s_1\in[0,1]$ satisfy
\[
 |s_1d_1-s_0d_0|\leq\gamma|d_1-d_0|
 \quad\text{for all }(d_0,d_1)\in I_0\times I_1,
\]
then $s_0=s_1$. If the box also contains an unequal pair, their common
value is at most $\gamma$. Conversely, a common value $s\leq\gamma$
satisfies the inequality for every pair.
\end{proposition}

\begin{proof}
At $(d,d)$, the right side is zero, forcing $s_0=s_1$.
For an unequal pair, divide by $|d_1-d_0|>0$.
The converse follows by multiplying that absolute gap by $s$.
\end{proof}

This result concerns constant scales and the specified box; it supplies
no statistical coverage for data-estimated intervals.

\paragraph{Resource information.}
For fixed probability vectors $p,p_0$, the relation
$a^\top p\leq a^\top p_0$ for every positive stake vector $a$
holds if and only if $p\leq p_0$ pointwise. Sufficiency follows by
summation. If $p_i>p_{0,i}$, choose that stake arbitrarily large and
the other stakes fixed to reverse the inequality.
Therefore any positive nonobjecting addition prevents an unconditional
resource-domination guarantee with unknown unrestricted stakes.
With known batch values, additions satisfying
\[
 \sum_{q_i=0}a_i(1-\pi_{0,i})h_i
 \leq(1-s)\sum_{q_i=1}a_i\pi_{0,i}
\]
preserve expected spend relative to baseline. A hard realized budget
also needs realized or coupled budget-respecting decisions.

\paragraph{Total-stake information.}
Let $Z$ be the classifier's conditioning variables and $W=A+B$.
For $t(Z)=E[W\mid Z]\in(0,\infty)$, the weighted proper-loss target is
$y_W(Z)=E[B\mid Z]/t(Z)$. Hence
\[
 E[A\mid Z]=t(Z)(1-y_W(Z)),\qquad
 E[B\mid Z]=t(Z)y_W(Z).
\]
If $W$ is known from $Z$, then $t(Z)=W$. This follows from
$E[A\mid Z]+E[B\mid Z]=t(Z)$. Known total stake and a correctly
specified classifier recover conditional means, not realized
surviving value. The outcome probability $y_W$ is distinct from the
criterion-objection flag $q$.

\paragraph{Marginal capture scope.}
For a budget-optimal reference with finite multiplier $\mu$, let $h_R$
remove objecting probability and let $k$ add only nonobjecting
probability, with $0\leq h_R\leq\pi^\star$ and
$0\leq k\leq1-\pi^\star$ on their respective sets.
Assume these changes are measurable in $Z$. Use conditional coefficients $a=E[A\mid Z]$ and
$b=E[B\mid Z]$, and put $F=E[Ah_R]$, $C=E[Ak]\leq F$, and $c=b-\mu a$.
The capture-regret identity gives
\[
 L^\star-L_{\rm new}
 =E[h_Rc]+E[k(-c)]+\mu(F-C).
\]
Cheap removal alone is insufficient: useful additions need adequate
nonobjecting capacity at comparable loss per value.
With $F=C$, the bounds $c\leq\delta_R a$ on removals and
$-c\leq\delta_N a$ on additions imply capture loss at most
$F(\delta_R+\delta_N)$.
These margins concern true or coherently fitted coefficients;
an estimated score's proximity alone does not establish them.

The common empirical class keeps nonobjecting baseline spend fixed.
It is budget-feasible exactly when that minimum spend is within the
requested cap, since $s=0,h=0$ attains it. Its one additional-cut hull
supports at most two adjacent cuts at each chosen cost.
This optimization is empirical and class-specific. Capture and
resource calibration require separate evaluation, and a fallback outside
the class does not inherit its relative-gap guarantee.

\paragraph{Primary outcomes and calibration transfer.}
We retain all \ManuscriptFWCellTotal{} original cells at the primary
$\rho=.95$, $v=.10$ specification. The full comparison uses the V-frozen
proposal. Its mean value-unit objection removal is
\ManuscriptSafeFullProposedValueRemoval{}\%
(median \ManuscriptSafeFullProposedValueRemovalMedian{}\%); mean retained
capture is \ManuscriptSafeFullProposedMeanRetained{}\%
(median \ManuscriptSafeFullProposedRetained{}\%).
Tables~\ref{tab:objection-safe-primary}
and~\ref{tab:objection-safe-calibration} separate that comparison from
the C-frozen rule on final T. Applied full outcomes contain the C sample
and remain descriptive.

\begin{table}[t]
\centering\small
\setlength{\tabcolsep}{4pt}
\caption{V-frozen primary outcomes on the full original test split over
\ManuscriptFWCellTotal{} cells. $R$ is mean value-unit objection removal (\%);
$L$ is median retained capture (\%). $C$ and $V$ count attainment of the
capture floor and value cap; $J$ additionally requires non-increase in
both value-unit group rates. Same-$s$ controls use each fitted keep factor.}
\label{tab:objection-safe-primary}
\input{sections/objection_safe_primary_table}
\end{table}
\input{sections/objection_safe_calibration_table}

\begin{figure*}[t]
\centering
\includegraphics[width=\textwidth]{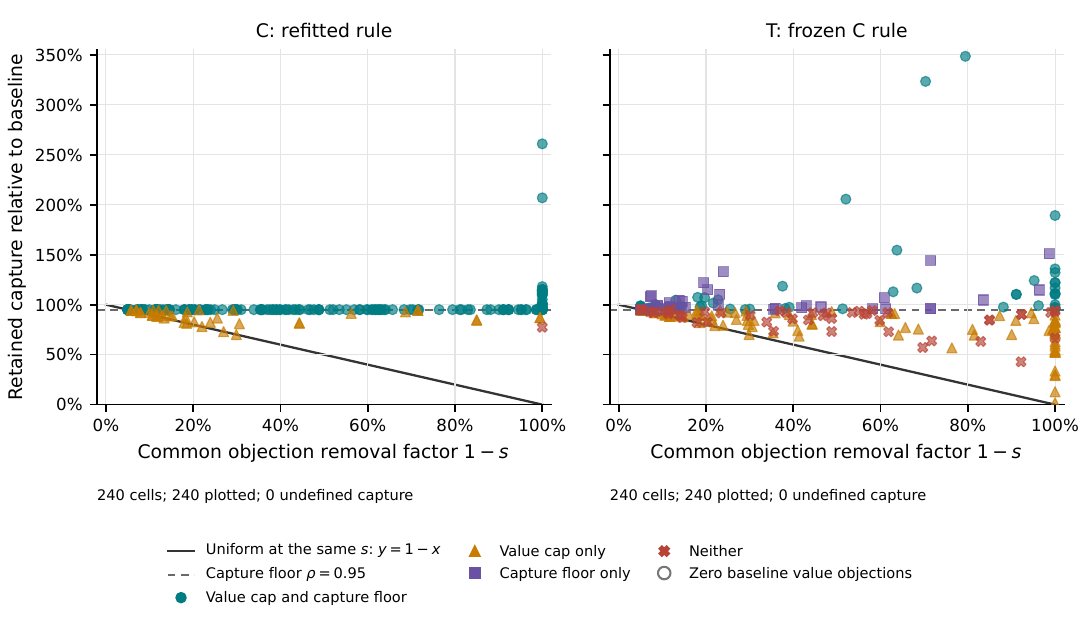}
\caption{Linear common contraction on C and the same
frozen rule on T. Each panel retains all \ManuscriptFWCellTotal{} cells;
support notes identify undefined capture ratios. The horizontal axis is
the saved factor $1-s$. It equals relative objection removal where baseline
mass is positive; hollow marks identify zero baseline value-objection mass.
Colors and symbols separate the capture and value checks. The diagonal
is uniform thinning at the same $s$, and the dashed line is the capture floor.}
\Description{Two panels plot common objection removal against retained
capture on calibration and final rows, using the same frozen C parameters.
The complete vertical range retains high-capture outcomes. The legend
distinguishes both business targets, either target alone, and neither.}
\label{fig:objection-safe-capture-removal}
\end{figure*}

\paragraph{Measurement units and support.}
Table~\ref{tab:objection-safe-units} reports aggregate removal separately
for count, value, and loss. On the full split, each unit has
\ManuscriptSafeFullProposedAllUnitsRemovalSupport{} positive baseline-mass
cases, so their common factors give identical mean and median removal.
On C, value-objection removal is undefined in
\ManuscriptSafeCAppliedValueRemovalUndefined{} cells with zero baseline mass.
The C programme completes in \ManuscriptSafeCCalibrationSolved{} cells;
strict value-objection removal and the joint constraints hold in
\ManuscriptSafeCAppliedValueStrictRemovedJoint{}.
Programme completion and a defined positive gain are distinct outcomes.

On T, count- and value-unit mean removal is
\ManuscriptSafeTAppliedValueRemoval{}\%
(median \ManuscriptSafeTAppliedValueRemovalMedian{}\%). Loss-unit removal
has mean \ManuscriptSafeTAppliedLossRemoval{}\%
(median \ManuscriptSafeTAppliedLossRemovalMedian{}\%) over
\ManuscriptSafeTAppliedLossRemovalSupport{} positive-mass cells;
\ManuscriptSafeTAppliedLossRemovalUndefined{} remains undefined.
Strict aggregate removal together with the value-rate and business checks
holds in \ManuscriptSafeTAppliedValueStrictRemovedJoint{} cells in value units
and \ManuscriptSafeTAppliedAllUnitsStrictRemovedJoint{} when strict removal
must be defined and positive in all three units. We retain every cell
in the scope and expose these support counts.

\begin{table*}[t]
\centering\small
\setlength{\tabcolsep}{4pt}
\caption{Aggregate objection removal on the full original test at the
primary target. $n_+$ counts positive baseline objection mass in the named
unit. Means and medians summarize defined cell ratios; $N$ counts
non-increase in both raw group objection masses. These counts do not
require the capture floor or value cap.}
\label{tab:objection-safe-units}
\input{sections/objection_safe_unit_table}
\end{table*}

\paragraph{Source outcomes and target sensitivities.}
Table~\ref{tab:objection-safe-sources} retains all ten source families
and their original cohorts and seeds. The overall means average defined
cell ratios; source summaries expose the unequal source sizes and variation
in removal, capture, and value transfer. These are descriptive outcomes,
with no claim that repeated cells supply independent source replications.
Table~\ref{tab:objection-safe-sensitivity} retains every prespecified
capture floor, gap envelope, design cap, and group margin. Tighter fitting
constraints have their own infeasibility flags and do not silently remove
cells from evaluation.

\begin{table*}[t]
\centering\small
\setlength{\tabcolsep}{4pt}
\caption{Primary V-frozen full-test results by source. $n$ retains every
original cell; $\theta_B$ is the validation baseline captured-loss share
on objecting candidates. $C$, $V$, and $J$ follow
Table~\ref{tab:objection-safe-primary}.}
\label{tab:objection-safe-sources}
\resizebox{\textwidth}{!}{%
\input{sections/objection_safe_source_table}}
\end{table*}

\begin{table*}[t]
\centering\scriptsize
\setlength{\tabcolsep}{3pt}
\caption{All prespecified V-frozen full-test arms. $K$ counts completed
validation programmes. The maximum-removal family fits its listed capture
floor $\rho$; the fixed-factor and group-specific families maximize capture
at their listed gap envelope $\gamma$, with $.95$ retained capture as an
evaluation requirement. $v$ is the design cap and $m$ the group margin.
Every row retains all \ManuscriptFWCellTotal{} cells, including flagged
fallbacks. $C$, $V$, and $J$ follow Table~\ref{tab:objection-safe-primary}.}
\label{tab:objection-safe-sensitivity}
\resizebox{\textwidth}{!}{%
\input{sections/objection_safe_sensitivity_table}}
\end{table*}

\paragraph{Group-specific safe-cut control.}
Table~\ref{tab:objection-safe-targeted} reports the prespecified
$\gamma=.5$, $m=0$, $v=.10$ control. Its C acceptance checks the already
fitted proposal; rejection applies the selective margin anchor. The
requested gap target, useful removal, and joint business checks remain
separate. In particular, preserving group levels after an anchor fallback
does not establish a gain or attainment of the requested gap.

\begin{table*}[t]
\centering\scriptsize
\setlength{\tabcolsep}{3pt}
\caption{Group-specific safe cuts at the half-gap target and zero margin.
$A_C$ counts C acceptance. Unit columns give mean aggregate removal, and
retained capture is the median. $Q$ counts the complete relative gap,
group-margin, capture, and value target; $J$ requires the capture, value,
and value-unit group non-increase checks. Full applied outcomes include C.}
\label{tab:objection-safe-targeted}
\resizebox{\textwidth}{!}{%
\input{sections/objection_safe_targeted_table}}
\end{table*}

\paragraph{Baseline direction across samples.}
Table~\ref{tab:objection-direction-stability} compares the V baseline's
signed value-unit gap with the full and T baselines using the same literal
group labels. We fixed the deadzone and materiality thresholds before
generating this post-run descriptive readout. Its complete frozen CSVs keep
zero and undefined direction transitions; the table summarizes strict flips.
Each evaluation policy contracts its own baseline group levels by the same
nonnegative $s$, regardless of which group has the larger baseline level.
Thus baseline direction changes across samples need no reversal of the
common controller's group treatment. All observed strict flips preserve
both group levels; the strict-removal and joint counts remain separate.
These direction changes describe empirical baseline rates; normative and
legal group roles are separate inputs, and sample flips do not establish
policy-induced harm or population sign changes.

\input{sections/objection_direction_table}

\paragraph{Cost of uniform contraction.}
Let $f=b-m$ be the adjusted loss coefficient, and let
$A_q=E[q\pi_0a]>0$ and $\bar r=E[q\pi_0 f]/A_q$.
Compare the full range of contested rates with a common retention
factor, using the same feasible set for the nonobjector policy and the
same blocked-value cap. The difference between their optimal values of
$E[\pi f]$ is at most
\[
 \frac12 E\big[q\pi_0\lvert f-\bar r a\rvert\big].
\]
To see this, write the contested rates of any feasible policy as
$q\pi=tq\pi_0$ with $0\leq t\leq1$, and choose the common factor
$s=E[tq\pi_0a]/A_q$. This keeps contested blocked value fixed.
Use the same nonobjector policy. The change in the objective from
this common policy to the selective policy is
$E[tq\pi_0(f-\bar r a)]$, since the centered coefficient has mean zero
under $q\pi_0$. Taking only positive contributions bounds the change by
half its mean absolute deviation. If the objection penalty is identical
per unit of value throughout the contested set, it cancels inside
$f-\bar r a$.

\paragraph{Converse for gap contraction.}
Suppose $q\pi\leq q\pi_0$ and there is a set of positive probability
with $q\pi_0=0$. If every absolute gap between two groups weakly
decreases in every nonnegative unit with defined levels, the release
has a common factor almost surely on $q\pi_0>0$.

To prove the converse, write $v=q\pi_0$ and let $r=q\pi/v$ wherever
$v>0$. If $r$ is nonconstant, choose two measurable sets of positive
probability on $v>0$ with different, separated ranges of $r$. Their
baseline count objection levels are positive. Put the zero-objection
set in the group with the larger level and weight it just enough to
equalize the two baseline levels. Give zero weight to the rest of
the population. Neither group's retention ratio changes with this
extra weight, since $q\pi=0$ wherever $v=0$. The new levels therefore
differ, opening a gap from zero.

\paragraph{Cost of pointwise protection.}
Let the baseline fill the blocked-value budget in yield order and let
the penalty be $m=q(\alpha a+\delta b)$ for fixed
$\alpha,\delta\geq0$. Then some optimizer of the unrestricted
penalty objective obeys $q\pi\leq q\pi_0$.
For $\delta<1$, the penalty preserves the order among contested
candidates and only lowers their scores. Any positive-value
contested increment beyond the baseline's rate therefore ranks no
higher than any increment in the baseline, including uncontested
ones. The baseline increments already fill the budget, so ties
can favor those increments. For $\delta\geq1$, the adjusted loss
on contested candidates is nonpositive, and the optimizer can
pass them. Any zero-value candidate with positive adjusted loss
already has positive loss and is fully blocked by a yield-optimal
baseline. The pointwise cap thus has zero objective cost in this
case. Count penalties and different penalties across criterion
patterns can change the relative order, and the same argument
then need not apply.

\paragraph{Capture floor and penalty.}
Write the floor problem as minimizing $W(\pi)$ subject to
$C(\pi)\geq\rho C_0$ and the blocked-value budget. If a positive
finite multiplier $\nu$ attains the floor dual in a convex policy
class, a floor optimum also minimizes $W(\pi)-\nu C(\pi)$ under
the budget. The corresponding penalty weight in
Eq.~\eqref{eq:weighted-objection-objective} is
$\eta=1/(C_0\nu)$. A penalty at the slope of a linear frontier
segment can give the same objective to every point on that segment.
An explicit capture floor selects the desired operating point.

\paragraph{Additional protected criteria.}
Fix the baseline and the existing union $q$. Adding a criterion
with pass probability $q_{\rm new}$ removes replacement value equal
to $E[(1-\pi_0)(1-q)\mathbf1(q_{\rm new}>0)A]$.
The identity counts candidates newly covered by the union and their
remaining block probability. In the add-only controller, a newly
contested baseline block can also become eligible for release.

%% file: sections/objection_safe_primary_table.tex
\begin{tabular}{lrrrrr}
\toprule
Policy & $R$ & $L$ & $C$ & $V$ & $J$ \\
\midrule
Removal fit & 42.90 & 94.70 & 100 & 155 & 64 \\
Selective, same $s$ & 42.90 & 85.86 & 30 & 221 & 25 \\
Uniform, same $s$ & 42.90 & 65.89 & 4 & 228 & 2 \\
Uniform $0.95$ & 5.00 & 95.00 & 240 & 193 & 193 \\
\bottomrule
\end{tabular}

%% file: sections/objection_safe_calibration_table.tex
\begin{table}[t]
\centering
\small
\setlength{\tabcolsep}{4pt}
\caption{Final-subset outcomes at $\rho=0.95$ and design value cap $0.10$. Common C fitting completed the requested programme in 197 of 240 cells. C fitting precedes T evaluation; completion is a sample feasibility result. $R$ is mean value-unit objection removal (\%), and $L$ is median retained capture relative to baseline (\%); both use defined ratios. $C$ meets the capture floor, $V$ meets the design cap, and $J$ also requires both value complaint rates not to increase.}
\label{tab:objection-safe-calibration}
\begin{tabular}{lrrrrr}
\toprule
Policy & $R$ & $L$ & $C$ & $V$ & $J$ \\
\midrule
V rule & 42.90 & 94.88 & 102 & 147 & 63 \\
C rule & 46.46 & 92.97 & 93 & 154 & 56 \\
Uniform C $s$ & 46.46 & 63.59 & 7 & 224 & 5 \\
Uniform $0.95$ & 5.00 & 95.00 & 240 & 186 & 186 \\
\bottomrule
\end{tabular}

\end{table}

%% file: sections/objection_safe_unit_table.tex
\begin{tabular}{llrrrr}
\toprule
Policy & Unit & $n_+$ & Mean removal (\%) & Median removal (\%) & $N$ \\
\midrule
Removal fit & Count & 240 & 42.90 & 34.11 & 240 \\
Removal fit & Value & 240 & 42.90 & 34.11 & 240 \\
Removal fit & Loss & 240 & 42.90 & 34.11 & 240 \\
Selective release at fitted $s$ & Count & 240 & 42.90 & 34.11 & 240 \\
Selective release at fitted $s$ & Value & 240 & 42.90 & 34.11 & 240 \\
Selective release at fitted $s$ & Loss & 240 & 42.90 & 34.11 & 240 \\
Uniform at fitted $s$ & Count & 240 & 42.90 & 34.11 & 240 \\
Uniform at fitted $s$ & Value & 240 & 42.90 & 34.11 & 240 \\
Uniform at fitted $s$ & Loss & 240 & 42.90 & 34.11 & 240 \\
Uniform at $s=0.95$ & Count & 240 & 5.00 & 5.00 & 240 \\
Uniform at $s=0.95$ & Value & 240 & 5.00 & 5.00 & 240 \\
Uniform at $s=0.95$ & Loss & 240 & 5.00 & 5.00 & 240 \\
\bottomrule
\end{tabular}

%% file: sections/objection_safe_source_table.tex
\begin{tabular}{lrrrrrrr}
\toprule
Source & $n$ & Median $\theta_B$ (\%) & Mean removal (\%) & Median retained (\%) & $C$ & $V$ & $J$ \\
\midrule
Australian MTPL & 20 & 84.23 & 32.90 & 92.68 & 7 & 10 & 4 \\
Belgian MTPL & 30 & 38.75 & 42.34 & 93.01 & 11 & 12 & 4 \\
Bondora & 10 & 99.42 & 5.04 & 95.02 & 7 & 5 & 4 \\
CMS provider fraud & 20 & 58.43 & 64.27 & 97.31 & 14 & 14 & 10 \\
EU health insurance & 10 & 19.82 & 46.13 & 94.04 & 3 & 3 & 0 \\
French portfolio & 90 & 61.46 & 54.21 & 90.77 & 38 & 64 & 27 \\
German credit & 20 & 58.32 & 27.03 & 93.43 & 7 & 17 & 4 \\
Norwegian auto & 10 & 3.76 & 81.91 & 89.87 & 4 & 9 & 4 \\
Swedish motorcycle & 10 & 73.82 & 23.02 & 94.21 & 1 & 5 & 1 \\
UCI credit & 20 & 97.87 & 5.11 & 94.97 & 8 & 16 & 6 \\
\bottomrule
\end{tabular}

%% file: sections/objection_safe_sensitivity_table.tex
\begin{tabular}{lrrrrrrrrr}
\toprule
Family & $\rho$ or $\gamma$ & $v$ & $m$ & $K$ & Mean removal (\%) & Median retained (\%) & $C$ & $V$ & $J$ \\
\midrule
Maximum removal & 0.9 & 0.095 & 0 & 227 & 50.50 & 89.86 & 107 & 158 & 68 \\
Maximum removal & 0.9 & 0.1 & 0 & 240 & 55.70 & 89.87 & 110 & 171 & 76 \\
Maximum removal & 0.95 & 0.095 & 0 & 162 & 37.43 & 93.98 & 71 & 139 & 39 \\
Maximum removal & 0.95 & 0.1 & 0 & 240 & 42.90 & 94.70 & 100 & 155 & 64 \\
Maximum removal & 0.98 & 0.095 & 0 & 80 & 32.83 & 94.51 & 24 & 137 & 20 \\
Maximum removal & 0.98 & 0.1 & 0 & 240 & 33.33 & 97.63 & 91 & 143 & 53 \\
Maximum removal & 1 & 0.095 & 0 & 65 & 30.91 & 94.64 & 17 & 136 & 15 \\
Maximum removal & 1 & 0.1 & 0 & 240 & 25.16 & 99.98 & 82 & 138 & 47 \\
Fixed common factor & 0 & 0.095 & 0 & 239 & 100.00 & 68.48 & 33 & 171 & 22 \\
Fixed common factor & 0 & 0.1 & 0 & 240 & 100.00 & 70.39 & 39 & 176 & 28 \\
Fixed common factor & 0.25 & 0.095 & 0 & 239 & 76.86 & 75.68 & 40 & 160 & 27 \\
Fixed common factor & 0.25 & 0.1 & 0 & 240 & 76.14 & 77.96 & 50 & 163 & 33 \\
Fixed common factor & 0.5 & 0.095 & 0 & 239 & 54.77 & 82.89 & 49 & 155 & 33 \\
Fixed common factor & 0.5 & 0.1 & 0 & 240 & 53.18 & 85.76 & 67 & 156 & 39 \\
Fixed common factor & 0.75 & 0.095 & 0 & 239 & 35.32 & 90.21 & 57 & 143 & 33 \\
Fixed common factor & 0.75 & 0.1 & 0 & 240 & 31.47 & 93.19 & 106 & 143 & 54 \\
Fixed common factor & 0.9 & 0.095 & 0 & 239 & 25.98 & 93.64 & 78 & 136 & 44 \\
Fixed common factor & 0.9 & 0.1 & 0 & 240 & 19.37 & 97.22 & 153 & 137 & 80 \\
Group-specific safe cuts & 0 & 0.095 & 0 & 239 & 36.33 & 95.23 & 121 & 124 & 56 \\
Group-specific safe cuts & 0 & 0.095 & 0.05 & 239 & 38.53 & 93.17 & 91 & 125 & 44 \\
Group-specific safe cuts & 0 & 0.1 & 0 & 240 & 33.05 & 97.48 & 145 & 126 & 69 \\
Group-specific safe cuts & 0 & 0.1 & 0.05 & 240 & 35.84 & 94.89 & 116 & 134 & 55 \\
Group-specific safe cuts & 0.25 & 0.095 & 0 & 239 & 34.69 & 95.25 & 122 & 127 & 61 \\
Group-specific safe cuts & 0.25 & 0.095 & 0.05 & 239 & 36.81 & 93.32 & 95 & 129 & 46 \\
Group-specific safe cuts & 0.25 & 0.1 & 0 & 240 & 30.56 & 97.73 & 147 & 127 & 71 \\
Group-specific safe cuts & 0.25 & 0.1 & 0.05 & 240 & 33.35 & 95.07 & 122 & 136 & 60 \\
Group-specific safe cuts & 0.5 & 0.095 & 0 & 239 & 33.78 & 95.09 & 122 & 122 & 60 \\
Group-specific safe cuts & 0.5 & 0.095 & 0.05 & 239 & 35.82 & 93.80 & 98 & 127 & 48 \\
Group-specific safe cuts & 0.5 & 0.1 & 0 & 240 & 28.30 & 98.26 & 161 & 130 & 80 \\
Group-specific safe cuts & 0.5 & 0.1 & 0.05 & 240 & 31.09 & 95.58 & 136 & 136 & 67 \\
Group-specific safe cuts & 0.75 & 0.095 & 0 & 239 & 33.20 & 95.16 & 123 & 118 & 58 \\
Group-specific safe cuts & 0.75 & 0.095 & 0.05 & 239 & 35.17 & 94.32 & 103 & 122 & 51 \\
Group-specific safe cuts & 0.75 & 0.1 & 0 & 240 & 26.76 & 98.79 & 165 & 125 & 82 \\
Group-specific safe cuts & 0.75 & 0.1 & 0.05 & 240 & 29.70 & 96.30 & 143 & 133 & 74 \\
Group-specific safe cuts & 0.9 & 0.095 & 0 & 239 & 33.13 & 96.17 & 128 & 118 & 59 \\
Group-specific safe cuts & 0.9 & 0.095 & 0.05 & 239 & 34.99 & 94.69 & 109 & 120 & 51 \\
Group-specific safe cuts & 0.9 & 0.1 & 0 & 240 & 26.48 & 99.23 & 169 & 120 & 81 \\
Group-specific safe cuts & 0.9 & 0.1 & 0.05 & 240 & 29.39 & 96.92 & 150 & 130 & 78 \\
\bottomrule
\end{tabular}

%% file: sections/objection_safe_targeted_table.tex
\begin{tabular}{llrrrrrrr}
\toprule
Subset & Rule & $A_C$ & Count removal (\%) & Value removal (\%) & Loss removal (\%) & Median retained (\%) & $Q$ & $J$ \\
\midrule
C & Proposed & 72 & 26.34 & 29.06 & 25.81 & 97.91 & 34 & 72 \\
C & Applied & 72 & 6.53 & 7.83 & 4.93 & 100.00 & 37 & 135 \\
T & Proposed & 72 & 25.99 & 28.08 & 26.69 & 98.57 & 30 & 81 \\
T & Applied & 72 & 6.33 & 7.14 & 6.01 & 100.00 & 14 & 117 \\
full & Proposed & 72 & 26.11 & 28.30 & 25.66 & 98.26 & 30 & 80 \\
full & Applied & 72 & 6.42 & 7.44 & 5.36 & 100.00 & 20 & 122 \\
\bottomrule
\end{tabular}

%% file: sections/objection_direction_table.tex
\begin{table}[H]
\centering
\small
\setlength{\tabcolsep}{3pt}
\caption{Frozen baseline value-unit complaint directions at the primary removal target. Each comparison retains 240 cells. A flip has opposite signs beyond the $10^{-10}$ deadzone; both absolute gaps must exceed $m$. $N$ counts both-group complaint nonincrease within flips; $J_+$ also requires strict value-objection removal, retained capture at least $0.95$, and blocked-value fraction at most $0.10$. Gap summaries use each flip's smaller absolute endpoint gap, in normalized group-rate units. Full uses the V proposal; T uses the C-frozen rule.}
\label{tab:objection-direction-stability}
\begin{tabular}{lrrrrrr}
\toprule
Pair & $m$ & $F$ & $N$ & $J_+$ & Min gap & Median gap \\
\midrule
V/full & 0 & 52 & 52 & 14 & 4.92887e-06 & 0.00508057 \\
V/full & 0.001 & 47 & 47 & 14 & 0.00102485 & 0.00582975 \\
V/full & 0.005 & 27 & 27 & 7 & 0.0050016 & 0.0222316 \\
V/full & 0.01 & 15 & 15 & 4 & 0.0120362 & 0.0315298 \\
V/T & 0 & 61 & 61 & 13 & 4.92887e-06 & 0.00535558 \\
V/T & 0.001 & 53 & 53 & 11 & 0.00118193 & 0.00682623 \\
V/T & 0.005 & 34 & 34 & 8 & 0.00501377 & 0.00984534 \\
V/T & 0.01 & 16 & 16 & 4 & 0.0108218 & 0.0301136 \\
\bottomrule
\end{tabular}

\end{table}